\documentclass[12pt]{article}
\usepackage[a4paper, margin=2cm]{geometry}
\usepackage{mathtools}

\DeclarePairedDelimiter\floor{\lfloor}{\rfloor}
\usepackage{authblk}
\usepackage[normalem]{ulem}

\usepackage{subfiles}
\usepackage{xr}
\usepackage[x11names]{xcolor}
\usepackage{sidecap}
\usepackage{booktabs}
\usepackage{mdframed}
\usepackage{graphicx}
\usepackage{wrapfig}
\usepackage[section]{placeins}

\usepackage[backend=bibtex, style=numeric, sorting=none]{biblatex}
\title{How to find a unicorn: Model-free detection of unique events in time series}

\author[1,2]{Zsigmond Benk\H{o}}
\author[1]{Tamás Bábel} 
\author[1]{Zoltán Somogyvári}

\affil[1]{Wigner Research Center for Physics, Department of Computational Sciences, Konkoly-Thege Miklos road 29-33., Budapest, H-1121, Hungary}
\affil[2]{János Szentágothai Doctoral School of Neurosciences, Semmelweis University, Ull\H{o}i road 26., Budapest, H-1085, Hungary}
\date{}

\newcommand{\soma}[1]{{\color{black}{#1}}}
\newcommand{\zsiga}[1]{{\color{black}{#1}}}

\makeatletter
\newcommand*{\addFileDependency}[1]{
	\typeout{(#1)}
	\@addtofilelist{#1}
	\IfFileExists{#1}{}{\typeout{No file #1.}}
}
\makeatother

\newcommand*{\myexternaldocument}[1]{%
    \externaldocument{#1}%
    \addFileDependency{#1.tex}%
    \addFileDependency{#1.aux}%
}

\myexternaldocument{SI}
\DeclareUnicodeCharacter{2009}{\,}

\begin{document}
\maketitle

\begin{abstract}
Recognition of anomalous events is a challenging but critical task in many scientific and industrial fields, especially when the properties of anomalies are unknown. 
In this paper, we introduce a new anomaly concept called “unicorn” or unique event and present a new, model-free, unsupervised detection algorithm to detect unicorns.
The key component of the new algorithm is the Temporal Outlier Factor (TOF) to measure the uniqueness of events in continuous data sets from dynamic systems.
The concept of unique events differs significantly from traditional outliers in many aspects: while repetitive outliers are no longer unique events, a unique event is not necessarily an outlier; it does not necessarily fall out from the distribution of normal activity.
The performance of our algorithm was examined in recognizing unique events on different types of simulated data sets with anomalies and it was compared with the Local Outlier Factor (LOF) and discord discovery algorithms.
TOF had superior performance compared to LOF and discord algorithms even in recognizing traditional outliers and it also recognized unique events that those did not.
Benefits of the unicorn concept and the new detection method were illustrated by example data sets from very different scientific fields.
Our algorithm successfully recognized unique events in those cases where they were already known such as the gravitational waves of a binary black hole merger on LIGO detector data and the signs of respiratory failure on ECG data series.
Furthermore, unique events were found on the LIBOR data set of the last 30 years.
\end{abstract}

    \begin{mdframed}[backgroundcolor=blue!20,linecolor=blue!20]
    \textbf{Significance statement}    
    
    {Anomalies in time series are rare and abnormal patterns that can be signs of transient, but significant changes, and therefore their automatic detection is often critical. 
This is especially difficult when we do not know which parameter of the anomalous pattern differs from normal activity.
We have developed a new anomaly detection method that measures the uniqueness of events in time series and based on this, finds special, unique patterns that we have named “unicorns”.
We have shown that this approach, in addition to finding the anomalies that traditional methods do, also recognizes anomalies that they do not.
This is demonstrated on data sets from different fields, from gravitational waves through ECG to economic indicators.}
	\end{mdframed}

Anomalies in time series are rare and non-typical patterns that deviate from normal observations and may indicate a transiently activated mechanism different from the generating process of normal data. Accordingly, recognition of anomalies is often important or critical, invoking interventions in various industrial and scientific applications.

Anomalies can be classified according to various aspects \cite{Chandola09,Blazquez-Garcia20}. These non-standard observations can be point outliers, whose amplitude is out of range from the standard amplitude or contextual outliers, whose measured values do not fit into some context. Combination of values can also form an anomaly named a collective outlier.
Thus, in case of point outliers, a single point is enough to distinguish between normal and anomalous states, whilst in the case of collective anomalies a pattern of multiple observations is required.
Two characteristic examples of extreme events are black swans and dragon kings, distinguishable by their generation process \cite{Taleb07,Sornette09}.
Black swans are generated by a powerlaw process and they are usually unpredictable by nature.
In contrast, the dragon king, such as stock market crashes, occurs after a phase transition and it is generated by different mechanisms from normal samples making it more predictable.
Both black swans and dragon kings are extreme events easily recognisable post-hoc (retrospectively), but not all the anomalies are so effortless to detect.
Even post-hoc detection can be a troublesome procedure when the amplitude of the event does not fall out of the data distribution.

Although the definition of an anomaly is not straightforward, two of its key features include rarity and dissimilarity from normal data.

Most, if not all the outlier detection algorithms approach the anomalies from the dissimilarity point of view.
They search for the most distant and deviant points without much emphasis on their rarity.
In contrast, our approach is the opposite: we quantify the rarity of a state, largely independent of the dissimilarity.

Here we introduce a new type of anomaly, the unique event, which is not an outlier in the classical sense of the word: it does not necessarily lie out from the background distribution, neither point-wise, nor collectively.
A unique event is defined as a unique pattern which appears only once during the investigated history of the system.
Based on their hidden nature and uniqueness one could call these unique events "unicorns" and add them to the strange zoo of anomalies.
Note that unicorns can be both traditional outliers appearing only once or patterns that do not differ from the normal population in any of their parameters.

But how do you find something you've never seen before, and the only thing you know about is that it only appeared once?

The answer would be straightforward for discrete patterns, but for continuous variables, where none of the states are exactly the same, it is challenging to distinguish the really unique states from a dynamical point of view.

Classical supervised, semi-supervised and unsupervised strategies have been used to detect anomalies \cite{Hodge04, Chandola09, Pimentel24} and recently deep learning techniques \cite{Chalapathy19, Kwon19, Braei20} were applied to detect extreme events of complex systems \cite{Qi20}.
Supervised outlier detection techniques can be applied to identify anomalies, when labeled training data is available for both normal and outlier classes.
Semi-supervised techniques also utilize labeled training data, but this is limited to the normal or the outlier class. \soma{Some of the semi-supervised methods do not need perfectly anomaly-free data to learn the normal class, but allows some outlier-contamination even in the training data \cite{Beggel19}}.
Model based pattern matching techniques can be applied to detect specific anomalies with best results when the mechanism causing the anomaly is well known and simple \cite{Abbott16}.
However when the background is less well known or the system is too complex to get analytical results (or to run detailed simulations), it is hard to detect even specific types of anomalies with model-based techniques due to the unknown nature of the waveforms.
Model-free unsupervised outlier detection techniques can be applied to detect unexpected events from time series in cases when no tractable models or training data is available
\soma{
The closest concept to our unicorns in the anomaly detection literature is the discord, defined as the unique subsequence, which is the farthest from the rest of the (non-overlapping) time series \cite{Keogh05}. Multiple model-free unsupervised anomaly detection methods have been built based on the discord concept \cite{Keogh05, Senin15}.
Other unsupervised anomaly detection techniques, such as the Local Outlier Factor (LOF) algorithm \cite{Breuniq00} are based on k Nearest Neighbor (kNN) distances.
The LOF algorithm was also adapted to time series data by Oehmcke et al.\cite{Oehmcke15}.
}

\begin{figure}[htb!]
\centering
\includegraphics[scale=0.6]{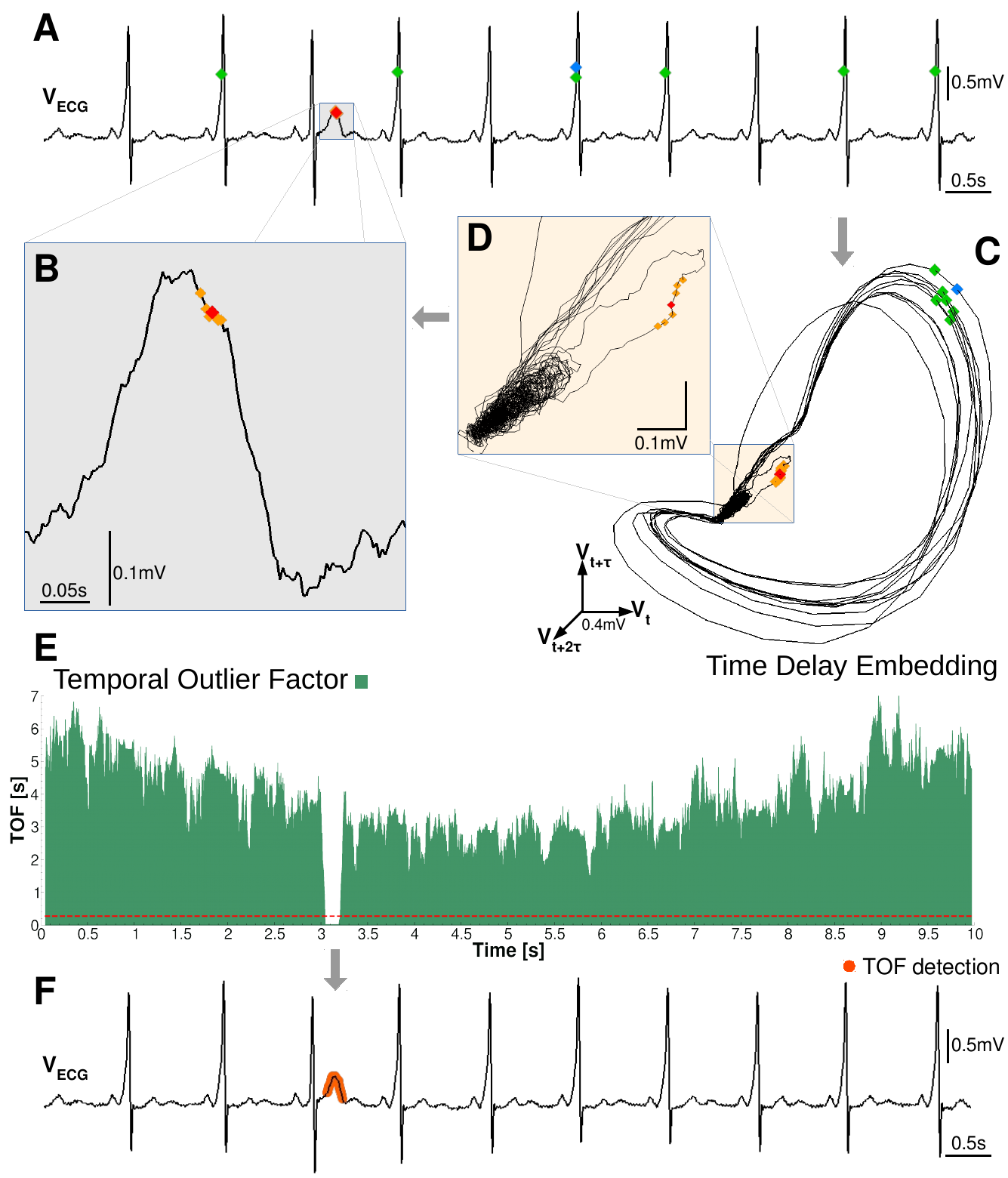}
\caption{\textbf{Schema of our unique event detection method and the Temporal Outlier Factor (TOF).}
(A) An ECG time series from a patient with Wolff-Parkinson-White Syndrome, a strange and unique T wave zoomed on the graph B. 
(C) The reconstructed attractor in the 3D state space by time delay embedding ($E=3, \tau=0.011\,s$). Two example states (red and blue diamonds) and their 6 nearest neighbors in the state space (orange and green diamonds respectively) are shown. The system returned several times back to the close vicinity of the blue state, thus the green diamonds are evenly distributed in time, on graph A. In contrast, the orange state space neighbors of the red point (zoomed on graph D) are close to the red point in time as well on graph B. These low temporal distances are show that the red point marks a unique event. (E) TOF measures the temporal dispersion of the $k$ nearest state space neighbors ($k=20$). Red dashed line is the threshold $\theta=0.28s$. Low values of TOF below the threshold mark the unique events, denoted by orange dots on the original ECG data on graph F.
}
\label{fig:berlin1}
\end{figure}

To adapt collective outlier-detection to time series data, nonlinear time series analysis provides the possibility to generate the multivariate state space from scalar observations.
The dynamical state of the system can be reconstructed from scalar time series \cite{Packard80} by taking the temporal context of each point according to Takens' embedding theorem \cite{Takens81}. This can be done via time delay embedding:
\begin{equation}\label{eq:embed}
    X(t) = [ x(t), x(t+\tau), x(t+2\tau), \ldots x(t+(E-1)\tau) ]
\end{equation}
where $X(t)$ is the reconstructed state at time $t$, $x(t)$ is the scalar time series. The procedure has two parameters: the embedding delay ($\tau$) and the embedding dimension ($E$).

\soma{Starting from an initial condition, the state of a dynamical system typically converges to a subset of its space space and forms a lower dimensional manifold, called the attractor, which describes the dynamics of the system in the long run.}
If E is sufficiently big ($E > 2*d$) compared to the dimension of the attractor ($d$), then the embedded (reconstructed) space is topologically equivalent to the system's state space, given some mild conditions on the the observation function generating the $x(t)$ time series are also met \cite{Takens81}.

As a consequence of Takens' theorem, small neighborhoods around points in the reconstructed state-space also form neighborhoods in the original state space, therefore a small neighborhood around a point represents nearly similar states.
This topological property has been leveraged to perform nonlinear prediction \cite{Ye15}, noise filtering \cite{Schreiber96,Hamilton16} and causality analysis \cite{Sugihara12, Benko19, Selmeczy19, Benko18}. 
Naturally, time delay embedding can be introduced as a preprocessing step before outlier detection (with already existing methods i.e. LOF) to create the contextual space for collective outlier detection from time series.

Besides the spatial information preserved in reconstructed state space, temporal relations in small neighborhoods can contain clues about the dynamics.
For example recurrence time statistics were applied to discover nonstationary time series \cite{Kennel97, Rieke02}, to measure attractor dimensions \cite{Gao99, Carletti06, Marwan07} and to detect changes in dynamics \cite{Gao13, Martinez-Rego16}.

 In the followings, we present a new model-free unsupervised anomaly detection algorithm to detect unicorns (unique events), that builds on nonlinear time series analysis techniques such as time delay embedding \cite{Takens81} and upgrades time-recurrence based non-stationarity detection methods \cite{Kennel97} by defining a local measure of uniqueness for each point.
\soma{
We validate the new method on simulated data, compare its performance with other modell-free unsupervised algorithms \cite{Breuniq00,Keogh05,Senin15} and we apply the new method to real-world data series, where the unique event is already known.
}

\section*{Results}
\subsection*{Temporal Outlier Factor}

The key question in unicorn search is how to measure the uniqueness of a state, as this is the only attribute of a unique event.
The simplest possible definition would be, that a unique state is one visited only once in the time series. A problem with this definition arises in the case of continuous valued observations, where almost every state is visited only once.
Thus, a different strategy should be applied to find the unicorns.
Our approach is based on measuring the temporal dispersion of the state-space neighbors.
If state space neighbors are separated by large time intervals, then the system returned to the same state time-to-time.
In contrast, if all the state space neighbors are temporal neighbors as well, then the system never returned to that state again.
This concept is shown on an example ECG data series from a patient with Wolff-Parkinson-White (WPW) Syndrome (Fig.\,\ref{fig:berlin1}).
The WPW syndrome is due to an aberrant atrio-ventricular connection in the heart. Its diagnostic signs are shortened PR-interval and appearance of the delta wave, a slurred upstroke of the QRS complex.
However, for our representational purpose, we have chosen a data segment, which contained one strange T wave with uniquely high amplitude (Fig.\,\ref{fig:berlin1}\,\textbf{A}).

To quantify the uniqueness on a given time series, the Temporal Outlier Factor (TOF) is calculated in the following steps (Fig.\,\ref{fig:berlin1}, S1):

Firstly, we reconstruct the system's state by time delay embedding (Eq.\,\ref{eq:embed}), resulting in a manifold, topologically equivalent to the attractor of the system (Fig.\,\ref{fig:berlin1}\,\textbf{C-D} and Fig.\,S1\,\textbf{B}).

Secondly, we search for the kNNs in the state space at each time instance on the attractor.
Two examples are shown on Fig.\,\ref{fig:berlin1}\,\textbf{C}: a red and a blue diamond and their 6 nearest neighbors marked by orange and green diamonds respectively. 

Thirdly, the Temporal Outlier Factor ($TOF$) is computed from the time indices of the kNN points (Fig.\,S1\,\textbf{C}):
\begin{equation}\label{eq:TOF}
	\mathrm{TOF} \left( t \right) =  \sqrt[\leftroot{-2}\uproot{16} q]{ \frac{\sum_{i=1}^{k}{\left| t-t_i \right| ^q} }{k} }.
\end{equation}
Where $t$ is the time index of the sample point ($X(t)$) and $t_i$ is the time index of the $i$-th nearest neighbor in reconstructed state-space.
Where $q\in \mathcal{R}^{+}$, in our case we use $q=2$ (Fig.\,\ref{fig:berlin1}\,\textbf{E}).

As a final step for identifying unicorns, a proper threshold $\theta$ should be defined for TOF (Fig.\,\ref{fig:berlin1}\,\textbf{E}, dashed red line), to mark unique events (orange dots, Fig.\,\ref{fig:berlin1}\,\textbf{F}).

TOF measures an expected temporal distance of the kNN neighbors in reconstructed state-space (Eq.\,\ref{eq:TOF}), thus it has time dimension. A high or medium value of TOF implies that neighboring points in state-space were not close in time, therefore the investigated part of state-space was visited on several different occasions by the system. In our example, green diamonds on (Fig.\,\ref{fig:berlin1}\,\textbf{C}) mark states which were the closest points to the blue diamond in the state space, but were evenly distributed in time, on Fig.\,\ref{fig:berlin1}\,\textbf{A}. Thus the state marked by the blue diamond was not a unique state, the system returned there several times.

However a small value of TOF implies that neighboring points in state-space were also close in time, therefore this part of the space was visited only once by the system. On Fig.\,\ref{fig:berlin1}\,\textbf{C} and \textbf{D} orange diamonds mark the closest states to the red diamond and they are also close to the red diamond in time, on the (Fig.\,\ref{fig:berlin1}\,\textbf{B}). This results in low value of TOF in the state marked by the red diamond and means that it was a unique state never visited again.
Thus, small TOF values feature the uniqueness of sample points in state-space, and can be interpreted as an outlier factor. Correspondingly, TOF values exhibit a clear breakdown at time interval of the anomalous T wave (Fig.\,\ref{fig:berlin1}\,\textbf{F}).  

The number of neighbors ($k$) used during the estimation procedure sets the minimal possible TOF value:
\begin{equation}\label{eq:minlength}
    \mathrm{TOF}_\mathrm{min} = 
    \sqrt{ \frac{ \sum_{i=-\floor{k/2} }^{\floor{k/2} + k \bmod 2}{i^2}}{k} } \Delta t,
\end{equation}
Where $\lfloor k/2 \rfloor$ is the integer part of $k/2$, $\bmod$ is the modulo operator and $\Delta t$ is the sampling period.

The approximate maximal possible TOF value is determined by the length ($T$) and neighborhood size ($k$) of the embedded time series:
\begin{equation}\label{eq:maxlength}
    \mathrm{TOF}_\mathrm{max} = \sqrt{ \frac{\sum_{i=0}^{k-1} (T - i \Delta t) ^ 2 }{k} }
\end{equation}

TOF shows a time-dependent mean baseline and variance (Fig.\,\ref{fig:berlin1}\,\textbf{E}, Fig.\,S2) which can be computed if the time indices of the nearest points are evenly distributed along the whole time series.
The approximate mean baseline is a square-root-quadratic expression, it has the lowest value in the middle and highest value at the edges (see exact derivation for continuous time limit and $q=1$ in the Supporting Information, Fig.\,S2-S3):

\begin{equation}\label{eq:mu_TOF}
	\sqrt{\left< \mathrm{{TOF}_{\mathrm{noise}}} \left( t \right) ^{2} \right>} =  \sqrt{t^2 - t T + \frac{T^2}{3}}
\end{equation}

\begin{equation}\label{eq:var_TOF}
        \mathrm{VAR} \left( {\mathrm{TOF}^{2}_{\mathrm{noise}}} \left( t \right) \right) =
        \frac{1}{k} \left( \frac{t^5 + (T-t)^5}{5 T} 
        - \left( t^2 - tT + \frac{T^2}{3} \right)^2 \right)
\end{equation}

Based on the above considerations, imposing a threshold $\theta$ on $TOF_{k}$ has a straightforward meaning: it sets a maximum detectable event length ($M$) or vice versa: 
\begin{equation}
    \theta = \sqrt{ \frac{\sum_{i=0}^{k-1}{\left( M-i \Delta t \right) ^2} }{k} } \quad \bigg| \quad k \Delta t \stackrel{!}{\leq} M
\label{eq:threshold}
\end{equation}
Where in the continuous limit, the threshold and the event length becomes equivalent:
\begin{equation}
       \lim_{\Delta t \to 0}{\theta(M)} = M  
\end{equation}
Also, the parameter $k$ sets a necessary detection-criteria on the minimal length of the detectable events: only events with \soma{length $M\ge k \Delta t$} may be detected. 
This property comes from the requirement, that there must be at least k neighbors within the unique dynamic regime of the anomaly.

\soma{The current implementation of the TOF algorithm contains a time delay embedding, a $k$NN search, the computation of TOF scores from the neighborhoods and a threshold application for it.
The time-limiting step is the neighbor-search, which uses the scipy cKDTree implementation of the kDTree algorithm \cite{Bentley75}.
The most demanding task is to build the data-structure; its complexity is $O(k n \log{n})$ \cite{Brown15}, while the nearest neighbor search has $O(\log n)$ complexity.}

\begin{figure}[htb!]
\centering
\includegraphics[width=\textwidth]{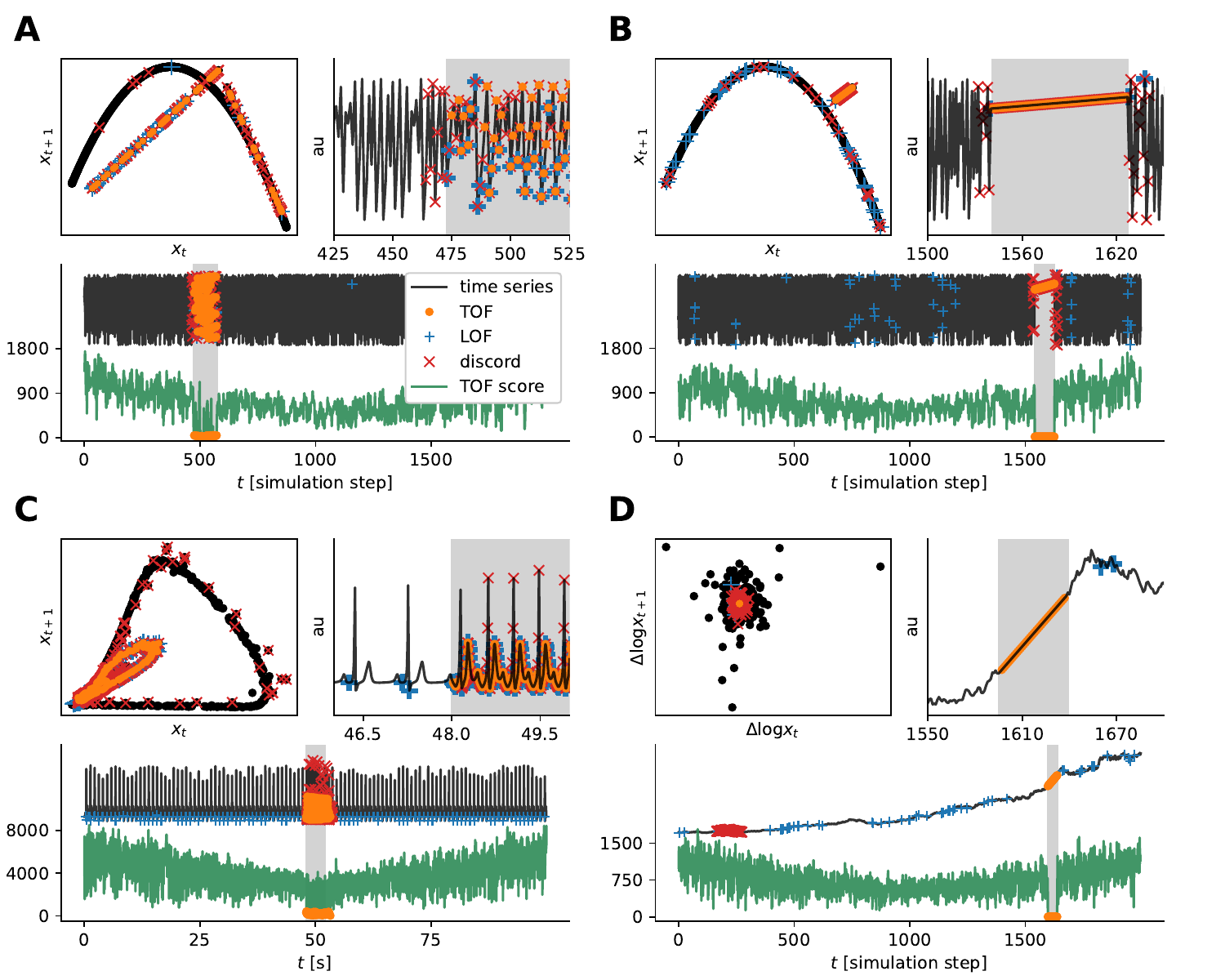}
\caption{\textbf{Simulated time series with anomalies of different kinds.}  (A) Logistic map time series with tent-map anomaly. (B) Logistic map time series with linear anomaly. (C) Simulated ECG time series with tachycardia. (D) Random walk time series with linear anomaly, where TOF was measured on the discrete time log derivative ($\Delta log x_t$). Each subplot shows an example time series of the simulations (black) in arbitrary units and in three forms: Top left the return map, which is the results of the 2D time delay embedding, and defines the dynamics of the system or its 2D projection. Bottom: Full length of the simulated time series (black) and the corresponding TOF values (green). Shaded areas show anomalous sections. Top right: Zoom to the onset of the anomaly. \soma{In all graphs, the outliers detected by TOF, LOF and Keogh's discord algorithms are marked by orange dots, blue plus and red x signs respectively. While anomalies form clear outliers on A and B, D shows an example where the unique event is clearly not an outlier, but it is located in the centre of the distribution. All the three algorithms detected the example anomaly well in case A, TOF and discord detected well the anomalies in B and C cases, but only TOF was able to detect all the four anomaly examples.}}
\label{fig:chart2}
\end{figure}

\soma{
\subsection*{Evaluation and comparison to previous methods}
We compare our method to widely used model-free, unsupervised outlier detection methods: the Local Outlier Factor (LOF) and two versions of discord detection \cite{Keogh05,Senin15} (see SI).
} The main purpose of the comparison is not to show that our method is superior to the others in outlier detection, but to present the fundamental differences between the \soma{previous} outlier concepts and the unicorns.

\soma{ The first steps of all three algorithms are parallel: While TOF and LOF use time-delay embedding as a preprocessing step to define a state-space, discord algorithm reaches the same by defining subsequences due to a sliding window. As a next step, state space distances are calculated in all of the three methods, but with slightly different focus. Both LOF and TOF search for the kNNs in the state-space for each time instance. As a key difference, the LOF calculates the distance of the actual points in state-space from their nearest neighbors and normalizes it with the mean distance of those nearest neighbors from their nearest neighbors, resulting in a relative local density measure. LOF values around 1 are considered the signs of normal behavior, while higher LOF values mark the outliers. While LOF concentrates on the densities of the nearest neighbours in the state-space, the discord concept is based on the distances directly. For each time instance, it searches for the closest, but temporary non-overlapping subsequence (state). This distance defines the distance of the actual state from the whole sequence and is called the matrix profile \cite{Yeh17}. Finally, the top discord is defined as the state, which is the most distant from the whole data sequence by this means. Besides this top-discord, any predefined number of discords can be defined by finding the next most distant subsequence which does not overlap with the already found discords.
The only parameter of discord detection is the expected length of the anomaly, which is given as the length of the subsequences used for the distance calculation. Senin et al.\cite{Senin15, jmotif} extended Keogh's method by calculating the matrix profile for different subsequence lengths, then normalizing the distances by the length of the subsequences and finally choosing the most distant subsequence according to the normalized distances. Through this method Senin's algorithm provides an estimation of the anomaly length as well.
Both Keogh's and Senin's algorithm can be implemented in a slower but exact way by calculating all the distances, or fastening them by using the Symbolic Aggregate approXimation (SAX) method. In our comparisons, Keogh's method was calculated exactly while SAX was used for Senin's algorithm only. 
}

\begin{figure}[htb!]
\centering
\includegraphics[width=\textwidth]{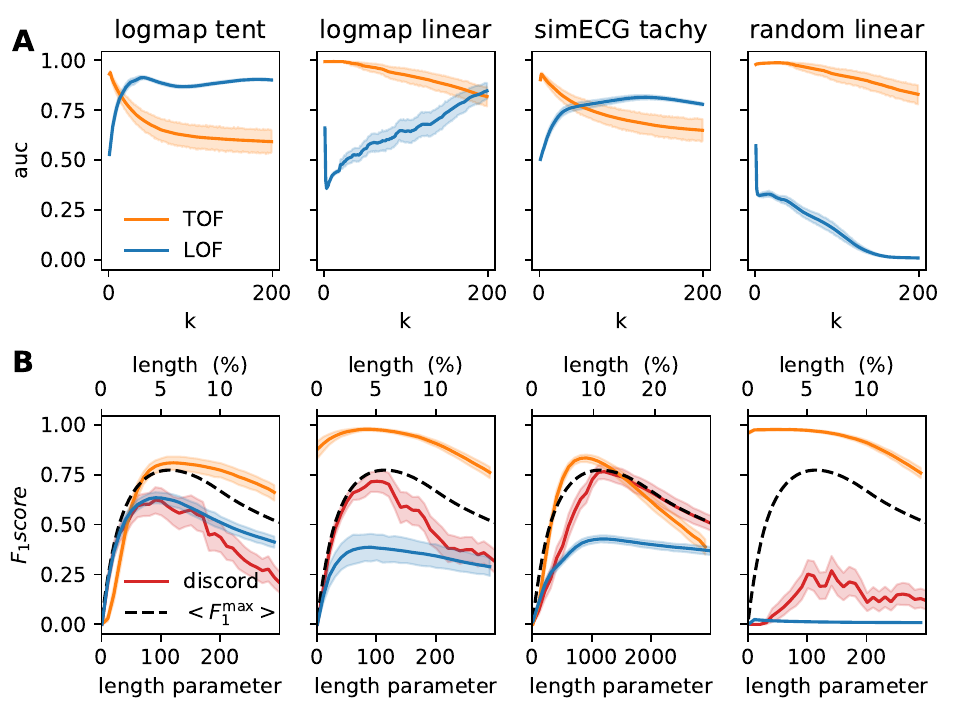}
\caption{\soma{\textbf{Performance evaluation of TOF, LOF and Keogh's discord algorithms on four simulated datasets.} (A) Mean Receiver Observer Characteristic Area Under Curve (ROC AUC) score and SD for TOF (orange) and LOF (blue) are showed as a function of neighborhood size ($k$). TOF showed the best results for small neighborhoods. In contrast, LOF showed better results for larger neighborhoods in the case of logistic map and ECG datasets, but did not reach reasonable performance on random walk with linear outliers.
(B) Mean $\mathrm{F}_1$ score for TOF (orange), LOF (blue) and Keogh's discord (red) algorithms as a function of the expected anomaly length (for TOF) given in either data percentage (for LOF) or window length parameter (for discord). Black dashed lines show the theoretical maximum of the mean $\mathrm{F}_1$ score for algorithms with prefixed detection numbers or lengths (LOF and discord), but this upper limit does apply for TOF. The $\mathrm{F}_1$ score of TOF was very high for the linear anomalies and slightly lower for logistic map - tent map anomaly and ECG datasets, but it was higher than the $\mathrm{F}_1$ score of the two other methods and their theoretical limits in all cases. Note, that the only comparable performance was shown by discord on ECG anomaly, while neither discord nor LOF was able to detect the linear anomaly on random background.}
}
\label{fig:performance}
\end{figure}

\begin{table}
    \caption{\textbf{Detection performance on simulations in terms of ROC AUC scores and the optimal neighborhood parameter $k$.} Maximal mean ROC AUC values and the corresponding SDs are shown. \soma{LOF was able to distinguish tent map and linear outliers from logistic background  and tachycardia from the normal rhythm with reasonable reliability but TOF outperformed LOF for all data series. Linear outliers can not be detected on random walk background by the LOF method at all, while TOF detected them almost perfectly. TOF reached its maximal performace mostly for low $k$ values, while LOF required larger $k$ for optimal performance on those three data series, on which it worked reasonably. While the ROC AUC was maximal at $k=30$ in case of random walk with linear outlier, the performance was not significantly lower for lower $k$ values.}}
    \centering
    \begin{tabular}{lcccccc}
\toprule\multicolumn{1}{c}{dataset} & $ $ & \multicolumn{2}{c}{TOF} &  $\,$ & \multicolumn{2}{c}{LOF} \\
{} & $ $ & $k$ &                AUC &  $ $ & $k$ &                AUC \\
\cmidrule{3-4} \cmidrule{6-7}logmap-tent    &   $\,$ &   2 &  $\mathbf{0.939 \pm 0.050}$ &  $\,$ &  42 &  $0.913 \pm 0.042$ \\
logmap-linear    &   $\,$ &   6 &  $\mathbf{0.994 \pm 0.007}$ &  $\,$ &  199 &  $0.847 \pm 0.213$ \\
sim ECG-tachy    &   $\,$ &   2 &  $\mathbf{0.931 \pm 0.039}$ &  $\,$ &  129 &  $0.815 \pm 0.056$ \\
randwalk-linear    &   $\,$ &   30 &  $\mathbf{0.988 \pm 0.014}$ &  $\,$ &  1 &  $0.572 \pm 0.015$ \\
\bottomrule
\end{tabular}
    \label{tab:auc}
\end{table}

\subsubsection*{Simulated data series}
\zsiga{We tested the TOF method on various types of simulated dataseries to demonstrate its wide applicability. These simulations are examples of deterministic discrete time systems, continuous dynamics and a stochastic process.}

We simulated two datasets with deterministic chaotic discrete-time dynamics generated by a logistic map \cite{May76} ($N=2000$, $100-100$ instances each) and inserted variable-length ($l=20-200$ step) outlier-segments into the time series at random times (Fig.\,\ref{fig:chart2}\,\textbf{A-B}).
Two types of outliers were used in these simulations, the first type was generated from a tent-map dynamics (Fig.\,\ref{fig:chart2}\,\textbf{A}) and the second type was simply a linear segment with low gradient (Fig.\,\ref{fig:chart2}\,\textbf{B}) for simulation details see the Supporting Information (SI).
The tent map demonstrates the case, where the underlying dynamics is changed for a short interval, but it generates a very similar periodic or chaotic oscillatory activity (depending on the parameters) to the original dynamics. 
This type of anomaly is hard to distinguish by naked eye. In contrast, a linear outlier is easy to identify for a human observer but not for many traditional outlier detecting algorithms. The linear segment is a collective outlier and all of its points represent a state that was visited only once during the whole data sequence, therefore they are unique events as well.

As a continuous deterministic dynamics with realistic features, we simulated electrocardiograms with short tachycardic periods where beating frequency was higher (Fig.\,\ref{fig:chart2}\,\textbf{C}).
The simulations were carried out according to the model of Rhyzhii \& Ryzhii \cite{Ryzhii14}, where the three heart pacemakers and muscle responses were modelled as a system of nonlinear differential equations (see SI).
We generated $100$ seconds of ECG and randomly inserted $2-20$ seconds long faster heart-rate segments, corresponding to tachycardia ($n=100$ realizations).

Takens' time delay embedding theorem is valid for time series generated by deterministic dynamical systems, but not for stochastic ones. In spite of this, we investigated the applicability of time delay embedded temporal and spatial outlier detection on stochastic signals with deterministic dynamics as outliers. 
We established a dataset of multiplicative random walks ($n=100$ instances, $T=2000$ steps each) with randomly inserted variable length linear outlier segments ($l=20-200$, see SI). 
As a preprocessing step, to make the random walk data series stationary, we took the log-difference of time series as is usually the case with economic data series. (Fig.\,\ref{fig:chart2}\,\textbf{D}).

\subsubsection*{Results on simulated data series}

\begin{table}[htb!]
    \caption{\textbf{Performance evaluation by $\mathrm{F}_1$, precision and recall scores on simulations.} \soma{The optimal expected anomaly length parameter (M) in time steps, mean scores and their standard deviations are shown for all methods and datasets; the highest scores are highlighted in bold. In case of TOF, $k=4$ neighbour number is used, while for LOF, the $k$ resulted the best ROC AUC were used from Table\,\ref{tab:auc}: $k=42$ for logmap-tentmap, $k=199$ for logmap-linear, $k=129$ for ECG tachycardia and $k=1$ for random walk-linear datasets. TOF resulted the highest $\mathrm{F}_1$ scores and highest precision for all datasets and the highest recall in three of the four cases but the simulated ECG tachycardia, where Keogh's discord algorithm reached slightly higher recall score. The only comparable performance was reached by Keogh's discord algorithm on ECG tachycardia in terms of $\mathrm{F}_1$ score while LOF produced reasonable results on logmap-tentmap anomaly series.
    Despite Senin's discord algorithm resulted in reasonable mean estimations for the lengths of the anomalies, its detection performance was worse than the other three algorithms.}}
    \centering
    \begin{tabular}{lcccc}
\toprule
method 	&	TOF 	    &	LOF	& 	Keogh	 &  Senin\\
\cmidrule{1-5}
dataset & \multicolumn{4}{c}{logistic map - tent map} \\
\cmidrule{1-5}
length (M)    &   $121$  &    $91$  &    $91$   &  $137.06 \pm 93.68$ \\
$\mathrm{F}_1$	& $\mathbf{0.810 \pm 0.175}$ 	& $0.635 \pm 0.141$ & $0.624 \pm 0.329$	&  $0.002 \pm 0.016$ \\
precision    & $\mathbf{0.920 \pm 0.139}$ 	& $0.702 \pm 0.231$ & $0.720 \pm 0.387$	&  $0.002 \pm 0.014$ \\
recall	& $\mathbf{0.734 \pm 0.185}$ 	& $0.659 \pm 0.149$ & $0.586 \pm 0.337$	&  $0.003 \pm 0.019$ \\
\cmidrule{1-5}
dataset & \multicolumn{4}{c}{logistic map - linear} \\
\cmidrule{1-5}
length (M)    &   $81$  &    $91$  &    $101$   &  $146.56 \pm 91.17$ \\
$\mathrm{F}_1$	& $\mathbf{0.978 \pm 0.038}$ 	& $0.387 \pm 0.353$ & $0.717 \pm 0.273$	&  $0.267 \pm 0.358$ \\
precision    & $\mathbf{0.978 \pm 0.053}$ 	& $0.382 \pm 0.366$ & $0.766 \pm 0.332$	&  $0.220 \pm 0.308$ \\
recall	& $\mathbf{0.981 \pm 0.038}$ 	& $0.459 \pm 0.428$ & $0.752 \pm 0.289$	&  $0.370 \pm 0.473$ \\
\cmidrule{1-5}
dataset & \multicolumn{4}{c}{sim ECG - tachycardia} \\
\cmidrule{1-5}
length (M)    &   $910$  &    $1110$  &    $1210$   &  $1128.04 \pm 1024.98$ \\
$\mathrm{F}_1$	& $\mathbf{0.834 \pm 0.094}$ 	& $0.428 \pm 0.092$ & $0.765 \pm 0.177$	&  $0.368 \pm 0.381$ \\
precision    & $\mathbf{0.861 \pm 0.115}$ 	& $0.425 \pm 0.119$ & $0.751 \pm 0.267$	&  $0.305 \pm 0.344$ \\
recall	& $0.815 \pm 0.091$ 	& $0.498 \pm 0.144$ & $\mathbf{0.894 \pm 0.141}$	&  $0.548 \pm 0.498$ \\
\cmidrule{1-5}
dataset & \multicolumn{4}{c}{random walk - linear} \\
\cmidrule{1-5}
length (M)    &   $51$  &    $11$  &    $141$   &  $161.01 \pm 80.38$ \\
$\mathrm{F}_1$	& $\mathbf{0.977 \pm 0.018}$ 	& $0.024 \pm 0.024$ & $0.269 \pm 0.393$	&  $0.007 \pm 0.034$ \\
precision    & $\mathbf{0.999 \pm 0.004}$ 	& $0.127 \pm 0.092$ & $0.284 \pm 0.425$	&  $0.006 \pm 0.030$ \\
recall	& $\mathbf{0.956 \pm 0.033}$ 	& $0.014 \pm 0.015$ & $0.266 \pm 0.387$	&  $0.015 \pm 0.104$ \\
\bottomrule
\end{tabular}
    \label{tab:F1}
\end{table}

\soma{TOF and LOF calculates scores on which thresholds should be applied to reach final detections. In contrast, the discord algorithms do not apply a threshold on the matrix profile values, but choose the highest peak as a top discord. The effectiveness of TOF and LOF scores to distinguish anomalous points from the background can be evaluated by measuring the area under receiver operator characteristic curve (ROC AUC, see Methods\,\ref{Metrics}). This evaluation method considers all the possible thresholds, thus provides a threshold-independent measure of the detection potential for a score, where 1 means that a threshold can separate all the anomalous points from the background. Thus, we applied ROC AUC to evaluate TOF and LOF scores on the four datasets mentioned above with fixed embedding parameters $E=3$ and $\tau=1$ and determined its dependency on the neighborhood size ($k=1-200$) that was used for the calculations.

Fig.\,\ref{fig:performance}\,\textbf{A} shows the performance of the two methods in terms of mean ROC AUC and SD for $n=100$ realizations. TOF produced higher maximal ROC AUC than LOF in all the four experimental setups. The ROC AUC values reached their maxima at small $k$ neighbourhood sizes in all of the four cases, and decreased with increasing $k$ afterwards. In contrast, LOF resulted in reasonable ROC AUC values in only three cases (logmap-tent anomaly, logmap-linear anomaly and ECG tachycardia), and it was not able to distinguish the linear anomaly from the random walk background at all. The ROC AUC values reached their maxima at typically higher $k$ neighbourhood size in the instances where LOF worked (Table \ref{tab:auc}).

In order to evaluate the final detection performance, as well as the type of errors made and the parameter dependency of these algorithms, $\mathrm{F}_1$ score, precision and recall were computed for all the four algorithms. $\mathrm{F}_1$ score is especially useful to evaluate detection performance in cases of highly unbalanced datasets as in our case, see Methods \ref{Metrics}}.

As TOF showed best performance in terms of ROC AUC with lower $k$ neighborhood sizes, the $\mathrm{F}_1$ scores were calculated at a fixed $k=4$ neighborhood forming a simplex in the 3 dimensional embedding space \cite{Sugihara12}. In contrast, as LOF showed stronger dependency on neighborhood size, the optimal neighborhood sizes were used for $\mathrm{F}_1$ score calculations. \soma{Discord uses no neighbourhood parameter, as it calculates all-to-all distances between points in the state space.}

\soma{Three among the four investigated algorithms require an estimation of the expected length of the anomaly, however this estimation become effective through different parameters within the different algorithms.
In case of LOF, the expected length of the anomaly can be translated into a threshold, which determines the number of time instances above the threshold. In the absence of this information, the threshold is hard to determine in any principled way.
In case of the Keogh's discord detection algorithm the length of the anomaly is the only parameter and no further threshold is required. Both LOF and Keogh's discord find the predefined number of time instances exactly. While the discord finds them in one continuous time interval, LOF detects independent points along the whole data.
The expected maximal anomaly length is necessary to determine the threshold in case of TOF as well (Eq.\,\ref{eq:threshold}).
As Senin's discord algorithm does not require predefined anomaly length, it was omitted from this test, and we calculated the F$_1$ score at the self-determined window length.

Fig.\,\ref{fig:performance}\,\textbf{B} shows the mean $\mathrm{F}_1$ scores for n=100 realizations, as a function of the the expected anomaly length, for the three algorithms and for all the four test datasets. Additionally, Fig.\,S8 shows the precision and the recall, which are the two constituents of the $\mathrm{F}_1$ score as a function of the expected anomaly length as well. The actual length of the anomalies were randomly chosen between 20 and 200 time steps for each realization in three of our four test cases and between 200 and 2000 time steps in ECG realizations, thus the effect of the expected length parameters were examined up to these lengths as well.

While it is realistic, that we only have a rough estimate on the expected length of the anomaly, it turns out, that the randomness in the anomaly length sets an upper bound (Fig.\,\ref{fig:performance}\,\textbf{B}, black dashed lines, Fig.\,S6), for the mean $\mathrm{F}_1$ scores for those algorithms, that work with exact predefined number of detections i.e. the LOF and the Keogh's discord. Although the expected length parameter and the randomness in the actual anomaly length affect the detection performance of TOF as well, they do not set a strict upper bound, as the number of detections is not in a one-to-one correspondence with the expected anomaly length.

For all the four test datasets, TOF algorithm reached higher maximal $\mathrm{F}_1$ scores than the LOF and Keogh's discord method (Fig.\,\ref{fig:performance}\,\textbf{B}, Fig.\,S8, orange lines). The maximal $\mathrm{F}_1$ score was even higher than the theoretical limit imposed by the variable anomaly lengths to the other methods. Similar to the results on ROC AUC values, performance of TOF algorithm was excellent on the the linear type anomalies and very good for the logmap-tentmap and the simulated ECG-tahycardia datasets.

In contrast, LOF algorithm showed good performance on the logmap-tentmap data series and mediocre results on logmap-linear anomalies and on the ECG-tachycardia dataseries. The linear outlier on random walk background was completely undetectable for the LOF method (Fig.\,\ref{fig:performance}\,\textbf{B}, Fig.\,S8, blue lines).

Keogh's discord algorithm displayed good $\mathrm{F}_1$ scores on three datasets, but weak results were given in case of the linear anomaly on the random walk background (Fig.\,\ref{fig:performance}\,\textbf{B}, Fig.\,S8, red lines). 

The simulated ECG dataset was the only one, where any of the competitor methods showed comparable performance to TOF: Keogh's discord reached its theoretical maximum, thus TOF resulted in an only slightly higher maximal $\mathrm{F}_1$ score in an optimal range of the length parameter. If the expectation significantly overestimated the actual length, the results of discord were slightly better.

The $\mathrm{F}_1$ scores reached their maxima when the expected anomaly length parameters were close to the mean of the actual anomaly lengths for all algorithms and for all detectable cases when the $\mathrm{F}_1$ score showed significant peaks (Table \ref{tab:F1}).
}

\soma{As we have seen, the variable and unknown length of the anomalies had significant effect on the detection performance of all methods, but especially LOF and discord. Senin et al. \cite{Senin14, Senin15} extended the discord detection method to overcome the problem of predefined anomaly length and to allow the algorithm to find the length of the anomalies. Thus, we have tested Senin's algorithm on our test data series and included the anomaly lengths found by this algorithm as well as the performance measures into the comparison in Table\,\ref{tab:F1}. While the mean estimated anomaly lengths were not far from the mean of the actual lengths, the performance of this algorithm lags well behind all three previously tested ones on all the four types of test data series.
}

\soma{We have identified several factors, which could explain the different detection patterns of different algorithms.} Table\,S1 shows, that the tent map and the tachycardia produce lower density, thus more dispersed points in the state space, presumably making them more detectable by the LOF. In contrast, linear segments resulted in similar density of points to the normal logistic activity or higher density of points compared to the random walk background. Detrending via differentiation of the logarithm was applied as a preprocessing step in the latter case, making the data series stationary and drastically increasing the state space density of the anomaly. 
\soma{LOF relays solely on the local density, thus it only counted the low density sets as outliers. In contrast, as discord method identifies anomalies based on the distances in the state space, it was able to detect linear anomaly on chaotic background, tent-map anomaly on log-map dataseries and tachycardia on the simulated ECG data, but failed on the detection of the linear anomaly on random walk background. The state space points belonging to the well detected anomalies are truly farther from the points in the manifolds of the background dynamics (Fig.\,\ref{fig:berlin1}\,\textbf{A}-\textbf{C}). In contrast, after discrete time derivation, the points belonging to a linear anomaly are placed near the center of the background distribution (Fig.\,\ref{fig:berlin1}\,\textbf{D}), making them undetectable either for LOF and discord algorithms.

The detection performance of TOF was less affected by the relation between the expected and the actual length of the anomalies in the linear cases. The reason behind this is that each point of the linear segment is a unique state in itself, thus it always falls below the expected maximal anomaly length. In contrast, the tent map and tachycardic anomalies produce short, but stationary segments, which can be less effectively detected if they are longer than the preset expected length.

We can conclude, that 1) TOF has reached better performance to detect anomalies in all the investigated cases, 2) there were special types of anomalies which can be detected only by TOF and can be considered unicorns but not outliers or discords.}

\subsection*{TOF detects unicorns}

\begin{figure}[htb!]
\centering
\includegraphics[width=8.7cm]{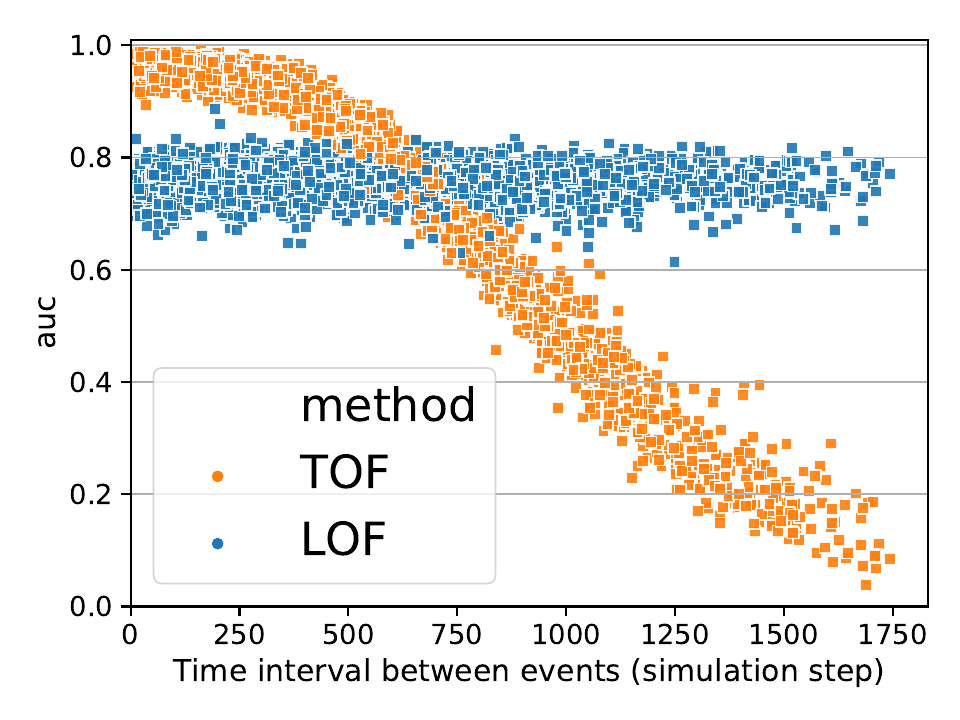}
\caption{\textbf{TOF detects unique events.} Detection performance measured by ROC AUC as a function of the minimum inter event interval (IEI) between two inserted tent-map outlier segments.
TOF was able to distinguish outliers from the background very well when IEIs were below 300 steps, and the two events can be considered one. However, the detection performance of TOF decreased for higher IEIs.
In contrast, LOF's peak performance was lower, but independent of the IEI.}
\label{fig:chart4}
\end{figure}

To show that TOF enables detection of only unique events, additional simulations were carried out, where two, instead of one, tent-map outlier segments were inserted into the logistic map simulations.
We detected outliers by TOF and LOF and subsequently ROC AUC values were analysed as a function of the inter event interval (IEI, Fig.\,\ref{fig:chart4}) of the outlier segments.
LOF performed independent of IEI, but TOF's performance showed strong IEI-dependence.
Highest TOF ROC AUC values were found at small IEI-s and AUC was decreasing with higher IEI.
Also the variance of ROC AUC values was increasing with IEI. This result showed, that TOF algorithm can detect only unique events: if two outlier events are close enough to each other, they can be considered as one unique event together. In this case, TOF can detect it with higher precision, compared to LOF. However if they are farther away than the time limit determined by the detection threshold, then the detection performance decreases rapidly.

The results also showed, that anomalies can be found by TOF only if they are alone, a second appearance decreases the detection rate significantly.

\subsection*{Application examples on real-world data series}

\begin{figure}[htb!]
\centering
\includegraphics[scale=0.7]{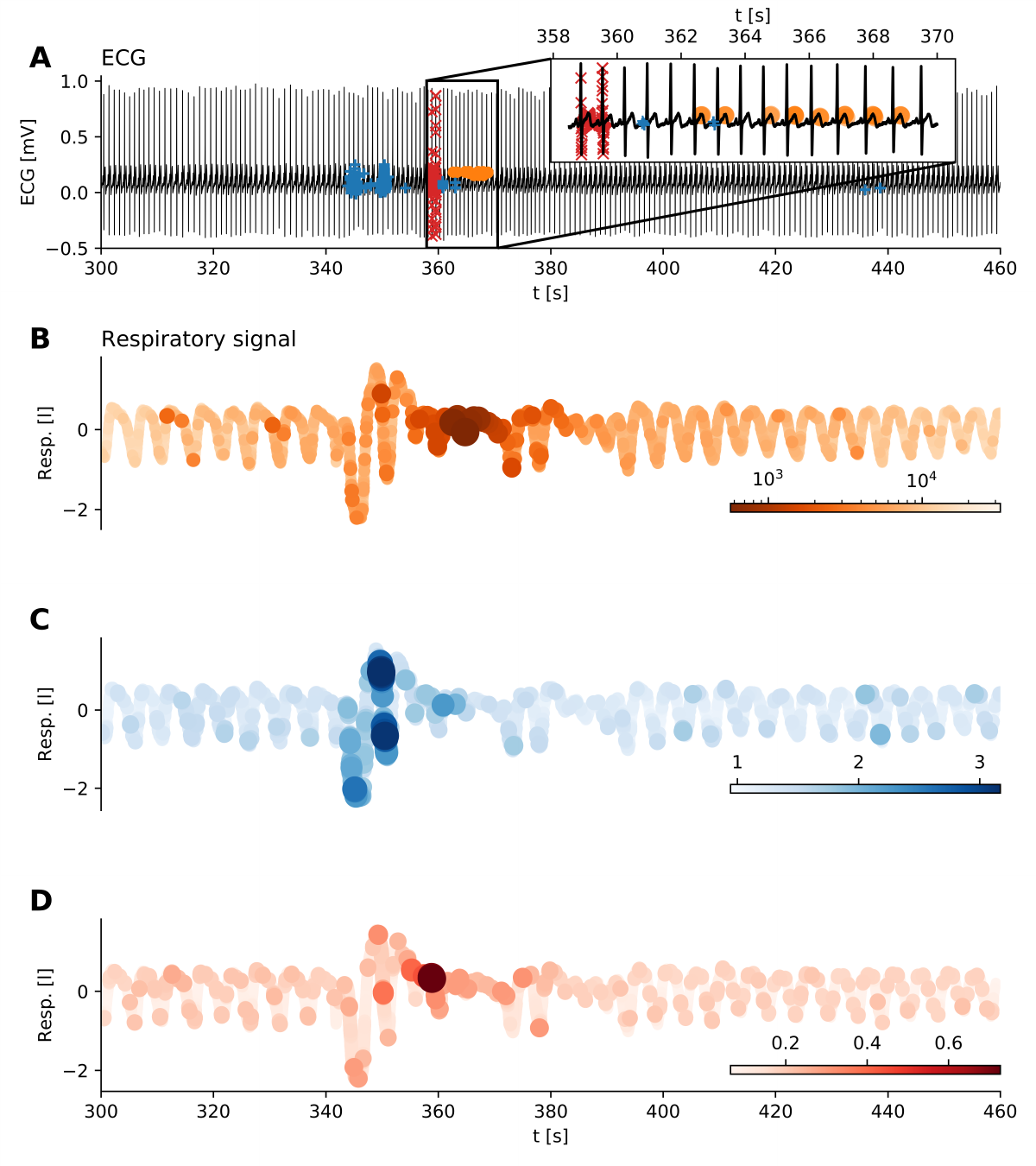}
\caption{\textbf{Detecting apnea with arousal on ECG.} (A) ECG time series with unique events detected by TOF (orange dots, $E=3, \tau=0.02 \mathrm{\,s}, k=11, M=5 \mathrm{\,s}$), outliers detected by LOF (blue + signs, $E=7,\tau=0.02 \mathrm{\,s}, k=100$, threshold$=0.5 \% $) and the top discord (red x signs, M=5\,s). The inset shows the more detailed pattern of detections: unique behavior mainly appears on the T waves. \soma{(B-D) Breathing air-flow time series parallel to the above ECG recording, colored according to the scores of the three anomaly methods. The anomaly starts with a period of irregular breathing at 340\,s, followed by the apnea when breathing almost stops (350-370\,s). After this anomalous period, an arousal restores the normal breathing. (B) Air flow colored according to the TOF score at each sample. Low values (darker colors) mark the anomaly corresponding to the period of apnea. (C) Air-flow time series with coloring corresponds to the LOF score at each sample. Higher LOF values mark the outliers. LOF finds irregular breathing preceding the apnea. (D) Air flow time series colored according to the matrix profile values by the discord algorithm. Discord finds the point of transition from irregular breathing to the apnea.}}
\label{fig:ecg}
\end{figure}

\subsubsection*{Detecting apnea event on ECG time series}

\soma{To demonstrate, that the TOF method can reveal unicorns in real world data, we have chosen data series where the existence and the position of the unique event already known.}

We applied TOF to ECG measurements from the MIT-BIH Polysomnographic Database’s \cite{Ichimaru99, Goldberger00}  to detect apnea event. Multichannel recordings were taken on $250$ Hz sampling frequency, and the ECG and respiratory signal of the first recording was selected for further analysis ($n=40000$ data points $1600$ seconds).

While the respiratory signal clearly showed the apnea, there were no observable changes on the parallel ECG signal.

We applied time delay embedding with $E_\mathrm{TOF}=3$, $E_\mathrm{LOF}=7$ and $\tau=0.02\mathrm{\,s}$ according to the first zerocrossing of the autocorrelation function (Fig.\,S9). 
TOF successfully detected apnea events in ECG time series; interestingly, the unique behaviour was found mostly during T waves when the breathing activity was almost shut down (Fig.\,\ref{fig:ecg}, $k=11$, $M=5\mathrm{\,s}$).
In contrast, LOF was sensitive to the increased and irregular breathing before apnea ($k=200$, threshold$=0.5\mathrm{\,\%}$), \soma{while the top discord ($M=5\mathrm{\,s}$) were found at the transient between the irregular breathing and the apnea.}
This example shows that our new method could be useful for biomedical signal processing and sensor data analysis.

\begin{figure}[tb!]
\centering
\includegraphics[width=\textwidth]{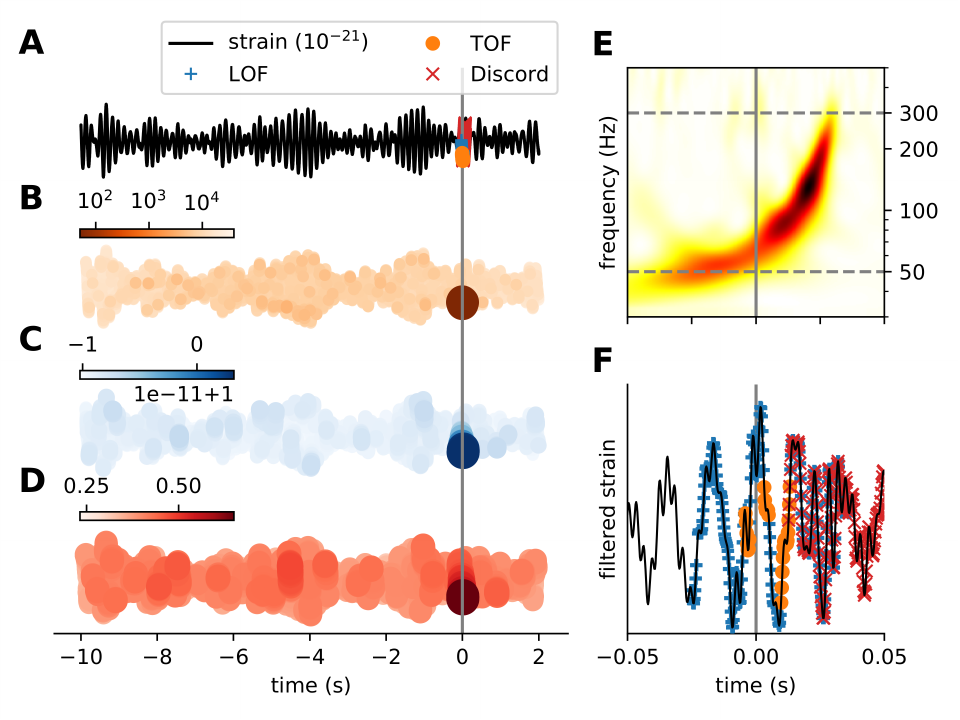}
\caption{\textbf{Detection of the GW150914 event on LIGO open data with TOF and LOF and discord.}
(A) Strain time series (black) from Hanford detector around GW150914 event (grey vertical line) with \soma{TOF (orange dots), LOF (blue plus) and discord (red x) detections.
TOF score values (B), LOF scores (C) and matrix profile scores (D) are mapped to the time series (orange, blue and red colors respectively), the strongest colors show the detected event around $0$ seconds.} 
(E) The Q-transform of the event shows a rapidly increasing frequency bump in the power spectra right before the merger event (grey). The grey horizontal dashed lines show the lower ($50$ Hz) and upper ($300$ Hz) cutoff frequencies of the bandpass filter, which was applied on the time series as a preprocessing step before anomaly detection.
(F) Filtered strain data at $0.1$ second neighborhood around the event. TOF, LOF and discord detected the merger event with different sensitivity. LOF detected more points of the event, while TOF found the period which has the highest power in the power spectra and discord detected the end of the event. 
($E_\mathrm{TOF}=6$, $\tau_\mathrm{TOF}=1.953$\,ms, $k_\mathrm{TOF}=12$, $M_{\mathrm{TOF}}=146.484$\,ms, $w=7$; $E_\mathrm{LOF}=11$, $\tau_\mathrm{LOF}=1.953$\,ms, $k_\mathrm{LOF}=100$, threshold$=0.5 $\%, \soma{$M_{discord}=146.484\,ms$})
}
\label{fig:grav}
\end{figure}

\subsubsection*{Detecting gravitational waves}
As a second example of real world datasets \soma{with known unique event}, we analyzed gravitational wave detector time series around the GW150914 merger event \cite{Abbott16} (Fig.\,\ref{fig:grav}).
The LIGO Hanford detector's signal ($4096$ Hz) was downloaded from the GWOSC database \cite{Abbott19}.
A 12\,s long segment of strain data around the GW150914 merger event was selected for further analysis.
As a preprocessing step, the signal was bandpass-filtered (50-300 Hz).
Time delay embedding was carried out with embedding delay of $8$ time-steps ($1.953$ ms) and embedding dimension of $E=6$ and $E=11$ for TOF and LOF respectively.
\soma{The neighbour parameter was set to $k=12$, for TOF and $k=100$ for LOF. The length of the event was set to $M=146.484\,ms$ for TOF and discord and correspondingly, the threshold to $0.5 \%$ for LOF (Fig.\,S10).

All three algorithms detected the merger event, albeit with some differences. LOF found the whole period, while TOF selectively detected the period when the chirp of the spiraling black holes was the loudest. Interestingly, discord found the end of the event (Fig.\,\ref{fig:grav}\,\textbf{B}, \textbf{C}, \textbf{D}).}

\zsiga{
To investigate the performance of TOF on detecting noise bursts called blip in LIGO detector data series, we applied the algorithm on the Gravity Spy \cite{Zevin_2017} blip data series downloaded from the GWOSC database \cite{Abbott19} (Fig.\,S7).
We determined the value of optimal threshold on the training set ($N=128$), then measured precision, F$_1$ score, recall and block-recall metrics on the test set ($N=29$).
We set the threshold value by the maximum precision ($M=36$, Fig.\,S7\,A).
TOF reached high precision ($1$), low F$_1$ score, low recall and high block-recall ($0.9$) values (Fig.\,S7\,B) on the test set.
The high precision shows, that the detected anomaly is likely to be a real blip and the high block recall (hit rate) implies that TOF found blips in the majority of the sample time series.
}

\begin{figure}[tb!]
\centering
\includegraphics[width=\textwidth]{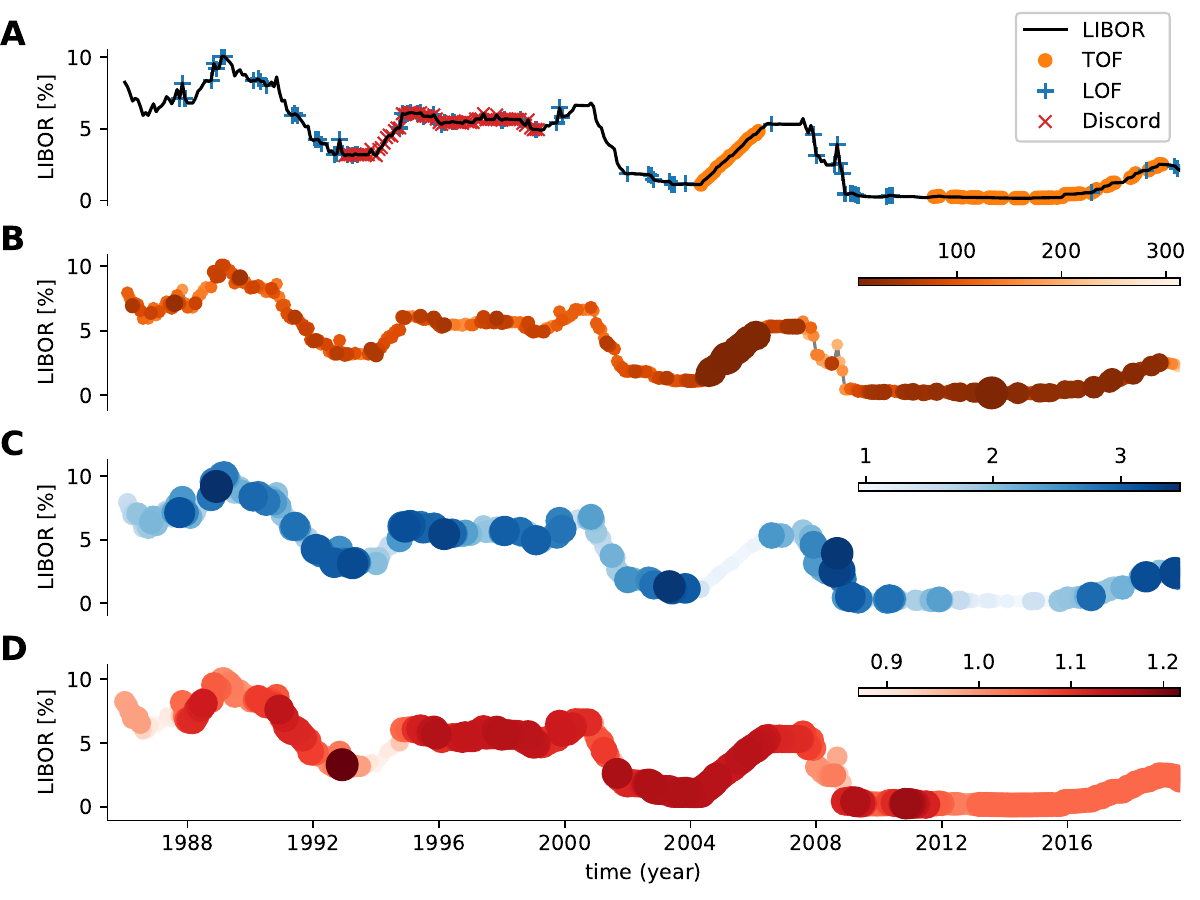}
\caption{
    \textbf{Analysis of LIBOR dataset.}  The detections were run on the temporal derivative of the LIBOR time series. (A) time series with detections. (B) TOF score values. (C) LOF score values. \soma{(D) Matrix profile scores by the discord algorithm.}
    TOF detected two rising periods: the first between 2005 and 2007 and a second, started in 2012 and lasts until now. While both periods exhibit unique dynamics, they differ from each other as well.
}
\label{fig:libor}
\end{figure}

\subsubsection*{London InterBank Offer Rate dataset}
\soma{Our final real world example is the application of TOF, LOF and discord algorithms on the London InterBank Offer Rate (LIBOR) dataset. In this case, we have no exact apriory knowledge about the appearance of unique events, but we assumed, that unique states found by TOF algorithm may have unique economical characteristics.} 

As a preprocessing step, discrete time derivative was calculated to eliminate global trends, then we applied TOF ($E=3, \tau=1, k=5, M=30$ month) and LOF ($E=3, \tau=1, k=30$, threshold$=18.86\,\%$) on the derivative (Fig.\,S11-S12).
TOF found the uprising period prior to the 2008 crisis and the slowly rising period from 2012 onwards as outlier segments.
LOF detected several points, but no informative pattern emerged from the detections (Fig.\,\ref{fig:libor}). \soma{Also, Discord detected a period between 1993 and 1999, with no obvious characteristic.

While in this case the ground-truth was not known, the two periods highlighted by TOF show specific patterns of monotonous growth.
Moreover, the fact that both of the two periods were detected by TOF shows that both dynamics are unique, therefore different from each other.}

\section*{Discussion}
In this paper we introduced a new concept of anomalous event called unicorn; unicorns are the unique states of the system, which were visited only once.
A new anomaly concept can be valid only if a proper detection algorithm is provided: we have defined the Temporal Outlier Factor to quantify the uniqueness of a state.
We demonstrated that TOF is a model-free, non-parametric, domain independent anomaly detection tool, which can detect unicorns.

TOF measures the temporal dispersion of state space neighbors for each point.
If state space neighbors are temporal neighbors as well, then the system has never returned to that state, therefore it is a unique event. ie. a unicorn.  

The unicorns are not just outliers in the usual sense, they are conceptually different.
As an example of their inherently different behavior, one can consider a simple linear data series: All of the points of this series are unique events; they are only visited once and the system never returned to either one of them. Whilst this property may seem counter-intuitive, it ensures that our algorithm finds unique events regardless of their other properties, such as amplitude or frequency. This example also shows, that the occurrences of unique events are not necessarily rare: actually, all the points of a time series can be unique. This property clearly differs from other anomaly concepts: most of them assume that there is a normal background behavior which generates the majority of the measurements and outliers form only a small minority.

\soma{Keogh’s discord detection algorithm \cite{Keogh05} differs from our method in an important aspect: Keogh’s algorithm finds one, or other predefined number of anomalies on any dataset. Thus Keogh’s algorithm can not be used to distinguish, whether there are any anomalies on the data or not, it will always find at least one. This property makes it inappropriate in many real world applications, since usually we do not know if there are any anomalies on the actual dataset or not. In contrast, our algorithm can return any number of anomalies, including zero.}

Detection performance comparison of TOF, LOF and discord on different simulated datasets highlighted the conceptual difference between the traditional outliers and the unique events as well. As our simulations showed, TOF with the same parameter settings was able to find both higher and lower density anomalies, based on the sole property that they were unique events. The algorithm has very low false detection rate, but not all the outlier points were found or not all the points of the event were unique. As an example, QRS waves of ECG simulations do not appear to be different from normal waves, hence the algorithms did not find them.

\soma{Of course our aim was not to compete with those specific algorithms that have been developed to detect sleep apnea events from ECG signal \cite{Sharma16}. Most of the methods extract and classify specific features of the R-R interval series called heart rate variability (HRV). It was shown, that sympathetic activation during apnea episodes leaves its mark on HRV \cite{Penzel03}, its spectral components, sample entropy \cite{AlAngari07} or correlation dimension \cite{Bock98}. Song et al.\cite{Song16} used discriminative Markov-chain models to classify HRV signals and reached $97\%$ precision for per-recording classification.}

While ECG analysis mostly concentrates on the temporal relations of the identified wave components, here we apply the detection methods to the continuous ECG data. \soma{Previously, it was shown, that apnea is associated with morphological changes of the P waves and the QRS complex in the ECG signal \cite{Penzel02,Boudaoud07, Sharma16}.}
Interestingly, TOF marked mainly the T waves of the heart cycle as anomalous points.
T waves are signs of the ventricular repolarization and are known to be largely variable, thus they are often omitted from the ECG analysis. This example showed, that they can carry relevant information as well.

\soma{The already identified gravitational wave GW150914 event was used to demonstrate the ability of our method to find another type of anomaly without prior knowledge about it.}

\soma{Clearly, specific model-based algorithms (such as matched filter methods \cite{Abbott16c}) or unmodelled algorithms that were originally used to recognize gravitational waves, such as coherent Wavebursts, omicron-LALInference-Bursts and BayesWave are much more sensitive to the actual waveforms generated by merger of black holes or neutron stars than our TOF method \cite{Abbott16b}. The unmodelled methods have only two basic assumptions: first, that gravitational wave-background (unlike ECG signal) is basically silent, thus detectors measure only Gaussian noise in the lack of an event. Thus, any increase in the observed wave-power needs to be detected and classified. Second, an increase in the coherent power between the far located detectors is the hallmark of candidate events of astrophysical origin. The detectors should observe similar waveforms with phase difference corresponding to the waves traveling with light-speed between them. In contrast, increased power in only one of the detectors should have terrestrial origin and these are called glitches. After the unmodelled detection of candidate waveforms, more specific knowledge about the possible waveforms can be incorporated into the analysis pipeline, such as analyzing time evolution of the central frequency of the signal, or comparison of the waveform to the model database, containing simulated waveforms generated by merger events.
Model-free methods can detect events with unpredicted waveforms may help to find glitches. The presence of different types of glitches significantly increases the noise level and decreases the useful data length of detectors, thus limits its sensitivity.}

\soma{In contrast to apnea and gravitational wave detection, the nature of anomalies are much less known in the economical context. Most of the anomaly detection methods concentrate on fraud detection on transaction or network traffic records and utilize clustering techniques to distinguish normal a fraudulent behaviours \cite{Ahmed16}.}

Whilst LOF showed no specific detection pattern, TOF detected two rising periods on the temporal derivate of the USD LIBOR dataset: one preceding the 2008 crisis and an other one from 2012 onwards.
Both detected periods showed unique dynamics: \soma{the large fluctuations are replaced by constant rising during these periods, the dynamics are 'frozen'. Note, that the rising speeds differ in the two periods.}
The period between 2005-2007 can be considered unique in many ways; not only was there an upswing of the global market, but investigations revealed that several banks colluded in manipulation and rigging of LIBOR rates in what came to be known as the infamous LIBOR scandal \cite{DeptJustice12}. Note, that this was not the only case, when LIBOR was manipulated: During the economic breakdown in 2008 the Barclys Bank submitted artificially low rates to show healthier appearance \cite{SniderYoule09, SniderYoule10, SniderYoule12}. As a consequence of these scandals, significant reorganization took place in controlling LIBOR calculation, starting from 2012.

To sum it up, gravitational waves of the merger black-holes on the filtered dataset formed a traditional outlier which was well detectable by all the TOF, the LOF and the discord algorithms, while LIBOR exhibited longer periods of unique events only detectable by TOF. 
Apnea generated a mixed event on ECG; the period of irregular breathing formed outliers detectable by LOF, while the period of failed respiration generated a unique event detectable only by the TOF. Meanwhile discord detected the transitory period between the two periods.

Comparing TOF, LOF and discord proved that temporal scoring has advantageous properties and adds a new aspect to anomaly detection.
One advantage of TOF can be experienced when it comes to threshold selection.
Since TOF score has time dimension, an actual threshold value means the maximal expected length of the event to be found.
Also, on the flipside the neighborhood size $k$ parameter sets the minimal event length. 
Because of these properties, domain knowledge about possible event lengths renders threshold selection to a simple task.

\soma{While TOF and LOF have similar computational complexity ($O(k n \log(n))$), the smaller embedding dimensions and neighborhood sizes, makes TOF computations faster and less memory hungry. In contrast, the exact discord algorithm has $O(k n^2 \log{n})$ complexity \cite{Keogh05}. While the running time of discord has been significantly fastened by the SAX approximation, our results may indicate, that the SAX approximation has limited the precision of Senin's algorithm seriously.

To measure the running time empirically, we applied TOF algorithm on random noise from $10^2-10^6$ sample size, $15$ instances each ($d=3$, $\tau=1$, $k=4$).
The runtime on the longest tested $10^6$ points long dataset was $15,144\pm0.351$ secs (Fig\,S4) on a laptop powered by Intel\textregistered Core\texttrademark i5-8265U. The fitted exponent of the scaling was 1.3. Based on these results, we have estimated that if memory issues could be solved, running a unicorn search on the whole 3 months length of the LIGO O1 data downsampled to 4096Hz would take 124 days on a single CPU (8 threads). A search through one week of ECG data would take 3 hours. As calculations on the ECG data are much shorter than the recording length; online processing is feasible as well.}

Time indices of k nearest neighbors have been previously utilized differently in nonlinear time series analysis to diagnose nonstationary time series \cite{Kennel97, Rieke02, Rieke04}, measure intrinsic dimensionality of system's attractors \cite{Gao99, Carletti06, Marwan07}, monitor changes in dynamics \cite{Gao13} and even for fault detection \cite{Martinez-Rego16}.
Rieke et al.\,\cite{Rieke02, Rieke04} utilized  very resembling statistics to TOF: the average absolute temporal distances of k nearest neighbors from the points. However they analyzed the distribution of temporal distances to determine nonsationarity and did not interpret the resulting distance scores locally.
Gao \& Hu and Martinez-Rego et al. \cite{Martinez-Rego16} used recurrence times to monitor dynamical changes in time series locally, but these statistics are not specialized for detecting extremely rare unique events.
TOF utilizes the temporal distance of k nearest neighbors at each point, thus provides a locally interpretable outlier score, which takes small values when the system visits an undiscovered territory of state-space for a short time period.

\soma{ The minimal detectable event length might be the strongest limitation of TOF method. We have shown, that the TOF method has a lower bound on the detectable event length ($\Theta_{min}$), which depends on the number of neighbors ($k$) used in the TOF calculations. This means that TOF is not well suited to detect point-outliers, which are easily detectable by many traditional outlier detection methods.

Furthermore, the shorter the analyzed time series and the smaller $k$ is used, the higher the chance, that the background random or chaotic dynamics spontaneously produce a unique event. Smaller $k$ results in higher fluctuations of the baseline TOF values, which makes the algorithm prone to produce false positive detections.

A further limitation arises from the difficulty of finding optimal parameters for the time delay embedding: the time delay $\tau$ and the embedding dimension $E$. Fig.\,S5 shows the sensitivity of the $F_1$ score to the time delay embedding parameters and the relation between the used and the optimal parameter pairs. This post hoc evaluation, which can be done for simulations but not in a real life data showed, that our general parameter setting ($E=3$, $\tau=1$) used during the tests was suboptimal for simulated ECG-tachycardia dataset. The optimal parameter settings ($E=7$, $\tau=6$) would have resulted in $0.94$ as the maximal $F_1$ score in stead of $0.83$, shown in Table\,\ref{tab:F1}).}

\soma{The model-free nature of these algorithms can be an advantage and a limitation at same time. The specific detection algorithms, which are designed on purpose and use specific a priory knowledge about the target pattern to be detected, can be much more effective than a model-free algorithm. Model-free methods are preferred when the nature of the anomaly is unknown. Consequently, detecting a unicorn tells us that the detected state of the system is unique and differs from all other observed states, but it is not often obvious in what sense; post-hoc analysis or domain experts are needed to interpret the results.

Preprocessing can eliminate information from the data series, thus can filter out aspects considered uninteresting. For example, we have seen that a strong global trend on a data can make all the points unique. By detrending the data, as done on random walk and LIBOR datasets, we defined that these points should not be considered unique solely based on this feature. Similarly, band-pass filtering of gravitational wave data define that states should not be considered unique based on the out-of-frequency-range waveforms.}

Future directions to develop TOF would be to form a model which is able to represent uncertainty over detections by creating temporal outlier probabilities just like Local Outlier Probabilities \cite{Kriegel09} created from LOF.
Moreover, an interesting possibility would be to make TOF applicable also on different classes of data, such as multi-channel data or point processes, like spike-trains, network traffic time-stamps or earthquake dates.

\section{Methods}{
\subsection*{TOF Analysis workflow}

\begin{enumerate}
    \item Preprocessing and applicability check: 
    
    This step varies from case to case, and depends on the data or on the goals of analysis.
    Usually it is advisable to make the data stationary.
    For example, in the case of oscillatory signals, the signal must contain many periods even from the lowest frequency components.
    If this latter condition does not hold, then Fourier filtering can be applied to get rid of the low frequency components of the signal.
        
    \item Time delay embedding:
    
    We embed the scalar time series into an $E$ dimensional space with even time delays ($\tau$, \eqref{eq:embed}, Fig.\,S1\,\textbf{A}).
    The embedding parameters can be set with prior knowledge of the dynamics or by other optimization methods.
    Figs.\,S9-S12 illustrates our parameter hunting procedure, where the $\tau$ was chosen as the first zero point of the autocorrelation function of the signal or as the first minima, if it does not reach the zero level. The embedding dimension was estimated by finding the embedding dimension where the estimated dimension started to deviate from the embedding dimension.
    This procedure worked well for dynamical systems (Fig,\,S9-S10) but not for the LIBOR which is more likely to be generated by a stochastic process.
    Here, the estimated dimension increased with the embedding dimension without reaching a plateau (Fig.\,S11).
    \zsiga{Thus in this case, the embedding dimension and delay was estimated based on the minimal normalized differential entropy \cite{Gautama03}, which selects the embedding with the most structure in it (Fig.\,S12).}
    
    \item kNN Neighbor search:
    
    We search for k-Neighborhoods around each datapoint using scipy cKDTree implementation of the kDTree algorithm in statespace and save the distance and temporal index of neighbors \cite{Bentley75}.
    
    \item TOF score computation according to equation \eqref{eq:TOF}.
    
    \item Threshold ($\theta$) application on TOF score to detect unicorns (Fig.\,S1\,\textbf{C}): 
    
    The threshold can be established by prior knowledge, by clustering techniques or supervised learning.
    The maximum event length parameter ($M$) determines the level of threshold on TOF score (Eq.\,\ref{eq:threshold}): we set the threshold according to prior knowledge about the longest possible occurrence of the event.
    After thresholding, we may apply a padding around detected points with symmetric window length $w=k/2$, since the $k$ parameter sets the minimal length of the detectable events.
\end{enumerate}

We implemented these steps in the python programming language (python3)\zsiga{, the software is available at {https://github.com/phrenico/uniqed}.}

\zsiga{Detailed description of the data generation process and analysis steps can be found in the Supporting Information.}

\subsection*{Model Evaluation metrics\label{Metrics}}
We used precision, recall, $\mathrm{F}_1$ score and ROC-AUC to evaluate the detection-performance on the simulated datasets.

The precision metrics measures the ratio of true positive hits among all the detections:
\begin{equation}
	\mathrm{precision} = \frac{\mathrm{true \, positives}}{\mathrm{true \, positives} + \mathrm{false \, positives}}
\end{equation}
The recall evaluates what fraction of the points to be detected were actually detected:
\begin{equation}
	\mathrm{recall} = \frac{\mathrm{true \, positives}}{\mathrm{true \, positives} + \mathrm{false \, negatives}}
\end{equation}
 $\mathrm{F}_1$ score is the harmonic mean of precision and recall and it provides a single scalar to rate model performance: 
\begin{equation}
	F_1 = 2 \, \frac{\mathrm{precision} \times \mathrm{recall}}{\mathrm{precision} + \mathrm{recall}}
\end{equation}

As an alternative evaluation metrics we applied the area under Receiver Operating Characteristic curve \cite{Bradley97}.
The ROC curve consists of point-pairs of True Positive Rate (recall) and False Positive Rate parametrized by a threshold ($\alpha$, Eq.\,\ref{eq:ROC}).
\begin{equation}\label{eq:ROC}
	ROC(\alpha) := \left( \mathrm{TPR}(\alpha), \mathrm{FPR}(\alpha) \right)
\end{equation}
where $\alpha \in [-\infty, \infty]$.

We computed the \soma{mean and standard deviation} from the 100 simulations on each simulated datasets (Fig.\,\ref{fig:performance}).
}






\printbibliography

\newpage

\section*{Appendix}
    
\appendix
\renewcommand\thefigure{\arabic{figure}}
\renewcommand\figurename{Supplementary Fig.}
\setcounter{figure}{0} 

\appendix
\begin{refsection}

\begin{figure}[htb!]
\includegraphics[width=\textwidth]{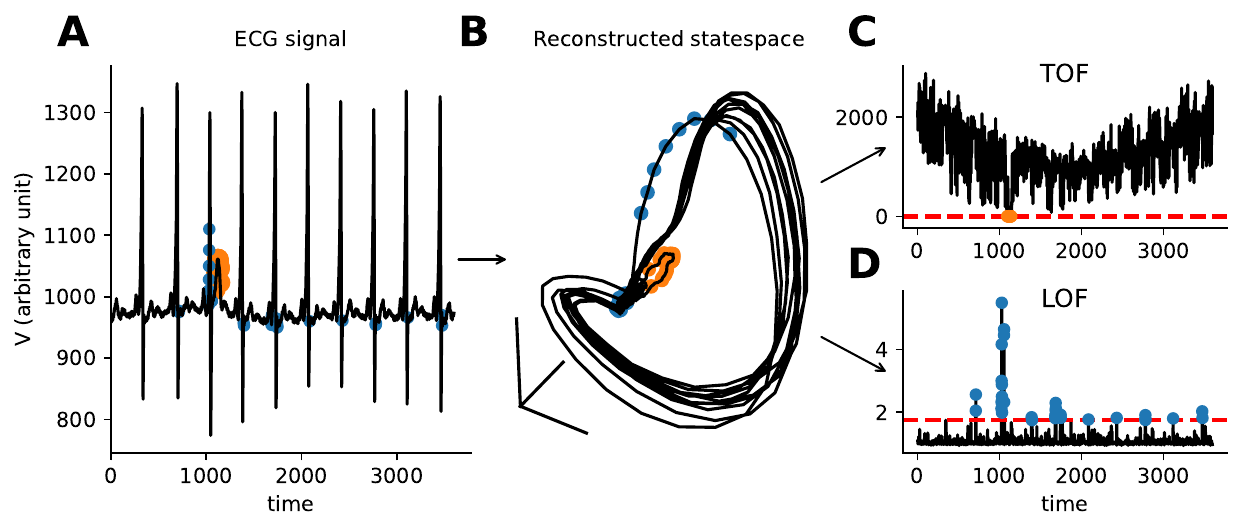}

    \caption{
        \textbf{The worklow for TOF and LOF analysis for time series.}
        (a) We start with a time series generated by a dynamical system; orange and blue marks TOF and LOF detections respectively. 
        (b) As a next step of our analysis we apply time delay embedding, then kNN search in the reconstructed state space.
        (c) We calculate TOF and LOF scores and apply thresholds on the outlier scores to detect anomalies.
    }
    \label{fig:tof_workflow}
\end{figure}

\section*{TOF Analysis workflow}

The main steps of the TOF analysis are recapitulated here for completeness:

\begin{enumerate}
    \item Preprocessing and applicability check:
    
    This step varies from case to case, and depends on the data or on the goals of analysis.
    Usually it is advisable to make the data stationary.
    For example, in the case of oscillatory signals, the signal must contain many periods even from the lowest frequency components.
    If this latter condition does not hold, then Fourier filtering can be applied to get rid of the low frequency components of the signal.
        
    \item Time delay embedding:
    
    We embed the scalar time series into an $E$ dimensional space with even time delays $\tau$ (Fig.\,\ref{fig:tof_workflow}\,\textbf{A}):
    
    \begin{equation}\label{eq:sup_embed}
    X(t) = [ x(t), x(t+\tau), x(t+2\tau), \ldots x(t+(E-1)\tau) ]
    \end{equation}

    The embedding parameters can be set with prior knowledge of the dynamics or by other optimization methods.
    Such optimization methods include the first minimum or zerocrossing of the autocorrelation function (for delay selection), the false nearest neighbor method \cite{Rhodes97, Krakovska15} or the differential entropy based embedding optimizer that we applied \cite{Gautama03}. Figs.\,S9-S12 illustrates our parameter hunting procedure, where the $\tau$ was chosen as the first zero point of the autocorrelation function of the signal or as the first minima, if it does not reach the zero level. The embedding dimension was estimated by finding the embedding dimension where the estimated dimension started to deviate from the embedding dimension. This procedure worked well for dynamical systems (Fig,\,S9-S10) but not for the LIBOR which is more likely to be generated by a stochastic process. Here, the estimated dimension increased with the embedding dimension without reaching a plateau (Fig.\,S11). Thus in this case, the embedding dimension was estimated based on the differential entropy (Fig.\,S12).
    
    \item kNN Neighbor search:
    
    We search for k-neighborhoods around each datapoint in the statespace using the kDTree algorithm and save the distance and temporal index of neighbors \cite{Bentley75}.
    
    \item Compute TOF score:
    
    \begin{equation}\label{eq:sup_TOF}
	\mathrm{TOF} \left( t \right) =  \sqrt[\leftroot{-2}\uproot{16} q]{ \frac{\sum_{i=1}^{k}{\left| t-t_i \right| ^q} }{k} }.
    \end{equation}
    Where $t$ is the time index of the sample point ($X(t)$) and $t_i$ is the time index of the $i$-th nearest neighbor in reconstructed state-space. Where $q\in \mathcal{R}^{+}$, in our case we use $q=2$.
    
    \item Apply a threshold $\theta$ on TOF score to detect unicorns (Fig.\,S1\,\textbf{C}): 
    
    The threshold can be established by prior knowledge, by clustering techniques or supervised learning.
    The maximum event length parameter ($M$) determines the level of threshold on TOF score:
    \begin{equation}
    \theta = \sqrt{ \frac{\sum_{i=0}^{k-1}{\left( M-i \Delta t \right) ^2} }{k} } \quad \bigg| \quad k \Delta t \stackrel{!}{\leq} M
    \label{eq:sup_threshold}
    \end{equation}
    We set the threshold according to prior knowledge about the longest possible occurence of the event.
    After thresholding, we may apply a padding around detected points with symmetric window length $w=k/2$, since the $k$ parameter sets the minimal length of the detectable events.
\end{enumerate}

We implemented these steps in the python programming language (python3), the software is available at

{https://github.com/phrenico/uniqed}.

The code builds on standard scientific python modules, i.\,e.\,the neighborhood search is established by the kd-tree algorithm of the scipy package \cite{2020SciPy-NMeth}.
Embedding parameter optimization was carried out by custom python scripts.
Furthermore, we used the scikit-learn package \cite{scikit-learn} to calculate LOF. 
We implemented the brute-force discord discovery algorithm \cite{Keogh05} (Keogh) by custom python and scilab scripts and we used the R implementation of RRA \cite{Senin15, jmotif} (Senin) discord discovery algorithm on all simulated datasets.

\section*{Mean and variance for $q=1$}

The mean and the variance of TOF can be computed for uncorrelated noise in the continuous-time limit, where the typical properties of the metrics can be introduced.
The expectation of the first neighbor is easy to compute (Eq. \ref{eq:tof_mean}), if we take the probability density function ($p(\tau)$) as uniform; this is the assumption of white noise.
Additionally, the pdf is independent of the rank of the neighbor ($k$), and thus the mean is the same for all neighborhood sizes.
By the previous assumptions, the  mean is simply a quadratic expression:

\begin{equation}\label{eq:tof_mean}
    \begin{array}{cc}
        \langle TOF_{q=1} \rangle & \;=\;
        \int_0^T | t - \tau | \; p(\tau) \; \mathrm{d}\tau \;=\;
        \frac{1}{T} \; \int_0^T | t - \tau | \;  \mathrm{d}\tau \;=\;
        \frac{t^2}{T} - t + \frac{T}{2}
    \end{array}  
\end{equation}

with the method of moments, we calculate the variance for $k=1$:

\begin{equation}
    \begin{array}{cc}
        \langle TOF_{q=1}^2 \rangle & \;=\;
        \int_0^T ( t - \tau )^2 \; p(\tau) \; \mathrm{d}\tau 
        \;=\; \frac{1}{T} \; \int_0^T ( t - \tau )^2 \;  \mathrm{d}\tau
        \;=\; t^2 - tT + \frac{T^2}{3}
    \end{array}    
\end{equation}

\begin{equation}
    \begin{array}{cc}
        \sigma_{q=1}^2 & \;=\;
        \langle TOF_{q=1}^2 \rangle - \langle TOF_{q=1} \rangle^2 \;=\;
        -\frac{t^4}{T^2} + \frac{2 t^3}{T} - t^2 + \frac{T^2}{12}
    \end{array}    
\end{equation}

if we have $k$ neighbors, then the variance is reduced by a $1/k$ factor:
\begin{equation}
    \begin{array}{cc}
        \sigma_{q=1, k}^2 & \;=\;
        \langle TOF_{q=1}^2 \rangle - \langle TOF_{q=1} \rangle^2 \;=\; \frac{1}{k}
        \left(-\frac{t^4}{T^2} + \frac{2 t^3}{T} - t^2 + \frac{T^2}{12} \right)
    \end{array}    
\end{equation}

\begin{figure}[htb!]
\includegraphics[width=0.9\linewidth]{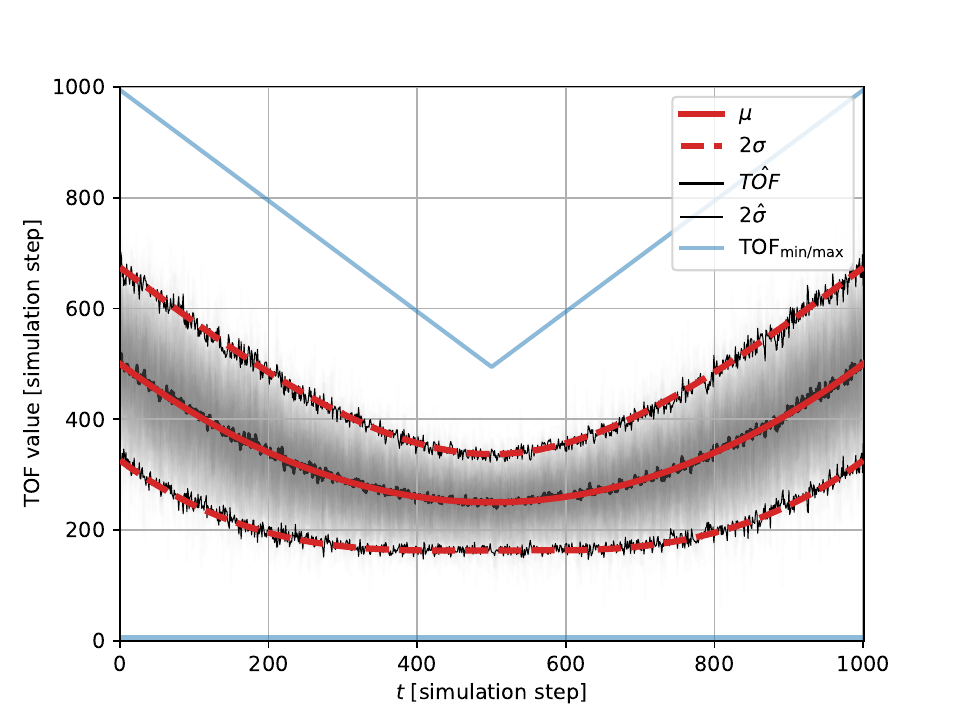}
    \caption{
        \textbf{Properties of TOF for white noise data: theory and simulations.} The expectation of TOF is computed as a function of temporal position in the time series ($q=1$, thick red line), also the standard deviation was calculated (dashed red line). The average (thick black line) and standard deviation (thin black line) of $n=100$ instances (grey shading). The minimal and maximal possible TOF vales are also charted (blue lines).
    }
    \label{fig:sup_tof_mean_q1}
\end{figure}

To test whether these theoretical arguments fit to data, we simulated random noise time series ($n=100, T=1000$) and computed the mean TOF score and standard deviation (Fig.\,\ref{fig:sup_tof_mean_q1}).
We found, that theoretical formulas described the behaviour of TOF perfectly.

\section*{Mean and variance for $q=2$}

\begin{figure}[htb!]
\includegraphics[width=0.9\linewidth]{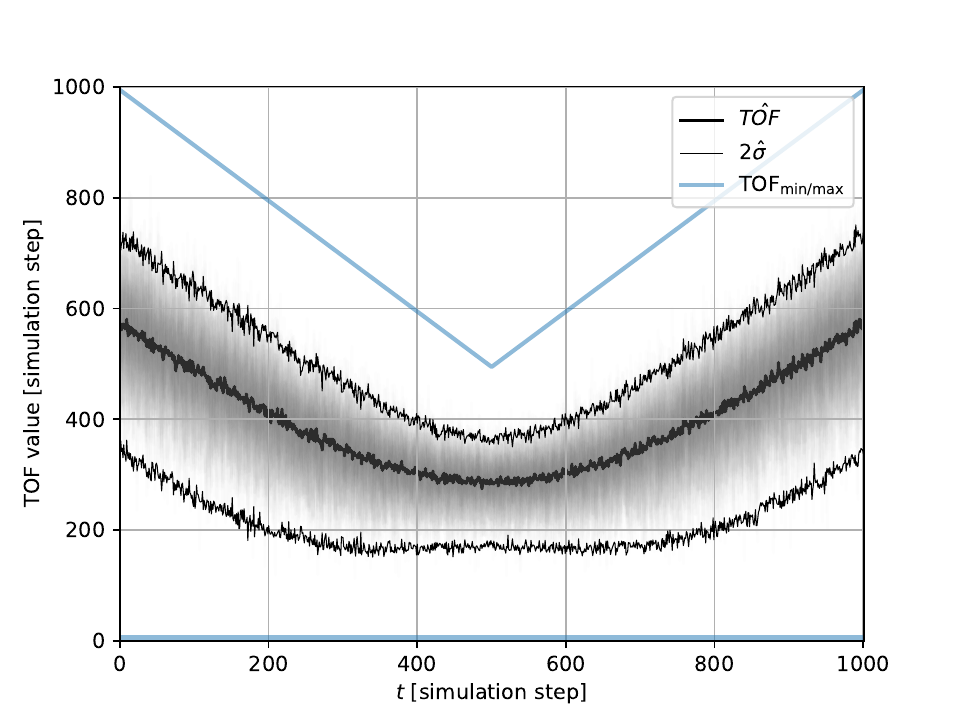}
    \caption{
        \textbf{Properties of TOF for white noise data 2: simulations}
        The baseline of TOF with $q=2$. 
        The average (thick black line) and standard deviation (thin black line) of $n=100$ instances (grey shading).
    }
    \label{fig:sup_tof_mean_q2}
\end{figure}

The exact statistics is hard to calculate, when the value of the $q$ exponent is not equal to one. Here we compute a vague approximation for $q=2$ (Fig.\,\ref{fig:sup_tof_mean_q2}). By computing the mean and variance for TOF squared, and taking the squareroot of these values can get a feeling about the properties of TOF$_{q=2}$ respectively.

\begin{equation}
    \begin{array}{r}
        \left< \mathrm{TOF}_{\mathrm{noise}, q=2}^{2} \right> = \int_0^T ( t - \tau )^2 \; p(\tau) \; \mathrm{d}\tau 
        \;=\; \frac{1}{T} \; \int_0^T ( t - \tau )^2 \;  \mathrm{d}\tau
        \;=\; t^2 - tT + \frac{T^2}{3}
    \end{array}    
\end{equation}

the second moment is as follows:

\begin{equation}
    \begin{array}{r}
        \left< \mathrm{TOF}_{\mathrm{noise}, q=2}^{4} \right> = \int_0^T ( t - \tau )^4 \; p(\tau) \; \mathrm{d}\tau 
        \;=\; \frac{1}{T} \; \int_0^T ( t - \tau )^4 \;  \mathrm{d}\tau
        \;=\; \frac{t^5 + (T-t)^5}{5T}
    \end{array}    
\end{equation}

Thus using the method of moments we can get the variance of the $TOF_{q=2}^2$:
\begin{equation}
    \begin{array}{r}
        \mathrm{Var} \left( \mathrm{TOF}_{\mathrm{noise}, q=2}^{2} \right) 
        \;=\; \frac{t^5 + (T-t)^5}{5 T} - \left( t^2 - tT + \frac{T^2}{3} \right)^2
    \end{array}    
\end{equation}

\section*{Generation of simulated datasets}

\subsubsection*{Simulated logistic map and stochastical datasets}

We simulated 4 systems: logistic map with linear tent map outlier segment, logistic map with linear outlier segment, simulated ECG data with tachycardia outlier segment and random walk with linear outlier segment.
The first three datasets stem from deterministic dynamics, whereas the last simulated dataset has stochastic nature.

We generated 100 time series from each type, the length and the position of outlier segments were determined randomly in each case. 

\subsubsection*{Logistic map with tent-map anomaly}\label{section:tentmap}

$100$ instances of logistic map data-series were simulated ($N=2000$) with one randomly (uniform) inserted outlier period in each dataset. The length of outlier periods was randomly chosen with length between $2-200$.
The basic dynamics in normal conditions were governed by the update rule:
\begin{equation}
	x_{t+1} = r x_t (1 - x_t)
\label{eq:logistic}
\end{equation}
where $r=3.9$.
The equation was changed during anomaly periods:
\begin{equation}
	x_{t+1} = 1.59 - 2.15 \times |x_t - 0.7| - 0.9 \times x_t
	\label{eq:tent}
\end{equation}
where $a=\pm 0.001$. To make sure that the time series was bounded in the $I=[0, 1]$ interval, the sign of $a$ was changed if required: initially $a>0$ and the sign is reversed when $x_t>=1$, thus restricting the time series to the desired interval $I$.

\subsubsection*{Logistic map with linear anomaly}

The background generation process exhibited the logistic dynamics (Eq.\,\ref{eq:logistic}) while the anomaly can be described by linear time dependence:
\begin{equation}\label{eq:lin}
	x_{t+1} = a * x_t + x_t
\end{equation}
Here we used $a=\pm 0.001$, where the sign of the slope is positive by default and changes when the border of the $(0, 1)$ domain is reached ensuring reflective boundary condition.

\subsubsection*{Random walk data with linear anomaly}
We simulated 100 instances of multiplicative random walks with 2-200 timestep long linear outlier-insets.
The generation procedure was as follows:
\begin{enumerate}
	\item Generate $w_i$ random numbers from a normal distribution with $\mu=0.001$ and $sigma=0.01$
	\item Transform $w_i$ to get the multiplicative random walk data as follows: $x_i = \prod_{j=1}^i (1 + w_j)$
	\item Draw the length ($L$) and position of outlier-section from discrete uniform distributions between $2-200$ and $1 - (N-L)$ respectively.
	\item Use linear interpolation between the section-endpoint values.
\end{enumerate}

\subsubsection*{Simulated ECG datasets with tachyarrhythmic segments}

We generated artificial ECG data series according to the model of Ryzhii and Ryzhii \cite{Ryzhii14}.
The pacemakers of the heart: the sinoatrial node (SA), the atroventricluar node (AV) and the His-Purkinje system (HP) are simulated by van der Pol equations:
\begin{equation}
    SN
    \begin{cases}
        \Dot{x}_1 &= y_1 \\
        \Dot{y}_1 &= -a_1 y_1 (x_1 - u_{11}) (x_1 - u_{12}) \\
        & \qquad{} - f_1 x_1 (x_1 + d_1) (x_1 + e_1)
    \end{cases}
\end{equation}

\begin{equation}
    AV
    \begin{cases}
        \Dot{x}_2 &= y_2 \\
        \Dot{y}_2 &= -a_2 y_2 (x_2 - u_{21}) (x_2 - u_{22})  \\
        & \qquad{} - f_2 x_2 (x_2 + d_2) (x_2 + e_2)  \\
        & \qquad{} + K_{SA-AV} (y_1^{\tau_{SA-AV}}-y_2)
    \end{cases}
\end{equation}

\begin{equation}
    HP
    \begin{cases}
        \Dot{x}_3 &= y_3 \\
        \Dot{y}_3 &= -a_3 y_3 (x_3 - u_{31}) (x_1 - u_{32})  \\
        & \qquad{} - f_3 x_3 (x_3 + d_3) (x_3 + e_3)  \\
        & \qquad{} + K_{AV-HP} (y_2^{\tau_{AV-HP}}-y_3)
    \end{cases}
\end{equation}
where the parameters were set according to Ryzhii\cite{Ryzhii14}: $a_1=40$, $a_2=a_3=50$, $u_{11} = u_{21} = u_{31} = 0.83$, $u_{12} = u_{22} = u_{32} = -0.83$, $f_1 = 22$, $f_2 = 8.4$, $f_3 = 1.5$, $d_1 = d_2 = d_3 = 3$, $e_1 = 3.5$, $e_2 = 5$, $e_3 = 12$ and $K_{\mathrm{SA-AV}} = K_{\mathrm{AV-HP}} = f_1$.

The following FitzHugh-Nagumo equations describe the atrial and ventricular muscle depolarization and repolarization responses to pacemaker activity:

\begin{equation}
    \mathrm{P \; wave}
    \begin{cases}
        \Dot{z}_1 &= k_1 (-c_1 z_1 (z_1 - w_{11}) (z_1 - w_{12})  \\
        & \qquad{} - b_1 v_1 - d_1 v_1 z_1 + I_{\mathrm{AT}_{\mathrm{De}}}) \\
        \Dot{v}_1 &= k_1 h_1 (z_1 - g_1 v_1)
    \end{cases}
\end{equation}

\begin{equation}
    \mathrm{Ta \; wave} 
    \begin{cases}
        \Dot{z}_2 &= k_2 (-c_2 z_2 (z_2 - w_{21}) (z_2 - w_{22})  \\
        & \qquad{} - b_2 v_2 - d_2 v_2 z_2 + I_{\mathrm{AT}_{\mathrm{Re}}}) \\
        \Dot{v}_2 &= k_2 h_2 (z_2 - g_2 v_2)
    \end{cases}
\end{equation}

\begin{equation}
    \mathrm{QRS}
    \begin{cases}
        \Dot{z}_3 &= k_3 (-c_3 z_3 (z_3 - w_{31}) (z_2 - w_{32})  \\
        & \qquad{} - b_3 v_3 - d_3 v_3 z_3 + I_{\mathrm{VN}_{\mathrm{De}}}) \\
        \Dot{v}_3 &= k_3 h_3 (z_3 - g_3 v_3)
    \end{cases}
\end{equation}

\begin{equation}
    \mathrm{T \; wave}
    \begin{cases}
        \Dot{z}_4 &= k_4 (-c_4 z_4 (z_4 - w_{41}) (z_4 - w_{42})  \\
        & \qquad{} - b_4 v_4 - d_4 v_4 z_4 + I_{\mathrm{VN}_{\mathrm{Re}}}) \\
        \Dot{v}_4 &= k_4 h_4 (z_4 - g_4 v_4)
    \end{cases}
\end{equation}
where $k_1 = 2 \times 10^3$, $k_2 = 4 \times 10 ^ 2$, $k_3 = 10^4$, $k_4 = 2 \times 10^3$, $c_1 = c_2 = 0.26$, $c_3 = 0.12$, $c_4 = 0.1$ $b_1 = b_2 = b_4 = 0$, $b_3 = 0.015$, $d_1 = d_2 = 0.4$, $d_3 = 0.09$, $d_4 = 0.1$, $h1 = h2 = 0.004$, $h_3 = h_4 = 0.008$, $g_1 = g_2 =g_3 = g_4 = 1$, $w11 = 0.13$, $w_{12} = =w_{22} = 1$, $w_{21} = 0.19$, $w_{31} = 0.12$, $w_{32} = 0.11$, $w_{41} = 0.22$, $w_{42} = 0.8$.

The input-currents ($I_i$) are caused by pacemaker centra.

\begin{equation}
    I_{\mathrm{AT}_{\mathrm{De}}} = 
    \begin{cases}
        0 & \mathrm{for} \quad y_1 \leq 0 \\
        K_{\mathrm{AT}_{\mathrm{De}}} y_1 & \mathrm{for} \quad y_1 > 0
    \end{cases}
\end{equation}

\begin{equation}
    I_{\mathrm{AT}_{\mathrm{Re}}} = 
    \begin{cases}
        - K_{\mathrm{AT}_{\mathrm{Re}}} y_1 & \mathrm{for} \quad y_1 \leq 0 \\
        0 & \mathrm{for} \quad y_1 > 0 
    \end{cases}
\end{equation}

\begin{equation}
    I_{\mathrm{VN}_{\mathrm{De}}} = 
    \begin{cases}
        0 & \mathrm{for} \quad y_3 \leq 0 \\
        K_{\mathrm{VN}_{\mathrm{De}}} y_3 & \mathrm{for} \quad y_3 > 0
    \end{cases}
\end{equation}

\begin{equation}
    I_{\mathrm{VN}_{\mathrm{Re}}} = 
    \begin{cases}
        - K_{\mathrm{VN}_{\mathrm{Re}}} y_3 & \mathrm{for} \quad y_3 \leq 0 \\
        0 & \mathrm{for} \quad y_3 > 0 
    \end{cases}
\end{equation}
where $K_{\mathrm{AT}_{\mathrm{De}}} = 4 \times 10^{-5}$, $K_{\mathrm{AT}_{\mathrm{Re}}}= 4 \times 10^{-5}$, $K_{\mathrm{VN}_{\mathrm{De}}}= 9 \times 10^{-5}$ and $K_{\mathrm{VN}_{\mathrm{Re}}}= 6 \times 10^{-5}$. 

The net ECG signal is given by the weighted sum of muscle depolarization and repolarization responses:
\begin{equation}
    ECG = z_0 + z_1  - z_2 + z_3 + z_4
\end{equation}
where $z_0 = 0.2$ is a constant offset.

We simulated $100$ instances of $t=100$ seconds long ECG data with base rate parameter chosen from a Gaussian distribution ($f_1 \sim \mathcal{N}(\mu=22,\sigma=3)$).
We randomly inserted $2-20$ seconds long fast heart-beat segments by adjusting the rate parameter ($f_1 \sim \mathcal{N}(\mu=82,\sigma=3)$).
The simulations were carried out by the ddeint python package, with simulation time-step $\Delta t=0.001$ from random initial condition and warmup time of $2$ seconds.
Also, a $10 \times$ rolling-mean downsampling was applied on the data series before analysis.

\subsubsection*{Generating non-unique anomalies dataset}

To show the selectiveness of TOF for the detection of unicorns, we simulated logistic map data with two tent-map outlier segments.
The governing equations were the same as in the previous section, but instead of one, we randomly placed two non-overlapping outlier segments into the time series during data generation, ($N=2000, L=20-200$).

\section*{Analysis steps on simulations}

We applied optional preprocessing, and ran TOF, LOF, brute-force discord discovery \cite{Keogh05} (Keogh) and RRA \cite{Senin15} (Senin) discord discovery algorithms on all simulated datasets.

We applied the same preprocessing on the datasets for all anomaly detection methods on the four datasets.
For the logistic map datasets no preprocessing was applied.
For the simulated ECG data we applied a tenfold downsampling, the sampling period became $\Delta t = 0.01$\,s.
For the multiplicative random walk with linear anomaly dataset we applied a logarithmic difference as a preprocessing step to get rid of nonstationarity in the time series (Eq.\,\ref{eq:logdiff}).
\begin{equation}\label{eq:logdiff}
    y_t = \log(x_t)-\log(x_{t-1})
\end{equation}
where $x$ is the original time series, $\log$ is the natural logarithm and $y$ is the preprocessed time series.

In the case of TOF and LOF, time delay embbeding was applied on the scalar time series.
For the  logisticmap - tentmap and - linear datasets the dynamics is well known and 1-dimensional, so $E=3$ is enough to embed the signal. Also, $\tau=1$ time-step was proper for an embedding delay.
For the ECG dataset the dynamics naively seems to be approximately 2-dimensional, so we set $E=3$, which may be enough to reconstruct the dynamics, also $\tau=0.01$ s was set as embedding delay.

After embedding, the ROC AUC score was computed to find optimal neighborhood sizes in the $k \in \{1, .. 199\}$ range with the TOF and the LOF methods (Fig.\,3\,A).

As a next step of comparison, a screening over the anomaly-length parameter was performed and optimal $\mathrm{F}_1$ score was registered for the TOF, LOF and Keogh (Fig.\,3\,B)).
More specifically, the $F_1$-score metrics, precision and recall were calculated on the simulated datasets in the function of event length parameter in the $(1, 300)$ integer range for the discrete-time datsets and in the $[1, 3000)$ integer range on the simulated ECG dataset.
The embedding dimension was set to $E=3$, and embedding delay $tau=1$, the neighborhood size parameter was set to $k=4$ in the case of TOF, and $k=28$, $1$, $99$, $1$ for LOF applied on the logistic map-tent map, logistic map-linear simulated ECG and random walk-linear datasets respectively.
We applied the brute force discord discovery on the simulated datasets, and calculated ROC AUC and $\mathrm{F}_1$ score in the function of neighborhood size and window length parameters respectively.
The window length parameter were varied the same way we changed the event length parameter for TOF or the percentage of outliers for LOF.

We ran Senin's Rare Rule Anomaly (RRA) algorithm on the simulated datasets for discord discovery with automated event length selection \cite{Senin15, jmotif}.
We set the maximal sliding window size to $200$ time-steps for the discrete time simulations and to $2000$ time-steps for the simulated ECG datasets. The $p_{aa}=4$ was set according to the example script and the alphabet size was set to a=8.

To show that TOF finds unique events, we applied the algorithm on time series with multiple anomalies.
We made no preprocessing on the dataset and the embedding parameters were set to $E=3$ and $\tau=1$.
Also the neighborhood size was set to $k_\mathrm{TOF}=4$ and $k_\mathrm{LOF}=28$ for TOF and LOF respectively.
We calculated the ROC AUC values for each simulated instance and plotted these values as the function of inter event interval (Fig.\,4).

\begin{figure}[htb!]
\includegraphics[width=0.9\linewidth]{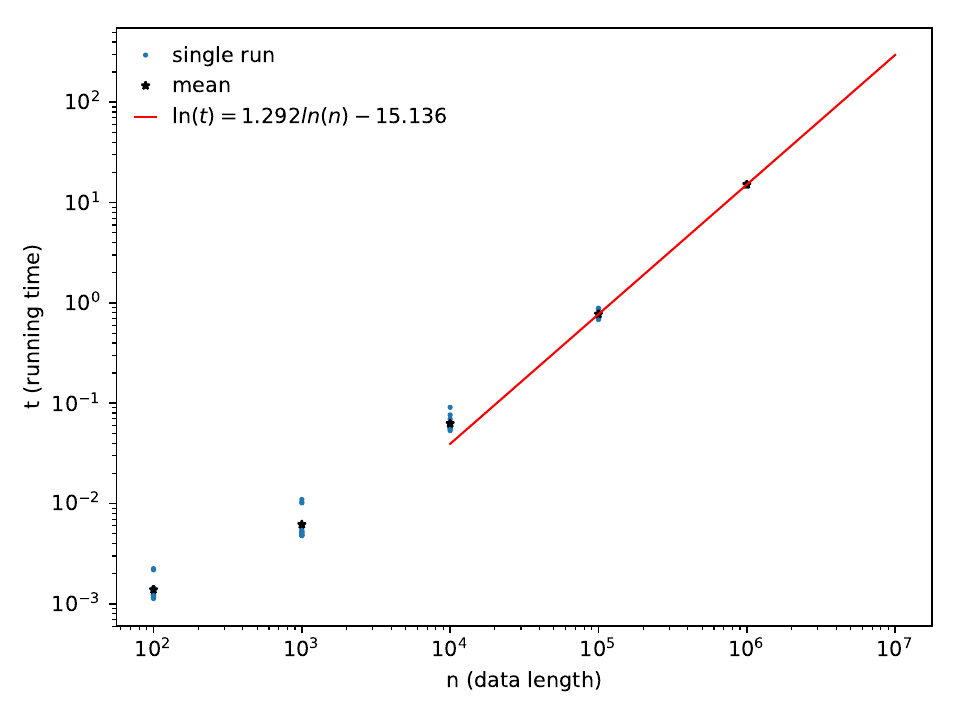}
    \caption{
        \textbf{Running time as a function of time series length.}
        Single runs (blue dots) and datalength-wise means (black stars) are shown along with the line fitted on the last two lengths (red line, ($d=3$, $\tau=1$, $k=4$)
    }
    \label{fig:sup_runtime}
\end{figure}

\section*{Computational complexity and running time}

The current implementation of the TOF algorithm contains a time delay embedding, a $k$NN search, the computation of TOF score from the neighborhoods and threshold application.
The time-limiting step is the neighbor-search, which uses the scipy cKDTree implementation of the kDTree algorithm \cite{Bentley75}.
The most demanding task is to build the tree data-structure; its complexity is $O(k n \log{n})$ \cite{Brown15}
 and the nearest neighbor search has $O(\log n)$ complexity.

We applied the TOF algorithm on random noise from $10^2-10^6$ sample size, $15$ instances each ($d=3$, $\tau=1$, $k=4$).
The running-time on the longest tested dataset containing $10^6$ points was $15,144\pm0.351$ secs (Fig.\,\ref{fig:sup_runtime}) on a laptop powered by Intel\textregistered Core\texttrademark i5-8265U CPU.

We fit a line on the log-log plot where the data-lengths were $n=10^5$ and $n=10^6$.
The following equation described the fitting line:
\begin{equation}
    \begin{array}{c}
         log(t) = 1.292 \ln(n) - 15.136
    \end{array}
\end{equation}
.

\begin{figure}[htb!]
\includegraphics[width=0.9\linewidth]{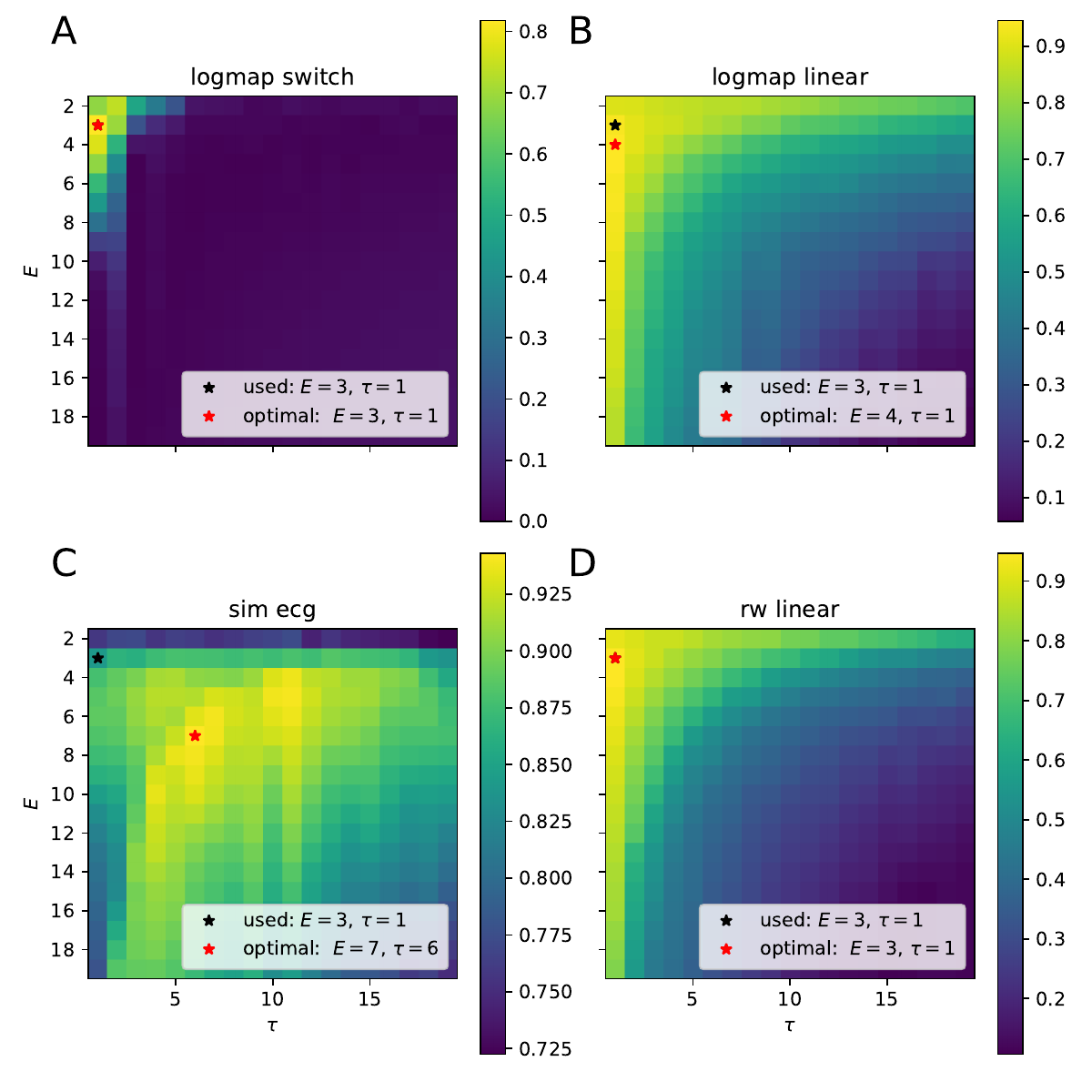}
    \caption{
        \textbf{F$_1$ score in the function of embedding dimension and embedding delay for the simulated datasets ($N=15$).}
        \textbf{A} Logistic map with tentmap anomaly ($E^*=3$, $\tau=1$, F$_1^{\mathrm{max}}=0.818$),
        \textbf{B} logistic map with linear anomaly ($E^*=4$, $\tau=1$, F$_1^{\mathrm{max}}=0.946$),
        \textbf{C} simulated ECG with tachycardia ($E^*=7$, $\tau=6$, F$_1^{\mathrm{max}}=0.942$) and
        \textbf{D} random walk with linear anomaly ($E^*=3$, $\tau=1$, F$_1^{\mathrm{max}}=0.947$).
    }
    \label{fig:sup_paramdep}
\end{figure}

\section*{Embedding-parameter dependence}

We investigated the parameter-dependence of TOF detection perfomance by measuring the F$_1$ score on a range of embedding dimension ($d \in \{2, .. 19\}$) and embedding delay ($\tau \in \{1, .. 19\}$) pairs, while keeping the threshold parameter fixed on the simulated datasets ($N=15$ each, Fig.\,\ref{fig:sup_paramdep}).
The threshold parameter was set to $110$ for the discrete-time datasets, and $1100$ for the simulated ECG dataset.

We found that the performance was parameter-depedent, but near optimal parameters can be found in most cases with basic knowledge about the investigated system.

It is worth mentioning that the optimal and near-optimal parameter combinations traced out a hyperbola in the search space pointing a quazy-constant optimal embedding-window specific to each dataset.

\section*{Maximum expected F$_1$ score of the simulated dataset}

When the event length is unknown, the maximal achievable F$_1$ score may be limited by the event length parameter.

\begin{figure}[tb!]
\includegraphics[width=\linewidth]{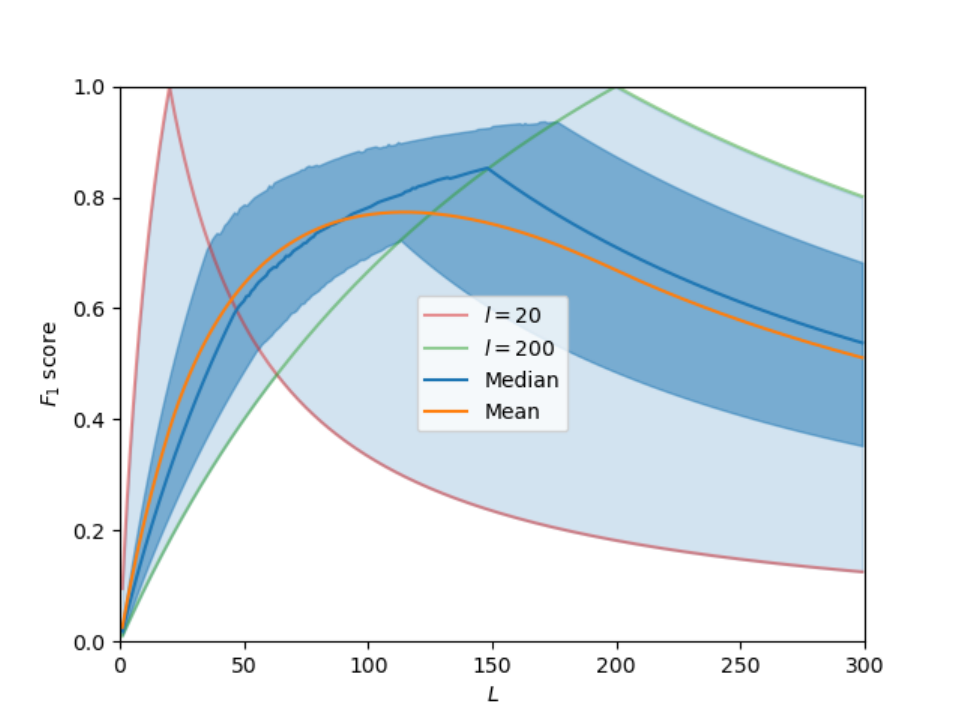}
    \caption{
        \textbf{Simulated Expectation of the upper limit F$_1$ score for LOF and discord in the function of length parameter on the simulated data}
        The figure shows the F$1$ scores of simulated time series ($N=10000$) with randomly varied anomaly length in the function of length parameter ($L$).
        The median (blue curve) and mean (yellow curve) of the F$_1$ scores is also marked on the figure.
        The shortest anomaly has the length of $l=20$ (red curve) time-steps and the longest one is $l=200$ time-steps long (green curve). These two curves mark the range of possible F$_1$ score values measured on the dataset (Blue shading) and to get some sense of the distribution, the inter-quartile range (strong shading) is also shown.
        The F$_1$ score strongly depends on the length parameter, the estimated maximum is at $114$ timesteps, which is around the expected event length ($110$) of the simulated outlier segments.
    }
    \label{fig:sup_f1_limit}
\end{figure}

We computed the maximal possible $F_1$ score given the length parameter of anomaly detection methods. We simulated N=10000 realizations of true event lengths drawn from a discrete uniform distibution over the [20, 200] range,  and computed the maximum possible $F_1$ score metric given the length parameter (L) in the (1, 300) range. We took the L-wise mean and median of the sample and plotted the results (Fig.\,\ref{fig:sup_f1_limit}).

\begin{figure}[tb!]
\includegraphics[width=\linewidth]{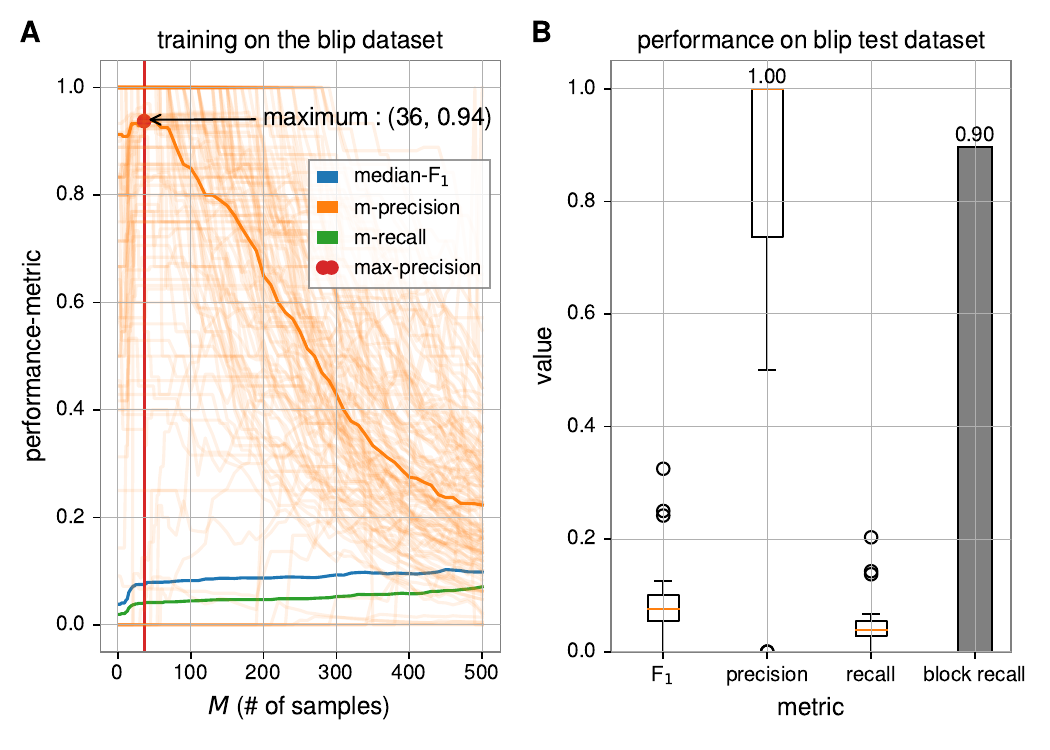}
    \caption{
        \textbf{TOF Results on the Gravity Spy blip dataset}
        \textbf{A} The mean F$_1$ score, precision and recall metrics from the training set are shown ($N=128$) and the maximal precision place ($M=36$) selected to test evaluation.
        \textbf{B} The test ($N=29$) F$_1$ score, precision,recall and block recall. The median precision is $1.00$ and the block recall (hit rate) is $0.9$.
    }
    \label{fig:sup_blip}
\end{figure}

\section*{Local Outlier Factor}

The Local Outlier Factor \cite{Breuniq00} compares local density around a point ($X$) with the density around its neighbors (Eq.\,\ref{eq:LOF}).

\begin{equation}\label{eq:LOF}
    \mathrm{LOF}_k(X) = \frac{1}{| N_k(X)|} \sum_{o \in N_{\mathrm{k}(X)}}{\frac{\mathrm{lrd}_k(o)}{\mathrm{lrd}_k(X)}}
\end{equation}
Where $| N_k(X)|$ is the cardinality of the $k$-distance neighborhood of $X$, lrd$_k$ is the local reaching density for $k$-neighborhood (see Breunig et al.\,\cite{Breuniq00} for details, Fig.\,S1).

\section*{Analysis of real-world data}

\subsubsection*{Polysomnography dataset}
We analysed a part of the first recording of the MIT-BIH polysomnographic database \cite{Ichimaru99} on Physionet \cite{Goldberger00}. The ECG data was sampled at 250 Hz.
A $160$\,s long segment was selected to be analysed, starting at $300$\,s of the recording.
The embedding parameters were set by a manual procedure to $E_\mathrm{TOF}=3$ and $\tau=0.02$\,s. The embedding delay was set according to the first zero-crossing of the autocorrelation function, embedding dimension was determined by an iterative embedding process, where the intrinsic dimensionality \cite{Szepesvari07} of the dataset was measured for various embedding dimensions (Fig\,S9).
The embedding dimension where the intrinsic dimensionality started to saturate was selected.
For LOF, the embedding dimension was set higher ($E_\mathrm{LOF}=7$), because the results became more informative about the apnea event.
The neihgborhood size was set according to simulation results; we used a smaller neighborhood for TOF ($k=11$) and a large neighborhood for LOF ($k=200$).
Moreover we set the event length to $M=5$\,s for TOF, corresponding to $3.125\%$ for LOF, which turned out to be a too loose condition. Therefore we used the more conservative $0.5\%$ threshold for LOF to get more informative results. 

\subsubsection*{Gravitational wave dataset}
We analysed the 4096 Hz sampling rate strain data of the LIGO Hanford (H1) detector around the GW150914 merger event.
The analysed 12 s recording starts 10 s before the event.
We investigated the q transform spectrogram of the time series around the event at $5 \times 10^{-4}$\,s time resolution by using the gwpy python package \cite{duncan_macleod_2020_3598469}.
Based on the spectrogram we applied 50-300 Hz bandpass filtering on the time series as a preprocessing step.
Embedding parameters were selected manually (Fig.\,S10), by choosing the first minima of the autocorrelation function for the embedding delay ($\tau=8$ sampling periods $\approx 1.95$\,ms) and then we selected the embedding dimension according to a manual procedure.
Successive embedding of the time series into higher and higher dimensional space showed, that the intrinsic dimensionality of the dataset starts to deviate from the embedding dimension at $E=6$.
Thus, we set this latter value as embedding dimension for TOF. 
For LOF a higher embedding dimension ($E=11$) led to informative results.
We set the neighborhood sizes based on our experiences with the simulated datasets: smaller value was set for TOF ($k=12$) and larger for LOF ($k=100$).
The event length was set to $M=146$\,ms for TOF as the visible length of the chirp on the spectrogram and $0.5$\,\% for LOF.
Also, a $w=7$ widening window was applied on TOF detections.

\subsubsection*{Gravity Spy blip dataset}
\textbf{Data acquisiton:}
We downloaded randomly chosen blip events registered in the Gravity Spy \cite{Zevin_2017} database from the GWOSC \cite{Abbott19} servers using the gwpy python package \cite{duncan_macleod_2020_3598469}.
Time series length was randomly chosen (0.15-2sec) around the blip events.
The start time and duration of each event was acquired from the Gravity Spy metadata file, and a random-length pre and post segment were added to the event.
After downloading the data, the data-files containing missing values were removed.
At the end of the download and quality check steps, the training set contained $N=128$ and the test set contained $N=29$ blip time series. 

\textbf{Preprocessing and application of TOF:}
We bandpass-filtered the signals ($100$-$300$ Hz) with default parameters two times (mne.filter/filter\_data function) and cropped the time series to get rid of distorted edges ($200$ timesteps).
We applied time delay embedding ($d=3$, $\tau=1$) and applied TOF to predict anomalies in the function of the event-threshold in the $1$-$500$ time-step range for the training data.
We applied the TOF algorithm on the test set with optimal threshold parameter (see below).

\textbf{Performance metrics:}
We calculated F$_1$ score, precision and recall values for the threshold range and we optimized the median precision value to select a threshold value ($M=36$, precision$=0.94$, Fig.\,S7).
Furthermore, we computed the block recall metric on the test set, which measures the ratio of datasets in which TOF found points of blip events. 

\textbf{Training and test results:}
The metrics showed high median precision, low median recall and low median F$_1$ score for the training set (Fig.\,S7\,A). 
On the test set, we applied TOF with the optimized threshold ($M=36$) and got very high median precision ($1.00$, IQR:$0.263$) and high block recall ($0.9$) with low F$_1$ score and recall (Fig.\,S7\,B).

\subsubsection*{LIBOR dataset}
The monthly LIBOR dataset was analysed to identify interesting periods.
As a preprocessing step, the first difference was applied for detrending purposes.

Optimal Embedding parameters were selected according to the minima of the relative entropy ($E=3$, $\tau=1$ month, Fig.\,S11-S12).
The neighborhood size was set manually to $k_{\mathrm{TOF}}=5$ and $k_{\mathrm{TOF}}=30$ for TOF and LOF respectively.
Also, the event length was $M=30$ for TOF and the threshold was set to $18.86$\,\% for LOF.
Also, a widening window $w=3$ was applied on TOF detections.

\newpage
\section*{Additional Tables and Figures}

\begin{figure}[htb!]
\includegraphics[width=\linewidth]{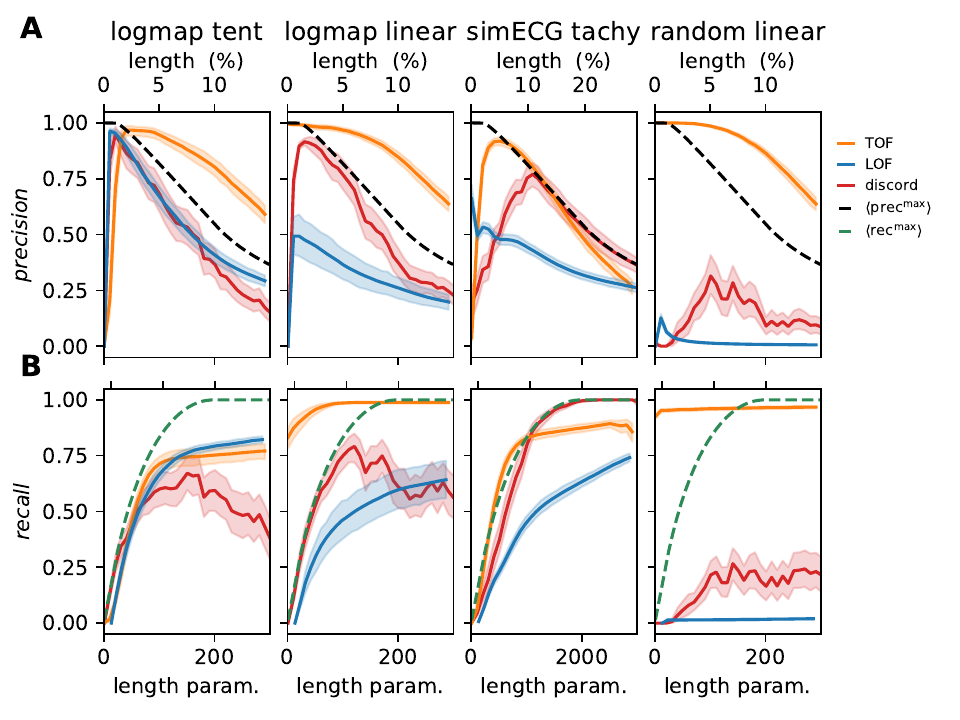}
    \caption{
        \textbf{Precision and recall and its dependence on the expected length parameters of the three methods on four different test datasets}
        Upper row mean precision, lower row mean recall and SD, over n=100 test dataseries for all four dataset types. Orange line: TOF, Blue line: LOF, Red line: discord, black-dashed: mean maximal precision for LOF and discord methods, green-dashed: mean maximal recall for LOF and discord.
        \bf{a} Precision score reached by the methods on the test data-sets.
        \bf{b} Recall score reached by the methods on the test data-sets.
    }
    \label{fig:sup_precision_recall}
\end{figure}


\begin{table*}
    \caption{\textbf{State space densities and LOF values within normal and anomalous activity.} Median and median absolute difference of the density of the points and LOF values in the reconstructed state space are shown, calculated from the distance of the 20 nearest neighbors. The density of the anomaly was significantly lower than the density generated by normal activity in two cases: the tent map anomaly in logistic backgroud the tachycardia within the normal heart rhythm.  These cases also resulted in higher LOF values of anomalies. While the density of the linear anomaly segments was not significantly different from the logistic background, the linear anomalies generated much higher density than the normal random walk time series after detrending. Correspondingly, LOF values were not significantly higher in these two cases within the anomaly than in the normal activity.} 
    \centering
    \begin{tabular}{lcccccc}
\toprule
\multicolumn{1}{c}{dataset} & $\,\,$ & \multicolumn{2}{c}{Density} &  $\,$ & \multicolumn{2}{c}{LOF} \\
{} & $\,\,$ & $Normal$ &                Anomaly &  $\,$ & $Normal$ &                Anomaly \\
\cmidrule{3-4} \cmidrule{6-7}
logmap tent     &   $\,$ &   $95.759 \pm 12.070$ &  $11.606 \pm 1.146$ &  $\,$ & $1.039 \pm 0.010$ &  $3.424 \pm 1.990$ \\
logmap linear   &   $\,$ &   $95.190 \pm 9.305$ &  $97.413 \pm 51.289$ &  $\,$ &   $1.040 \pm 0.012$ &  $1.398 \pm 0.451$ \\
sim ECG  tachy  &   $\,$ &   $10146 \pm 2227$ &  $168.370 \pm 38.699$ &  $\,$ &  $1.106 \pm 0.022$ &  $1.264 \pm 0.227$ \\
randwalk linear &   $\,$ &  $197.919 \pm 3.866 $ &  $52590 \pm 61527$ &  $\,$ &   $1.623 \pm 0.661$ &  $1.872 \pm 0.920$ \\
\bottomrule
\end{tabular}

    \label{tab:density}
\end{table*}

\begin{figure}[htb!]
\centering
\includegraphics[scale=0.6]{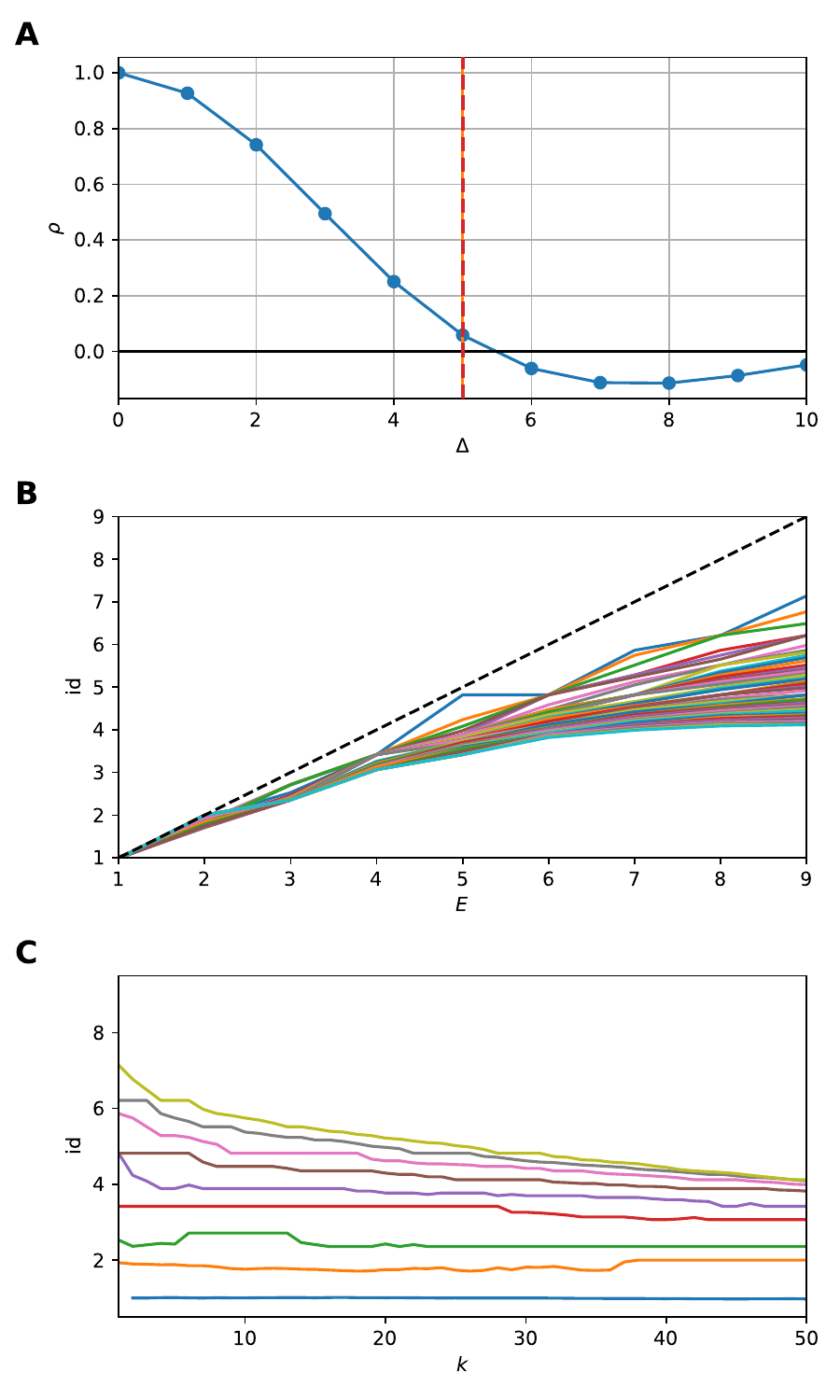}
    \caption{
    \textbf{Embedding parameter selection for the polysomnography data.}
    \textbf{a} Embedding delay was selected ($\tau=5$ sampling period) according to the first zerocrossing of the autocorrelation function. The timeshift ensures the most linearly independent axes in reconstructed state space.
    \textbf{b} The intrinsic dimensionality is measured as a function of embedding dimension ($E$) for various neighborhood sizes. The dimension-estimates start to deviate from the diagonal at $E=3$.
    \textbf{c} Intrinsic dimensionality in the function of neighborhood size ($k$) for various embedding dimensions.
    }
    \label{fig:sup_polysomn_embed}
\end{figure}

\begin{figure}[htb!]
\centering
\includegraphics[scale=0.6]{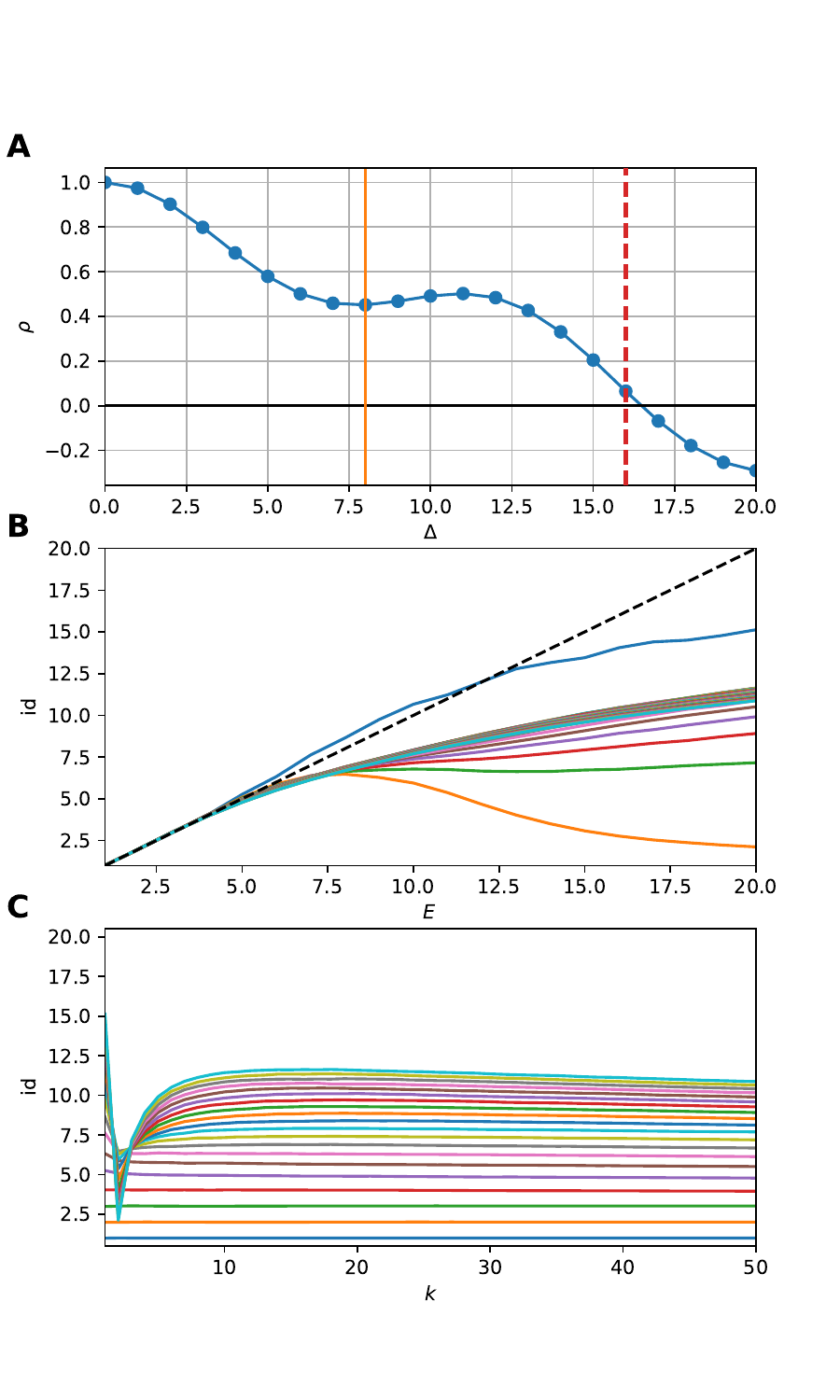}
   \caption{
        \textbf{Embedding parameter selection for the gravitational wave data.}
        \textbf{a} Embedding delay was selected ($\tau=8$ sampling period) according to the first minima of the autocorrelation function. The first zeropoint was between $16$ and $17$ sampling periods.
        \textbf{b} The intrinsic dimensionality is measured as a function of embedding dimension ($E$) for various neighborhood sizes. The dimension estimates start to deviate from the diagonal at $E=5$.
        \textbf{c} Intrinsic dimensionality as a function of neighborhood size ($k$) for various embedding dimensions.
    }
    \label{fig:sup_grav_embed}
\end{figure}

\begin{figure}[htb!]
\centering
\includegraphics[scale=0.6]{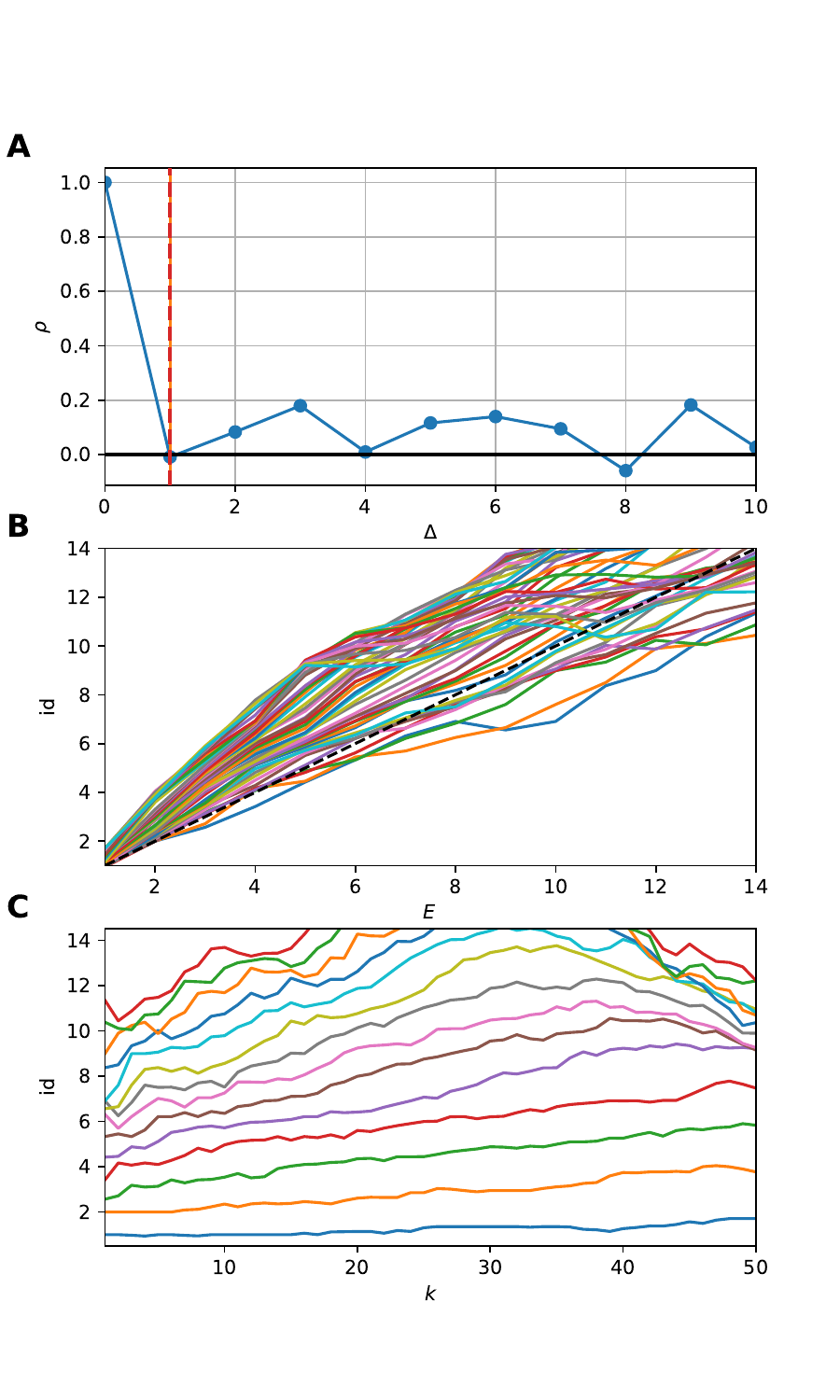}
    \caption{
        \textbf{Autocorrelation and intrinsic dimension measurement of the preprocessed LIBOR time series.}
    }
    \label{fig:sup_libor_embeda}
\end{figure}

\begin{figure}[htb!]
\includegraphics[width=\linewidth]{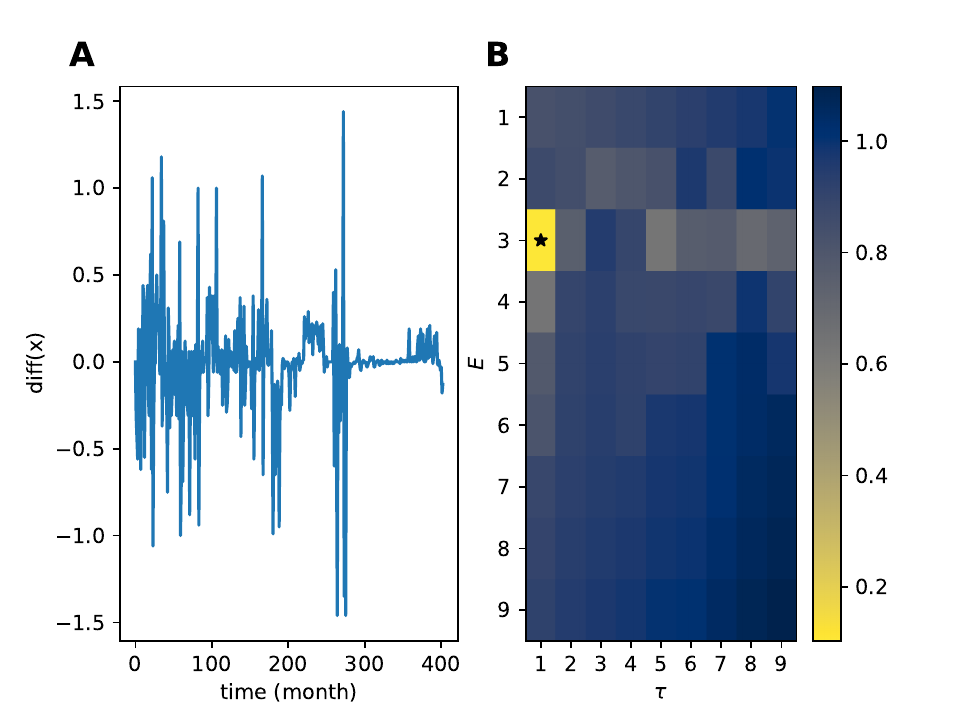}
    \caption{
        \textbf{Preprocessing and embedding parameter selection for the LIBOR time series with differential entropy.}
        (left) The discrete time derivative of the original time series was calculated to detrend the data
        (right). The minima of the entropy landscape marks the optimal embedding parameters ($E=3, \tau=1$ timestep).
    }
    \label{fig:sup_libor_embed}
\end{figure}

\FloatBarrier

\printbibliography

@article{Beggel19,
author = {Beggel, Laura and Kausler, Bernhard X. and Schiegg, Martin and Pfeiffer, Michael and Bischl, Bernd},
doi = {10.1007/s00180-018-0824-9},
issn = {16139658},
journal = {Computational Statistics},
title = {{Time series anomaly detection based on shapelet learning}},
year = {2019}
}

@article{Blazquez-Garcia20,
journal = {arXiv:2002.04236},
author = {Bl{\'{a}}zquez-Garc{\'{i}}a, Ane and Conde, Angel and Mori, Usue and Lozano, Jose A.},
eprint = {2002.04236},
title = {{A review on outlier/anomaly detection in time series data}},
url = {http://arxiv.org/abs/2002.04236},
year = {2020}
}

@inproceedings{Keogh05,
author = {Keogh, Eamonn and Lin, Jessica and Fu, Ada},
booktitle = {Proceedings - IEEE International Conference on Data Mining, ICDM},
doi = {10.1109/ICDM.2005.79},
isbn = {0769522785},
issn = {15504786},
title = {{HOT SAX: Efficiently finding the most unusual time series subsequence}},
year = {2005}
}

@article{Sharma16,
author = {Sharma, Hemant and Sharma, K K},
doi = {10.1016/j.compbiomed.2016.08.012},
issn = {18790534},
journal = {Computers in Biology and Medicine},
pmid = {27543782},
title = {{An algorithm for sleep apnea detection from single-lead ECG using Hermite basis functions}},
year = {2016}
}

@misc{Penzel03,
author = {Penzel, T.},
booktitle = {European Respiratory Journal},
doi = {10.1183/09031936.03.00102003},
issn = {09031936},
pmid = {14680069},
title = {{Is heart rate variability the simple solution to diagnose sleep apnoea?}},
year = {2003}
}

@article{AlAngari07,
author = {Al-Angari, H. M. and Sahakian, A.V.},
doi = {10.1109/TBME.2006.889772},
issn = {00189294},
journal = {IEEE Transactions on Biomedical Engineering},
pmid = {17926691},
title = {{Use of sample entropy approach to study heart rate variability in obstructive sleep apnea syndrome}},
year = {2007}
}

@article{Bock98,
author = {Bock, Joel and Gough, David A.},
doi = {10.1109/10.725330},
issn = {00189294},
journal = {IEEE Transactions on Biomedical Engineering},
pmid = {9805832},
title = {{Toward prediction of physiological state signals in sleep apnea}},
year = {1998}
}

@article{Song16,
author = {Song, Changyue and Liu, Kaibo and Zhang, Xi and Chen, Lili and Xian, Xiaochen},
doi = {10.1109/TBME.2015.2498199},
issn = {15582531},
journal = {IEEE Transactions on Biomedical Engineering},
pmid = {26560867},
title = {{An Obstructive Sleep Apnea Detection Approach Using a Discriminative Hidden Markov Model from ECG Signals}},
year = {2016}
}

@article{Penzel02,
author = {Penzel, T. and McNames, J. and {De Chazal}, P. and Raymond, B. and Murray, A. and Moody, G.},
doi = {10.1007/BF02345072},
issn = {01400118},
journal = {Medical and Biological Engineering and Computing},
pmid = {12227626},
title = {{Systematic comparison of different algorithms for apnoea detection based on electrocardiogram recordings}},
year = {2002}
}

@article{Boudaoud07,
author = {Boudaoud, S. and Rix, H. and Meste, O. and Heneghan, C. and O'Brien, C.},
doi = {10.1155/2007/32570},
issn = {11108657},
journal = {Eurasip Journal on Advances in Signal Processing},
title = {{Corrected integral shape averaging applied to obstructive sleep apnea detection from the electrocardiogram}},
year = {2007}
}

@article{Abbott16b,
archivePrefix = {arXiv},
arxivId = {1602.03843},
author = {Abbott, B. P. and Abbott, R. and Abbott, T. D. and Abernathy, M. R. and Acernese, F. and Ackley, K. and Adams, C. and Adams, T. and Addesso, P. and Adhikari, R. X. and Adya, V. B. and Affeldt, C. and Agathos, M. and Agatsuma, K. and Aggarwal, N. and Aguiar, O. D. and Aiello, L. and Ain, A. and Ajith, P. and Allen, B. and Allocca, A. and Altin, P. A. and Anderson, S. B. and Anderson, W. G. and Arai, K. and Araya, M. C. and Arceneaux, C. C. and Areeda, J. S. and Arnaud, N. and Arun, K. G. and Ascenzi, S. and Ashton, G. and Ast, M. and Aston, S. M. and Astone, P. and Aufmuth, P. and Aulbert, C. and Babak, S. and Bacon, P. and Bader, M. K.M. and Baker, P. T. and Baldaccini, F. and Ballardin, G. and Ballmer, S. W. and Barayoga, J. C. and Barclay, S. E. and Barish, B. C. and Barker, D. and Barone, F. and Barr, B. and Barsotti, L. and Barsuglia, M. and Barta, D. and Bartlett, J. and Bartos, I. and Bassiri, R. and Basti, A. and Batch, J. C. and Baune, C. and Bavigadda, V. and Bazzan, M. and Behnke, B. and Bejger, M. and Bell, A. S. and Bell, C. J. and Berger, B. K. and Bergman, J. and Bergmann, G. and Berry, C. P.L. and Bersanetti, D. and Bertolini, A. and Betzwieser, J. and Bhagwat, S. and Bhandare, R. and Bilenko, I. A. and Billingsley, G. and Birch, J. and Birney, R. and Biscans, S. and Bisht, A. and Bitossi, M. and Biwer, C. and Bizouard, M. A. and Blackburn, J. K. and Blackburn, L. and Blair, C. D. and Blair, D. G. and Blair, R. M. and Bloemen, S. and Bock, O. and Bodiya, T. P. and Boer, M. and Bogaert, G. and Bogan, C. and Bohe, A. and Bojtos, P. and Bond, C. and Bondu, F. and Bonnand, R. and Boom, B. A. and Bork, R. and Boschi, V. and Bose, S. and Bouffanais, Y. and Bozzi, A. and Bradaschia, C. and Brady, P. R. and Braginsky, V. B. and Branchesi, M. and Brau, J. E. and Briant, T. and Brillet, A. and Brinkmann, M. and Brisson, V. and Brockill, P. and Brooks, A. F. and Brown, D. A. and Brown, D. D. and Brown, N. M. and Buchanan, C. C. and Buikema, A. and Bulik, T. and Bulten, H. J. and Buonanno, A. and Buskulic, D. and Buy, C. and Byer, R. L. and Cadonati, L. and Cagnoli, G. and Cahillane, C. and {Calder{\'{o}}n Bustillo}, J. and Callister, T. and Calloni, E. and Camp, J. B. and Cannon, K. C. and Cao, J. and Capano, C. D. and Capocasa, E. and Carbognani, F. and Caride, S. and {Casanueva Diaz}, J. and Casentini, C. and Caudill, S. and Cavagli{\`{a}}, M. and Cavalier, F. and Cavalieri, R. and Cella, G. and Cepeda, C. B. and {Cerboni Baiardi}, L. and Cerretani, G. and Cesarini, E. and Chakraborty, R. and Chatterji, S. and Chalermsongsak, T. and Chamberlin, S. J. and Chan, M. and Chao, S. and Charlton, P. and Chassande-Mottin, E. and Chen, H. Y. and Chen, Y. and Cheng, C. and Chincarini, A. and Chiummo, A. and Cho, H. S. and Cho, M. and Chow, J. H. and Christensen, N. and Chu, Q. and Chua, S. and Chung, S. and Ciani, G. and Clara, F. and Clark, J. A. and Clark, M. and Cleva, F. and Coccia, E. and Cohadon, P. F. and Colla, A. and Collette, C. G. and Cominsky, L. and Constancio, M. and Conte, A. and Conti, L. and Cook, D. and Corbitt, T. R. and Cornish, N. and Corsi, A. and Cortese, S. and Costa, C. A. and Coughlin, M. W. and Coughlin, S. B. and Coulon, J. P. and Countryman, S. T. and Couvares, P. and Cowan, E. E. and Coward, D. M. and Cowart, M. J. and Coyne, D. C. and Coyne, R. and Craig, K. and Creighton, J. D.E. and Cripe, J. and Crowder, S. G. and Cumming, A. and Cunningham, L. and Cuoco, E. and {Dal Canton}, T. and Danilishin, S. L. and D'Antonio, S. and Danzmann, K. and Darman, N. S. and Dattilo, V. and Dave, I. and Daveloza, H. P. and Davier, M. and Davies, G. S. and Daw, E. J. and Day, R. and Debra, D. and Debreczeni, G. and Degallaix, J. and {De Laurentis}, M. and Del{\'{e}}glise, S. and {Del Pozzo}, W. and Denker, T. and Dent, T. and Dereli, H. and Dergachev, V. and Derosa, R. T. and {De Rosa}, R. and Desalvo, R. and Dhurandhar, S. and D{\'{i}}az, M. C. and {Di Fiore}, L. and {Di Giovanni}, M. and {Di Lieto}, A. and {Di Pace}, S. and {Di Palma}, I. and {Di Virgilio}, A. and Dojcinoski, G. and Dolique, V. and Donovan, F. and Dooley, K. L. and Doravari, S. and Douglas, R. and Downes, T. P. and Drago, M. and Drever, R. W.P. and Driggers, J. C. and Du, Z. and Ducrot, M. and Dwyer, S. E. and Edo, T. B. and Edwards, M. C. and Effler, A. and Eggenstein, H. B. and Ehrens, P. and Eichholz, J. and Eikenberry, S. S. and Engels, W. and Essick, R. C. and Etzel, T. and Evans, M. and Evans, T. M. and Everett, R. and Factourovich, M. and Fafone, V. and Fair, H. and Fairhurst, S. and Fan, X. and Fang, Q. and Farinon, S. and Farr, B. and Farr, W. M. and Favata, M. and Fays, M. and Fehrmann, H. and Fejer, M. M. and Ferrante, I. and Ferreira, E. C. and Ferrini, F. and Fidecaro, F. and Fiori, I. and Fiorucci, D. and Fisher, R. P. and Flaminio, R. and Fletcher, M. and Fournier, J. D. and Franco, S. and Frasca, S. and Frasconi, F. and Frei, Z. and Freise, A. and Frey, R. and Frey, V. and Fricke, T. T. and Fritschel, P. and Frolov, V. V. and Fulda, P. and Fyffe, M. and Gabbard, H. A.G. and Gair, J. R. and Gammaitoni, L. and Gaonkar, S. G. and Garufi, F. and Gatto, A. and Gaur, G. and Gehrels, N. and Gemme, G. and Gendre, B. and Genin, E. and Gennai, A. and George, J. and Gergely, L. and Germain, V. and Ghosh, Archisman and Ghosh, S. and Giaime, J. A. and Giardina, K. D. and Giazotto, A. and Gill, K. and Glaefke, A. and Goetz, E. and Goetz, R. and Gondan, L. and Gonz{\'{a}}lez, G. and {Gonzalez Castro}, J. M. and Gopakumar, A. and Gordon, N. A. and Gorodetsky, M. L. and Gossan, S. E. and Gosselin, M. and Gouaty, R. and Graef, C. and Graff, P. B. and Granata, M. and Grant, A. and Gras, S. and Gray, C. and Greco, G. and Green, A. C. and Groot, P. and Grote, H. and Grunewald, S. and Guidi, G. M. and Guo, X. and Gupta, A. and Gupta, M. K. and Gushwa, K. E. and Gustafson, E. K. and Gustafson, R. and Haas, R. and Hacker, J. J. and Hall, B. R. and Hall, E. D. and Hammond, G. and Haney, M. and Hanke, M. M. and Hanks, J. and Hanna, C. and Hannam, M. D. and Hanson, J. and Hardwick, T. and Harms, J. and Harry, G. M. and Harry, I. W. and Hart, M. J. and Hartman, M. T. and Haster, C. J. and Haughian, K. and Healy, J. and Heidmann, A. and Heintze, M. C. and Heitmann, H. and Hello, P. and Hemming, G. and Hendry, M. and Heng, I. S. and Hennig, J. and Heptonstall, A. W. and Heurs, M. and Hild, S. and Hinder, I. and Hoak, D. and Hodge, K. A. and Hofman, D. and Hollitt, S. E. and Holt, K. and Holz, D. E. and Hopkins, P. and Hosken, D. J. and Hough, J. and Houston, E. A. and Howell, E. J. and Hu, Y. M. and Huang, S. and Huerta, E. A. and Huet, D. and Hughey, B. and Husa, S. and Huttner, S. H. and Huynh-Dinh, T. and Idrisy, A. and Indik, N. and Ingram, D. R. and Inta, R. and Isa, H. N. and Isac, J. M. and Isi, M. and Islas, G. and Isogai, T. and Iyer, B. R. and Izumi, K. and Jacqmin, T. and Jang, H. and Jani, K. and Jaranowski, P. and Jawahar, S. and Jim{\'{e}}nez-Forteza, F. and Johnson, W. W. and Jones, D. I. and Jones, R. and Jonker, R. J.G. and Ju, L. and Haris, K. and Kalaghatgi, C. V. and Kalogera, V. and Kandhasamy, S. and Kang, G. and Kanner, J. B. and Karki, S. and Kasprzack, M. and Katsavounidis, E. and Katzman, W. and Kaufer, S. and Kaur, T. and Kawabe, K. and Kawazoe, F. and K{\'{e}}f{\'{e}}lian, F. and Kehl, M. S. and Keitel, D. and Kelley, D. B. and Kells, W. and Kennedy, R. and Key, J. S. and Khalaidovski, A. and Khalili, F. Y. and Khan, I. and Khan, S. and Khan, Z. and Khazanov, E. A. and Kijbunchoo, N. and Kim, C. and Kim, J. and Kim, K. and Kim, Nam Gyu and Kim, Namjun and Kim, Y. M. and King, E. J. and King, P. J. and Kinsey, M. and Kinzel, D. L. and Kissel, J. S. and Kleybolte, L. and Klimenko, S. and Koehlenbeck, S. M. and Kokeyama, K. and Koley, S. and Kondrashov, V. and Kontos, A. and Korobko, M. and Korth, W. Z. and Kowalska, I. and Kozak, D. B. and Kringel, V. and Kr{\'{o}}lak, A. and Krueger, C. and Kuehn, G. and Kumar, P. and Kuo, L. and Kutynia, A. and Lackey, B. D. and Laguna, P. and Landry, M. and Lange, J. and Lantz, B. and Lasky, P. D. and Lazzarini, A. and Lazzaro, C. and Leaci, P. and Leavey, S. and Lebigot, E. O. and Lee, C. H. and Lee, H. K. and Lee, H. M. and Lee, K. and Lenon, A. and Leonardi, M. and Leong, J. R. and Leroy, N. and Letendre, N. and Levin, Y. and Levine, B. M. and Li, T. G.F. and Libson, A. and Littenberg, T. B. and Lockerbie, N. A. and Logue, J. and Lombardi, A. L. and Lord, J. E. and Lorenzini, M. and Loriette, V. and Lormand, M. and Losurdo, G. and Lough, J. D. and L{\"{u}}ck, H. and Lundgren, A. P. and Luo, J. and Lynch, R. and Ma, Y. and Macdonald, T. and Machenschalk, B. and Macinnis, M. and Macleod, D. M. and Maga{\~{n}}a-Sandoval, F. and Magee, R. M. and Mageswaran, M. and Majorana, E. and Maksimovic, I. and Malvezzi, V. and Man, N. and Mandel, I. and Mandic, V. and Mangano, V. and Mansell, G. L. and Manske, M. and Mantovani, M. and Marchesoni, F. and Marion, F. and M{\'{a}}rka, S. and M{\'{a}}rka, Z. and Markosyan, A. S. and Maros, E. and Martelli, F. and Martellini, L. and Martin, I. W. and Martin, R. M. and Martynov, D. V. and Marx, J. N. and Mason, K. and Masserot, A. and Massinger, T. J. and Masso-Reid, M. and Matichard, F. and Matone, L. and Mavalvala, N. and Mazumder, N. and Mazzolo, G. and McCarthy, R. and McClelland, D. E. and McCormick, S. and McGuire, S. C. and McIntyre, G. and McIver, J. and McManus, D. J. and McWilliams, S. T. and Meacher, D. and Meadors, G. D. and Meidam, J. and Melatos, A. and Mendell, G. and Mendoza-Gandara, D. and Mercer, R. A. and Merilh, E. and Merzougui, M. and Meshkov, S. and Messenger, C. and Messick, C. and Meyers, P. M. and Mezzani, F. and Miao, H. and Michel, C. and Middleton, H. and Mikhailov, E. E. and Milano, L. and Miller, J. and Millhouse, M. and Minenkov, Y. and Ming, J. and Mirshekari, S. and Mishra, C. and Mitra, S. and Mitrofanov, V. P. and Mitselmakher, G. and Mittleman, R. and Moggi, A. and Mohan, M. and Mohapatra, S. R.P. and Montani, M. and Moore, B. C. and Moore, C. J. and Moraru, D. and Moreno, G. and Morriss, S. R. and Mossavi, K. and Mours, B. and Mow-Lowry, C. M. and Mueller, C. L. and Mueller, G. and Muir, A. W. and Mukherjee, Arunava and Mukherjee, D. and Mukherjee, S. and Mukund, N. and Mullavey, A. and Munch, J. and Murphy, D. J. and Murray, P. G. and Mytidis, A. and Nardecchia, I. and Naticchioni, L. and Nayak, R. K. and Necula, V. and Nedkova, K. and Nelemans, G. and Neri, M. and Neunzert, A. and Newton, G. and Nguyen, T. T. and Nielsen, A. B. and Nissanke, S. and Nitz, A. and Nocera, F. and Nolting, D. and Normandin, M. E. and Nuttall, L. K. and Oberling, J. and Ochsner, E. and O'Dell, J. and Oelker, E. and Ogin, G. H. and Oh, J. J. and Oh, S. H. and Ohme, F. and Oliver, M. and Oppermann, P. and Oram, Richard J. and O'Reilly, B. and O'Shaughnessy, R. and Ottaway, D. J. and Ottens, R. S. and Overmier, H. and Owen, B. J. and Pai, A. and Pai, S. A. and Palamos, J. R. and Palashov, O. and Palomba, C. and Pal-Singh, A. and Pan, H. and Pankow, C. and Pannarale, F. and Pant, B. C. and Paoletti, F. and Paoli, A. and Papa, M. A. and Page, J. and Paris, H. R. and Parker, W. and Pascucci, D. and Pasqualetti, A. and Passaquieti, R. and Passuello, D. and Patricelli, B. and Patrick, Z. and Pearlstone, B. L. and Pedraza, M. and Pedurand, R. and Pekowsky, L. and Pele, A. and Penn, S. and Perreca, A. and Phelps, M. and Piccinni, O. and Pichot, M. and Piergiovanni, F. and Pierro, V. and Pillant, G. and Pinard, L. and Pinto, I. M. and Pitkin, M. and Poggiani, R. and Popolizio, P. and Post, A. and Powell, J. and Prasad, J. and Predoi, V. and Premachandra, S. S. and Prestegard, T. and Price, L. R. and Prijatelj, M. and Principe, M. and Privitera, S. and Prodi, G. A. and Prokhorov, L. and Puncken, O. and Punturo, M. and Puppo, P. and P{\"{u}}rrer, M. and Qi, H. and Qin, J. and Quetschke, V. and Quintero, E. A. and Quitzow-James, R. and Raab, F. J. and Rabeling, D. S. and Radkins, H. and Raffai, P. and Raja, S. and Rakhmanov, M. and Rapagnani, P. and Raymond, V. and Razzano, M. and Re, V. and Read, J. and Reed, C. M. and Regimbau, T. and Rei, L. and Reid, S. and Reitze, D. H. and Rew, H. and Reyes, S. D. and Ricci, F. and Riles, K. and Robertson, N. A. and Robie, R. and Robinet, F. and Rocchi, A. and Rolland, L. and Rollins, J. G. and Roma, V. J. and Romano, R. and Romanov, G. and Romie, J. H. and Rosi{\'{n}}ska, D. and Rowan, S. and R{\"{u}}diger, A. and Ruggi, P. and Ryan, K. and Sachdev, S. and Sadecki, T. and Sadeghian, L. and Salconi, L. and Saleem, M. and Salemi, F. and Samajdar, A. and Sammut, L. and Sanchez, E. J. and Sandberg, V. and Sandeen, B. and Sanders, J. R. and Sassolas, B. and Sathyaprakash, B. S. and Saulson, P. R. and Sauter, O. and Savage, R. L. and Sawadsky, A. and Schale, P. and Schilling, R. and Schmidt, J. and Schmidt, P. and Schnabel, R. and Schofield, R. M.S. and Sch{\"{o}}nbeck, A. and Schreiber, E. and Schuette, D. and Schutz, B. F. and Scott, J. and Scott, S. M. and Sellers, D. and Sengupta, A. S. and Sentenac, D. and Sequino, V. and Sergeev, A. and Serna, G. and Setyawati, Y. and Sevigny, A. and Shaddock, D. A. and Shah, S. and Shahriar, M. S. and Shaltev, M. and Shao, Z. and Shapiro, B. and Shawhan, P. and Sheperd, A. and Shoemaker, D. H. and Shoemaker, D. M. and Siellez, K. and Siemens, X. and Sigg, D. and Silva, A. D. and Simakov, D. and Singer, A. and Singer, L. P. and Singh, A. and Singh, R. and Singhal, A. and Sintes, A. M. and Slagmolen, B. J.J. and Smith, J. R. and Smith, N. D. and Smith, R. J.E. and Son, E. J. and Sorazu, B. and Sorrentino, F. and Souradeep, T. and Srivastava, A. K. and Staley, A. and Steinke, M. and Steinlechner, J. and Steinlechner, S. and Steinmeyer, D. and Stephens, B. C. and Stone, R. and Strain, K. A. and Straniero, N. and Stratta, G. and Strauss, N. A. and Strigin, S. and Sturani, R. and Stuver, A. L. and Summerscales, T. Z. and Sun, L. and Sutton, P. J. and Swinkels, B. L. and Szczepa{\'{n}}czyk, M. J. and Tacca, M. and Talukder, D. and Tanner, D. B. and T{\'{a}}pai, M. and Tarabrin, S. P. and Taracchini, A. and Taylor, R. and Theeg, T. and Thirugnanasambandam, M. P. and Thomas, E. G. and Thomas, M. and Thomas, P. and Thorne, K. A. and Thorne, K. S. and Thrane, E. and Tiwari, S. and Tiwari, V. and Tokmakov, K. V. and Tomlinson, C. and Tonelli, M. and Torres, C. V. and Torrie, C. I. and T{\"{o}}yr{\"{a}}, D. and Travasso, F. and Traylor, G. and Trifir{\`{o}}, D. and Tringali, M. C. and Trozzo, L. and Tse, M. and Turconi, M. and Tuyenbayev, D. and Ugolini, D. and Unnikrishnan, C. S. and Urban, A. L. and Usman, S. A. and Vahlbruch, H. and Vajente, G. and Valdes, G. and {Van Bakel}, N. and {Van Beuzekom}, M. and {Van Den Brand}, J. F.J. and {Van Den Broeck}, C. and Vander-Hyde, D. C. and {Van Der Schaaf}, L. and {Van Heijningen}, J. V. and {Van Veggel}, A. A. and Vardaro, M. and Vass, S. and Vas{\'{u}}th, M. and Vaulin, R. and Vecchio, A. and Vedovato, G. and Veitch, J. and Veitch, P. J. and Venkateswara, K. and Verkindt, D. and Vetrano, F. and Vicer{\'{e}}, A. and Vinciguerra, S. and Vine, D. J. and Vinet, J. Y. and Vitale, S. and Vo, T. and Vocca, H. and Vorvick, C. and Voss, D. and Vousden, W. D. and Vyatchanin, S. P. and Wade, A. R. and Wade, L. E. and Wade, M. and Walker, M. and Wallace, L. and Walsh, S. and Wang, G. and Wang, H. and Wang, M. and Wang, X. and Wang, Y. and Ward, R. L. and Warner, J. and Was, M. and Weaver, B. and Wei, L. W. and Weinert, M. and Weinstein, A. J. and Weiss, R. and Welborn, T. and Wen, L. and We{\ss}els, P. and Westphal, T. and Wette, K. and Whelan, J. T. and White, D. J. and Whiting, B. F. and Williams, D. and Williams, R. D. and Williamson, A. R. and Willis, J. L. and Willke, B. and Wimmer, M. H. and Winkler, W. and Wipf, C. C. and Wittel, H. and Woan, G. and Worden, J. and Wright, J. L. and Wu, G. and Yablon, J. and Yam, W. and Yamamoto, H. and Yancey, C. C. and Yap, M. J. and Yu, H. and Yvert, M. and Zadrozny, A. and Zangrando, L. and Zanolin, M. and Zendri, J. P. and Zevin, M. and Zhang, F. and Zhang, L. and Zhang, M. and Zhang, Y. and Zhao, C. and Zhou, M. and Zhou, Z. and Zhu, X. J. and Zucker, M. E. and Zuraw, S. E. and Zweizig, J.},
doi = {10.1103/PhysRevD.93.122004},
eprint = {1602.03843},
issn = {24700029},
journal = {Physical Review D},
title = {{Observing gravitational-wave transient GW150914 with minimal assumptions}},
year = {2016}
}

@article{Ahmed16,
author = {Ahmed, Mohiuddin and Mahmood, Abdun Naser and Islam, Md Rafiqul},
doi = {10.1016/j.future.2015.01.001},
issn = {0167739X},
journal = {Future Generation Computer Systems},
title = {{A survey of anomaly detection techniques in financial domain}},
year = {2016}
}

@inproceedings{Yeh17,
author = {Yeh, Chin Chia Michael and Zhu, Yan and Ulanova, Liudmila and Begum, Nurjahan and Ding, Yifei and Dau, Hoang Anh and Silva, Diego Furtado and Mueen, Abdullah and Keogh, Eamonn},
booktitle = {Proceedings - IEEE International Conference on Data Mining, ICDM},
doi = {10.1109/ICDM.2016.89},
isbn = {9781509054725},
issn = {15504786},
title = {{Matrix profile I: All pairs similarity joins for time series: A unifying view that includes motifs, discords and shapelets}},
year = {2017}
}

@inproceedings{Senin14,
author = {Senin, Pavel and Lin, Jessica and Wang, Xing and Oates, Tim and Gandhi, Sunil and Boedihardjo, Arnold P. and Chen, Crystal and Frankenstein, Susan and Lerner, Manfred},
booktitle = {Lecture Notes in Computer Science (including subseries Lecture Notes in Artificial Intelligence and Lecture Notes in Bioinformatics)},
doi = {10.1007/978-3-662-44845-8_37},
isbn = {9783662448441},
issn = {16113349},
title = {{GrammarViz 2.0: A tool for grammar-based pattern discovery in time series}},
year = {2014}
}

@article{Chandola09,
archivePrefix = {arXiv},
arxivId = {arXiv:1011.1669v3},
author = {Chandola, Varun and Banerjee, Arindam and Kumar, Vipin},
doi = {10.1145/1541880.1541882},
eprint = {arXiv:1011.1669v3},
isbn = {0361-7734},
issn = {03600300},
journal = {ACM Computing Surveys},
number = {3},
pages = {1--58},
pmid = {21834704},
title = {{Anomaly detection: A survey}},
url = {http://portal.acm.org/citation.cfm?doid=1541880.1541882},
volume = {41},
year = {2009}
}

@article{Pimentel24,
author = {Pimentel, M A F and Clifton, D A and Clifton, L and Tarassenko, L},
doi = {10.1016/j.sigpro.2013.12.026},
issn = {01651684},
journal = {Signal Processing},
pages = {215--249},
title = {{A review of novelty detection}},
url = {https://www.scopus.com/inward/record.uri?eid=2-s2.0-84893296219{\&}doi=10.1016{\%}2Fj.sigpro.2013.12.026{\&}partnerID=40{\&}md5=0d3380f6f44195f22684bd2e83d06cd8},
volume = {99},
year = {2014}
}

@article{Abbott16,
author = {Abbott, B. P. and Abbott, R. and Abbott, T. D. and Abernathy, M. R. and Acernese, F. and Ackley, K. and Adams, C. and Adams, T. and Addesso, P. and Adhikari, R. X. and Adya, V. B. and Affeldt, C. and Agathos, M. and Agatsuma, K. and Aggarwal, N. and Aguiar, O. D. and Aiello, L. and Ain, A. and Ajith, P. and Allen, B. and Allocca, A. and Altin, P. A. and Anderson, S. B. and Anderson, W. G. and Arai, K. and Arain, M. A. and Araya, M. C. and Arceneaux, C. C. and Areeda, J. S. and Arnaud, N. and Arun, K. G. and Ascenzi, S. and Ashton, G. and Ast, M. and Aston, S. M. and Astone, P. and Aufmuth, P. and Aulbert, C. and Babak, S. and Bacon, P. and Bader, M. K. M. and Baker, P. T. and Baldaccini, F. and Ballardin, G. and Ballmer, S. W. and Barayoga, J. C. and Barclay, S. E. and Barish, B. C. and Barker, D. and Barone, F. and Barr, B. and Barsotti, L. and Barsuglia, M. and Barta, D. and Bartlett, J. and Barton, M. A. and Bartos, I. and Bassiri, R. and Basti, A. and Batch, J. C. and Baune, C. and Bavigadda, V. and Bazzan, M. and Behnke, B. and Bejger, M. and Belczynski, C. and Bell, A. S. and Bell, C. J. and Berger, B. K. and Bergman, J. and Bergmann, G. and Berry, C. P. L. and Bersanetti, D. and Bertolini, A. and Betzwieser, J. and Bhagwat, S. and Bhandare, R. and Bilenko, I. A. and Billingsley, G. and Birch, J. and Birney, R. and Birnholtz, O. and Biscans, S. and Bisht, A. and Bitossi, M. and Biwer, C. and Bizouard, M. A. and Blackburn, J. K. and Blair, C. D. and Blair, D. G. and Blair, R. M. and Bloemen, S. and Bock, O. and Bodiya, T. P. and Boer, M. and Bogaert, G. and Bogan, C. and Bohe, A. and Bojtos, P. and Bond, C. and Bondu, F. and Bonnand, R. and Boom, B. A. and Bork, R. and Boschi, V. and Bose, S. and Bouffanais, Y. and Bozzi, A. and Bradaschia, C. and Brady, P. R. and Braginsky, V. B. and Branchesi, M. and Brau, J. E. and Briant, T. and Brillet, A. and Brinkmann, M. and Brisson, V. and Brockill, P. and Brooks, A. F. and Brown, D. A. and Brown, D. D. and Brown, N. M. and Buchanan, C. C. and Buikema, A. and Bulik, T. and Bulten, H. J. and Buonanno, A. and Buskulic, D. and Buy, C. and Byer, R. L. and Cabero, M. and Cadonati, L. and Cagnoli, G. and Cahillane, C. and Bustillo, J. Calder{\'{o}}n and Callister, T. and Calloni, E. and Camp, J. B. and Cannon, K. C. and Cao, J. and Capano, C. D. and Capocasa, E. and Carbognani, F. and Caride, S. and Diaz, J. Casanueva and Casentini, C. and Caudill, S. and Cavagli{\`{a}}, M. and Cavalier, F. and Cavalieri, R. and Cella, G. and Cepeda, C. B. and Baiardi, L. Cerboni and Cerretani, G. and Cesarini, E. and Chakraborty, R. and Chalermsongsak, T. and Chamberlin, S. J. and Chan, M. and Chao, S. and Charlton, P. and Chassande-Mottin, E. and Chen, H. Y. and Chen, Y. and Cheng, C. and Chincarini, A. and Chiummo, A. and Cho, H. S. and Cho, M. and Chow, J. H. and Christensen, N. and Chu, Q. and Chua, S. and Chung, S. and Ciani, G. and Clara, F. and Clark, J. A. and Cleva, F. and Coccia, E. and Cohadon, P.-F. and Colla, A. and Collette, C. G. and Cominsky, L. and Constancio, M. and Conte, A. and Conti, L. and Cook, D. and Corbitt, T. R. and Cornish, N. and Corsi, A. and Cortese, S. and Costa, C. A. and Coughlin, M. W. and Coughlin, S. B. and Coulon, J.-P. and Countryman, S. T. and Couvares, P. and Cowan, E. E. and Coward, D. M. and Cowart, M. J. and Coyne, D. C. and Coyne, R. and Craig, K. and Creighton, J. D. E. and Creighton, T. D. and Cripe, J. and Crowder, S. G. and Cruise, A. M. and Cumming, A. and Cunningham, L. and Cuoco, E. and Canton, T. Dal and Danilishin, S. L. and D'Antonio, S. and Danzmann, K. and Darman, N. S. and {Da Silva Costa}, C. F. and Dattilo, V. and Dave, I. and Daveloza, H. P. and Davier, M. and Davies, G. S. and Daw, E. J. and Day, R. and De, S. and DeBra, D. and Debreczeni, G. and Degallaix, J. and {De Laurentis}, M. and Del{\'{e}}glise, S. and {Del Pozzo}, W. and Denker, T. and Dent, T. and Dereli, H. and Dergachev, V. and DeRosa, R. T. and {De Rosa}, R. and DeSalvo, R. and Dhurandhar, S. and D{\'{i}}az, M. C. and {Di Fiore}, L. and {Di Giovanni}, M. and {Di Lieto}, A. and {Di Pace}, S. and {Di Palma}, I. and {Di Virgilio}, A. and Dojcinoski, G. and Dolique, V. and Donovan, F. and Dooley, K. L. and Doravari, S. and Douglas, R. and Downes, T. P. and Drago, M. and Drever, R. W. P. and Driggers, J. C. and Du, Z. and Ducrot, M. and Dwyer, S. E. and Edo, T. B. and Edwards, M. C. and Effler, A. and Eggenstein, H.-B. and Ehrens, P. and Eichholz, J. and Eikenberry, S. S. and Engels, W. and Essick, R. C. and Etzel, T. and Evans, M. and Evans, T. M. and Everett, R. and Factourovich, M. and Fafone, V. and Fair, H. and Fairhurst, S. and Fan, X. and Fang, Q. and Farinon, S. and Farr, B. and Farr, W. M. and Favata, M. and Fays, M. and Fehrmann, H. and Fejer, M. M. and Feldbaum, D. and Ferrante, I. and Ferreira, E. C. and Ferrini, F. and Fidecaro, F. and Finn, L. S. and Fiori, I. and Fiorucci, D. and Fisher, R. P. and Flaminio, R. and Fletcher, M. and Fong, H. and Fournier, J.-D. and Franco, S. and Frasca, S. and Frasconi, F. and Frede, M. and Frei, Z. and Freise, A. and Frey, R. and Frey, V. and Fricke, T. T. and Fritschel, P. and Frolov, V. V. and Fulda, P. and Fyffe, M. and Gabbard, H. A. G. and Gair, J. R. and Gammaitoni, L. and Gaonkar, S. G. and Garufi, F. and Gatto, A. and Gaur, G. and Gehrels, N. and Gemme, G. and Gendre, B. and Genin, E. and Gennai, A. and George, J. and Gergely, L. and Germain, V. and Ghosh, Abhirup and Ghosh, Archisman and Ghosh, S. and Giaime, J. A. and Giardina, K. D. and Giazotto, A. and Gill, K. and Glaefke, A. and Gleason, J. R. and Goetz, E. and Goetz, R. and Gondan, L. and Gonz{\'{a}}lez, G. and Castro, J. M. Gonzalez and Gopakumar, A. and Gordon, N. A. and Gorodetsky, M. L. and Gossan, S. E. and Gosselin, M. and Gouaty, R. and Graef, C. and Graff, P. B. and Granata, M. and Grant, A. and Gras, S. and Gray, C. and Greco, G. and Green, A. C. and Greenhalgh, R. J. S. and Groot, P. and Grote, H. and Grunewald, S. and Guidi, G. M. and Guo, X. and Gupta, A. and Gupta, M. K. and Gushwa, K. E. and Gustafson, E. K. and Gustafson, R. and Hacker, J. J. and Hall, B. R. and Hall, E. D. and Hammond, G. and Haney, M. and Hanke, M. M. and Hanks, J. and Hanna, C. and Hannam, M. D. and Hanson, J. and Hardwick, T. and Harms, J. and Harry, G. M. and Harry, I. W. and Hart, M. J. and Hartman, M. T. and Haster, C.-J. and Haughian, K. and Healy, J. and Heefner, J. and Heidmann, A. and Heintze, M. C. and Heinzel, G. and Heitmann, H. and Hello, P. and Hemming, G. and Hendry, M. and Heng, I. S. and Hennig, J. and Heptonstall, A. W. and Heurs, M. and Hild, S. and Hoak, D. and Hodge, K. A. and Hofman, D. and Hollitt, S. E. and Holt, K. and Holz, D. E. and Hopkins, P. and Hosken, D. J. and Hough, J. and Houston, E. A. and Howell, E. J. and Hu, Y. M. and Huang, S. and Huerta, E. A. and Huet, D. and Hughey, B. and Husa, S. and Huttner, S. H. and Huynh-Dinh, T. and Idrisy, A. and Indik, N. and Ingram, D. R. and Inta, R. and Isa, H. N. and Isac, J.-M. and Isi, M. and Islas, G. and Isogai, T. and Iyer, B. R. and Izumi, K. and Jacobson, M. B. and Jacqmin, T. and Jang, H. and Jani, K. and Jaranowski, P. and Jawahar, S. and Jim{\'{e}}nez-Forteza, F. and Johnson, W. W. and Johnson-McDaniel, N. K. and Jones, D. I. and Jones, R. and Jonker, R. J. G. and Ju, L. and Haris, K. and Kalaghatgi, C. V. and Kalogera, V. and Kandhasamy, S. and Kang, G. and Kanner, J. B. and Karki, S. and Kasprzack, M. and Katsavounidis, E. and Katzman, W. and Kaufer, S. and Kaur, T. and Kawabe, K. and Kawazoe, F. and K{\'{e}}f{\'{e}}lian, F. and Kehl, M. S. and Keitel, D. and Kelley, D. B. and Kells, W. and Kennedy, R. and Keppel, D. G. and Key, J. S. and Khalaidovski, A. and Khalili, F. Y. and Khan, I. and Khan, S. and Khan, Z. and Khazanov, E. A. and Kijbunchoo, N. and Kim, C. and Kim, J. and Kim, K. and Kim, Nam-Gyu and Kim, Namjun and Kim, Y.-M. and King, E. J. and King, P. J. and Kinzel, D. L. and Kissel, J. S. and Kleybolte, L. and Klimenko, S. and Koehlenbeck, S. M. and Kokeyama, K. and Koley, S. and Kondrashov, V. and Kontos, A. and Koranda, S. and Korobko, M. and Korth, W. Z. and Kowalska, I. and Kozak, D. B. and Kringel, V. and Krishnan, B. and Kr{\'{o}}lak, A. and Krueger, C. and Kuehn, G. and Kumar, P. and Kumar, R. and Kuo, L. and Kutynia, A. and Kwee, P. and Lackey, B. D. and Landry, M. and Lange, J. and Lantz, B. and Lasky, P. D. and Lazzarini, A. and Lazzaro, C. and Leaci, P. and Leavey, S. and Lebigot, E. O. and Lee, C. H. and Lee, H. K. and Lee, H. M. and Lee, K. and Lenon, A. and Leonardi, M. and Leong, J. R. and Leroy, N. and Letendre, N. and Levin, Y. and Levine, B. M. and Li, T. G. F. and Libson, A. and Littenberg, T. B. and Lockerbie, N. A. and Logue, J. and Lombardi, A. L. and London, L. T. and Lord, J. E. and Lorenzini, M. and Loriette, V. and Lormand, M. and Losurdo, G. and Lough, J. D. and Lousto, C. O. and Lovelace, G. and L{\"{u}}ck, H. and Lundgren, A. P. and Luo, J. and Lynch, R. and Ma, Y. and MacDonald, T. and Machenschalk, B. and MacInnis, M. and Macleod, D. M. and Maga{\~{n}}a-Sandoval, F. and Magee, R. M. and Mageswaran, M. and Majorana, E. and Maksimovic, I. and Malvezzi, V. and Man, N. and Mandel, I. and Mandic, V. and Mangano, V. and Mansell, G. L. and Manske, M. and Mantovani, M. and Marchesoni, F. and Marion, F. and M{\'{a}}rka, S. and M{\'{a}}rka, Z. and Markosyan, A. S. and Maros, E. and Martelli, F. and Martellini, L. and Martin, I. W. and Martin, R. M. and Martynov, D. V. and Marx, J. N. and Mason, K. and Masserot, A. and Massinger, T. J. and Masso-Reid, M. and Matichard, F. and Matone, L. and Mavalvala, N. and Mazumder, N. and Mazzolo, G. and McCarthy, R. and McClelland, D. E. and McCormick, S. and McGuire, S. C. and McIntyre, G. and McIver, J. and McManus, D. J. and McWilliams, S. T. and Meacher, D. and Meadors, G. D. and Meidam, J. and Melatos, A. and Mendell, G. and Mendoza-Gandara, D. and Mercer, R. A. and Merilh, E. and Merzougui, M. and Meshkov, S. and Messenger, C. and Messick, C. and Meyers, P. M. and Mezzani, F. and Miao, H. and Michel, C. and Middleton, H. and Mikhailov, E. E. and Milano, L. and Miller, J. and Millhouse, M. and Minenkov, Y. and Ming, J. and Mirshekari, S. and Mishra, C. and Mitra, S. and Mitrofanov, V. P. and Mitselmakher, G. and Mittleman, R. and Moggi, A. and Mohan, M. and Mohapatra, S. R. P. and Montani, M. and Moore, B. C. and Moore, C. J. and Moraru, D. and Moreno, G. and Morriss, S. R. and Mossavi, K. and Mours, B. and Mow-Lowry, C. M. and Mueller, C. L. and Mueller, G. and Muir, A. W. and Mukherjee, Arunava and Mukherjee, D. and Mukherjee, S. and Mukund, N. and Mullavey, A. and Munch, J. and Murphy, D. J. and Murray, P. G. and Mytidis, A. and Nardecchia, I. and Naticchioni, L. and Nayak, R. K. and Necula, V. and Nedkova, K. and Nelemans, G. and Neri, M. and Neunzert, A. and Newton, G. and Nguyen, T. T. and Nielsen, A. B. and Nissanke, S. and Nitz, A. and Nocera, F. and Nolting, D. and Normandin, M. E. N. and Nuttall, L. K. and Oberling, J. and Ochsner, E. and O'Dell, J. and Oelker, E. and Ogin, G. H. and Oh, J. J. and Oh, S. H. and Ohme, F. and Oliver, M. and Oppermann, P. and Oram, Richard J. and O'Reilly, B. and O'Shaughnessy, R. and Ott, C. D. and Ottaway, D. J. and Ottens, R. S. and Overmier, H. and Owen, B. J. and Pai, A. and Pai, S. A. and Palamos, J. R. and Palashov, O. and Palomba, C. and Pal-Singh, A. and Pan, H. and Pan, Y. and Pankow, C. and Pannarale, F. and Pant, B. C. and Paoletti, F. and Paoli, A. and Papa, M. A. and Paris, H. R. and Parker, W. and Pascucci, D. and Pasqualetti, A. and Passaquieti, R. and Passuello, D. and Patricelli, B. and Patrick, Z. and Pearlstone, B. L. and Pedraza, M. and Pedurand, R. and Pekowsky, L. and Pele, A. and Penn, S. and Perreca, A. and Pfeiffer, H. P. and Phelps, M. and Piccinni, O. and Pichot, M. and Pickenpack, M. and Piergiovanni, F. and Pierro, V. and Pillant, G. and Pinard, L. and Pinto, I. M. and Pitkin, M. and Poeld, J. H. and Poggiani, R. and Popolizio, P. and Post, A. and Powell, J. and Prasad, J. and Predoi, V. and Premachandra, S. S. and Prestegard, T. and Price, L. R. and Prijatelj, M. and Principe, M. and Privitera, S. and Prix, R. and Prodi, G. A. and Prokhorov, L. and Puncken, O. and Punturo, M. and Puppo, P. and P{\"{u}}rrer, M. and Qi, H. and Qin, J. and Quetschke, V. and Quintero, E. A. and Quitzow-James, R. and Raab, F. J. and Rabeling, D. S. and Radkins, H. and Raffai, P. and Raja, S. and Rakhmanov, M. and Ramet, C. R. and Rapagnani, P. and Raymond, V. and Razzano, M. and Re, V. and Read, J. and Reed, C. M. and Regimbau, T. and Rei, L. and Reid, S. and Reitze, D. H. and Rew, H. and Reyes, S. D. and Ricci, F. and Riles, K. and Robertson, N. A. and Robie, R. and Robinet, F. and Rocchi, A. and Rolland, L. and Rollins, J. G. and Roma, V. J. and Romano, J. D. and Romano, R. and Romanov, G. and Romie, J. H. and Rosi{\'{n}}ska, D. and Rowan, S. and R{\"{u}}diger, A. and Ruggi, P. and Ryan, K. and Sachdev, S. and Sadecki, T. and Sadeghian, L. and Salconi, L. and Saleem, M. and Salemi, F. and Samajdar, A. and Sammut, L. and Sampson, L. M. and Sanchez, E. J. and Sandberg, V. and Sandeen, B. and Sanders, G. H. and Sanders, J. R. and Sassolas, B. and Sathyaprakash, B. S. and Saulson, P. R. and Sauter, O. and Savage, R. L. and Sawadsky, A. and Schale, P. and Schilling, R. and Schmidt, J. and Schmidt, P. and Schnabel, R. and Schofield, R. M. S. and Sch{\"{o}}nbeck, A. and Schreiber, E. and Schuette, D. and Schutz, B. F. and Scott, J. and Scott, S. M. and Sellers, D. and Sengupta, A. S. and Sentenac, D. and Sequino, V. and Sergeev, A. and Serna, G. and Setyawati, Y. and Sevigny, A. and Shaddock, D. A. and Shaffer, T. and Shah, S. and Shahriar, M. S. and Shaltev, M. and Shao, Z. and Shapiro, B. and Shawhan, P. and Sheperd, A. and Shoemaker, D. H. and Shoemaker, D. M. and Siellez, K. and Siemens, X. and Sigg, D. and Silva, A. D. and Simakov, D. and Singer, A. and Singer, L. P. and Singh, A. and Singh, R. and Singhal, A. and Sintes, A. M. and Slagmolen, B. J. J. and Smith, J. R. and Smith, M. R. and Smith, N. D. and Smith, R. J. E. and Son, E. J. and Sorazu, B. and Sorrentino, F. and Souradeep, T. and Srivastava, A. K. and Staley, A. and Steinke, M. and Steinlechner, J. and Steinlechner, S. and Steinmeyer, D. and Stephens, B. C. and Stevenson, S. P. and Stone, R. and Strain, K. A. and Straniero, N. and Stratta, G. and Strauss, N. A. and Strigin, S. and Sturani, R. and Stuver, A. L. and Summerscales, T. Z. and Sun, L. and Sutton, P. J. and Swinkels, B. L. and Szczepa{\'{n}}czyk, M. J. and Tacca, M. and Talukder, D. and Tanner, D. B. and T{\'{a}}pai, M. and Tarabrin, S. P. and Taracchini, A. and Taylor, R. and Theeg, T. and Thirugnanasambandam, M. P. and Thomas, E. G. and Thomas, M. and Thomas, P. and Thorne, K. A. and Thorne, K. S. and Thrane, E. and Tiwari, S. and Tiwari, V. and Tokmakov, K. V. and Tomlinson, C. and Tonelli, M. and Torres, C. V. and Torrie, C. I. and T{\"{o}}yr{\"{a}}, D. and Travasso, F. and Traylor, G. and Trifir{\`{o}}, D. and Tringali, M. C. and Trozzo, L. and Tse, M. and Turconi, M. and Tuyenbayev, D. and Ugolini, D. and Unnikrishnan, C. S. and Urban, A. L. and Usman, S. A. and Vahlbruch, H. and Vajente, G. and Valdes, G. and Vallisneri, M. and van Bakel, N. and van Beuzekom, M. and van den Brand, J. F. J. and {Van Den Broeck}, C. and Vander-Hyde, D. C. and van der Schaaf, L. and van Heijningen, J. V. and van Veggel, A. A. and Vardaro, M. and Vass, S. and Vas{\'{u}}th, M. and Vaulin, R. and Vecchio, A. and Vedovato, G. and Veitch, J. and Veitch, P. J. and Venkateswara, K. and Verkindt, D. and Vetrano, F. and Vicer{\'{e}}, A. and Vinciguerra, S. and Vine, D. J. and Vinet, J.-Y. and Vitale, S. and Vo, T. and Vocca, H. and Vorvick, C. and Voss, D. and Vousden, W. D. and Vyatchanin, S. P. and Wade, A. R. and Wade, L. E. and Wade, M. and Waldman, S. J. and Walker, M. and Wallace, L. and Walsh, S. and Wang, G. and Wang, H. and Wang, M. and Wang, X. and Wang, Y. and Ward, H. and Ward, R. L. and Warner, J. and Was, M. and Weaver, B. and Wei, L.-W. and Weinert, M. and Weinstein, A. J. and Weiss, R. and Welborn, T. and Wen, L. and We{\ss}els, P. and Westphal, T. and Wette, K. and Whelan, J. T. and Whitcomb, S. E. and White, D. J. and Whiting, B. F. and Wiesner, K. and Wilkinson, C. and Willems, P. A. and Williams, L. and Williams, R. D. and Williamson, A. R. and Willis, J. L. and Willke, B. and Wimmer, M. H. and Winkelmann, L. and Winkler, W. and Wipf, C. C. and Wiseman, A. G. and Wittel, H. and Woan, G. and Worden, J. and Wright, J. L. and Wu, G. and Yablon, J. and Yakushin, I. and Yam, W. and Yamamoto, H. and Yancey, C. C. and Yap, M. J. and Yu, H. and Yvert, M. and Zadro{\.{z}}ny, A. and Zangrando, L. and Zanolin, M. and Zendri, J.-P. and Zevin, M. and Zhang, F. and Zhang, L. and Zhang, M. and Zhang, Y. and Zhao, C. and Zhou, M. and Zhou, Z. and Zhu, X. J. and Zucker, M. E. and Zuraw, S. E. and Zweizig, J.},
doi = {10.1103/PhysRevLett.116.061102},
issn = {0031-9007},
journal = {Physical Review Letters},
month = {feb},
number = {6},
pages = {061102},
title = {{Observation of Gravitational Waves from a Binary Black Hole Merger}},
url = {https://link.aps.org/doi/10.1103/PhysRevLett.116.061102},
volume = {116},
year = {2016}
}

@article{Hodge04,
author = {Hodge, Victoria J and Austin, Jim},
doi = {10.1007/s10462-004-4304-y},
issn = {1573-7462},
journal = {Artificial Intelligence Review},
number = {2},
pages = {85--126},
title = {{A Survey of Outlier Detection Methodologies}},
url = {https://doi.org/10.1007/s10462-004-4304-y},
volume = {22},
year = {2004}
}

@article{Takens81,
archivePrefix = {arXiv},
arxivId = {arXiv:1011.1669v3},
author = {Takens, Floris},
doi = {10.1007/BFb0091924},
eprint = {arXiv:1011.1669v3},
isbn = {978-3-540-11171-9},
issn = {1098-6596},
journal = {Dynamical Systems and Turbulence, Warwick 1980},
pages = {366--381},
pmid = {155},
title = {{Detecting strange attractors in turbulence}},
url = {http://link.springer.com/10.1007/BFb0091924},
volume = {898},
year = {1981}
}

@article{Ye15,
author = {Ye, Hao and Beamish, Richard J. and Glaser, Sarah M. and Grant, Sue C.H. and Hsieh, Chih Hao and Richards, Laura J. and Schnute, Jon T. and Sugihara, George},
doi = {10.1073/pnas.1417063112},
issn = {10916490},
journal = {Proceedings of the National Academy of Sciences of the United States of America},
number = {13},
pages = {E1569--E1576},
title = {{Equation-free mechanistic ecosystem forecasting using empirical dynamic modeling}},
volume = {112},
year = {2015}
}

@article{Schreiber96,
author = {Schreiber, Thomas and Kaplan, Daniel T.},
doi = {10.1063/1.166148},
issn = {1054-1500},
journal = {Chaos: An Interdisciplinary Journal of Nonlinear Science},
month = {mar},
number = {1},
pages = {87--92},
title = {{Nonlinear noise reduction for electrocardiograms}},
url = {http://aip.scitation.org/doi/10.1063/1.166148},
volume = {6},
year = {1996}
}

@article{Sornette09,
author = {Didier Sornette},
journal = {International Journal of Terraspace Science and Engineering},
number = {1},
pages = {1-18},
title = {{Dragon-Kings, Black Swans and the Prediction of Crises}},
url = {https://arxiv.org/abs/0907.4290},
volume = {2},
year = {2009}
}

@article{Hamilton16,
author = {Hamilton, Franz and Berry, Tyrus and Sauer, Timothy},
doi = {10.1103/PhysRevX.6.011021},
issn = {21603308},
journal = {Physical Review X},
title = {{Ensemble Kalman filtering without a model}},
year = {2016}
}

@article{Sugihara12,
author = {Sugihara, George and May, Robert and Ye, Hao and Hsieh, Chih-hao and Deyle, Ethan and Fogarty, Michael and Munch, Stephan},
doi = {10.1126/science.1227079},
issn = {1095-9203},
journal = {Science (New York, N.Y.)},
month = {oct},
number = {6106},
pages = {496--500},
pmid = {22997134},
title = {{Detecting causality in complex ecosystems.}},
volume = {338},
year = {2012}
}

@article{Gao13,
author = {Gao, Jianbo and Hu, Jing},
doi = {10.3389/fncom.2013.00122},
journal = {Frontiers in Computational Neuroscience},
number = {October},
pages = {1--8},
title = {{Fast monitoring of epileptic seizures using recurrence time statistics of electroencephalography}},
volume = {7},
year = {2013}
}

@article{Marwan07,
author = {MARWAN, N and CARMENROMANO, M and THIEL, M and KURTHS, J},
doi = {10.1016/j.physrep.2006.11.001},
issn = {03701573},
journal = {Physics Reports},
month = {jan},
number = {5-6},
pages = {237--329},
title = {{Recurrence plots for the analysis of complex systems}},
url = {https://linkinghub.elsevier.com/retrieve/pii/S0370157306004066},
volume = {438},
year = {2007}
}

@article{Martinez-Rego16,
author = {Mart{\'{i}}nez-Rego, David and Fontenla-Romero, Oscar and Alonso-Betanzos, Amparo and Principe, Jos{\'{e}} C.},
doi = {10.1016/j.patrec.2016.07.019},
issn = {01678655},
journal = {Pattern Recognition Letters},
pages = {8--14},
title = {{Fault detection via recurrence time statistics and one-class classification}},
volume = {84},
year = {2016}
}

@article{Gao99,
author = {Gao, J. B.},
doi = {10.1103/PhysRevLett.83.3178},
issn = {10797114},
journal = {Physical Review Letters},
number = {15},
pages = {3178--3181},
title = {{Recurrence time statistics for chaotic systems and their applications}},
volume = {83},
year = {1999}
}

@misc{Kennel97,
archivePrefix = {arXiv},
arxivId = {chao-dyn/9512005},
author = {Kennel, Matthew B.},
doi = {10.1103/PhysRevE.56.316},
eprint = {9512005},
isbn = {1063-651X},
issn = {1063651X},
journal = {Physical Review E - Statistical Physics, Plasmas, Fluids, and Related Interdisciplinary Topics},
number = {1},
pages = {316--321},
primaryClass = {chao-dyn},
title = {{Statistical test for dynamical nonstationarity in observed time-series data}},
volume = {56},
year = {1997}
}

@article{Rieke02,
author = {Rieke, Christoph and Sternickel, Karsten and Andrzejak, Ralph G. and Elger, Christian E. and David, Peter and Lehnertz, Klaus},
doi = {10.1103/PhysRevLett.88.244102},
issn = {10797114},
journal = {Physical Review Letters},
number = {24},
pages = {4},
pmid = {12059301},
title = {{Measuring Nonstationarity by Analyzing the Loss of Recurrence in Dynamical Systems}},
volume = {88},
year = {2002}
}

@article{Rieke04,
author = {Rieke, Christoph and Andrzejak, Ralph G. and Mormann, Florian and Lehnertz, Klaus},
doi = {10.1103/PhysRevE.69.046111},
issn = {1063651X},
journal = {Physical Review E - Statistical Physics, Plasmas, Fluids, and Related Interdisciplinary Topics},
number = {4},
pages = {9},
pmid = {15169073},
title = {{Improved statistical test for nonstationarity using recurrence time statistics}},
volume = {69},
year = {2004}
}

@article{Carletti06,
author = {Carletti, Timoteo and Galatolo, Stefano},
doi = {10.1016/j.physa.2005.10.003},
issn = {03784371},
journal = {Physica A: Statistical Mechanics and its Applications},
pages = {120--128},
title = {{Numerical estimates of local dimension by waiting time and quantitative recurrence}},
volume = {364},
year = {2006}
}

@article{Ichimaru99,
author = {Ichimaru, Y. and Moody, G. B.},
doi = {10.1046/j.1440-1819.1999.00527.x},
issn = {13231316},
journal = {Psychiatry and Clinical Neurosciences},
title = {{Development of the polysomnographic database on CD-ROM}},
year = {1999}
}

@article{Goldberger00,
author = {Goldberger, A. L. and Amaral, L. A. and Glass, L. and Hausdorff, J. M. and Ivanov, P. C. and Mark, R. G. and Mietus, J. E. and Moody, G. B. and Peng, C. K. and Stanley, H. E.},
issn = {15244539},
journal = {Circulation},
title = {{PhysioBank, PhysioToolkit, and PhysioNet: components of a new research resource for complex physiologic signals.}},
year = {2000}
}

@article{May76,
author = {May, Robert M},
doi = {10.1038/261459a0},
issn = {1476-4687},
journal = {Nature},
number = {5560},
pages = {459--467},
title = {{Simple mathematical models with very complicated dynamics}},
url = {https://doi.org/10.1038/261459a0},
volume = {261},
year = {1976}
}

@article{Ryzhii14,
author = {Ryzhii, E. and Ryzhii, M.},
issn = {18727565},
journal = {Computer Methods and Programs in Biomedicine},
number = {1},
pages = {40--49},
publisher = {Elsevier Ireland Ltd},
title = {{A heterogeneous coupled oscillator model for simulation of ECG signals}},
url = {http://dx.doi.org/10.1016/j.cmpb.2014.04.009},
volume = {117},
year = {2014}
}

@inproceedings{Kriegel09,
author = {Kriegel, Hans Peter and Kr{\"{o}}ger, Peer and Schubert, Erich and Zimek, Arthur},
booktitle = {International Conference on Information and Knowledge Management, Proceedings},
doi = {10.1145/1645953.1646195},
isbn = {9781605585123},
title = {{LoOP: Local outlier probabilities}},
year = {2009}
}

@article{Breuniq00,
author = {Breunig, Markus M. and Kriegel, Hans-Peter and Ng, Raymond T. and Sander, J\"{o}rg},
doi = {10.1145/335191.335388},
issn = {01635808},
journal = {SIGMOD Record (ACM Special Interest Group on Management of Data)},
title = {{LOF: Identifying density-based local outliers}},
year = {2000}
}

@article{Bradley97,
author = {Bradley, Andrew P.},
doi = {10.1016/S0031-3203(96)00142-2},
issn = {00313203},
journal = {Pattern Recognition},
number = {7},
pages = {1145--1159},
title = {{The use of the area under the ROC curve in the evaluation of machine learning algorithms}},
volume = {30},
year = {1997}
}

@article{Rhodes97,
author = {Rhodes, Carl and Morari, Manfred},
doi = {10.1016/S0098-1354(97)87657-0},
isbn = {0098-1354},
issn = {00981354},
journal = {Computers {\&} Chemical Engineering},
title = {{The false nearest neighbors algorithm: An overview}},
year = {1997}
}

@article{Krakovska15,
author = {Krakovsk{\'{a}}, Anna and Mezeiov{\'{a}}, Krist{\'{i}}na and Bud{\'{a}}{\v{c}}ov{\'{a}}, Hana},
doi = {10.1155/2015/932750},
issn = {2356-7244},
journal = {Journal of Complex Systems},
pages = {1--12},
title = {{Use of False Nearest Neighbours for Selecting Variables and Embedding Parameters for State Space Reconstruction}},
volume = {2015},
year = {2015}
}

@inproceedings{Gautama03,
author = {Gautama, Temujin and Mandic, Danilo P. and {Van Hulle}, Marc M.},
booktitle = {ICASSP, IEEE International Conference on Acoustics, Speech and Signal Processing - Proceedings},
doi = {10.1109/icassp.2003.1201610},
issn = {15206149},
pages = {29--32},
title = {{A differential entropy based method for determining the optimal embedding parameters of a signal}},
volume = {6},
year = {2003}
}

@book{Taleb07,
author = {Taleb, Nassim Nicholas},
booktitle = {The Review of Austrian Economics},
isbn = {9781400063512},
issn = {08893047},
title = {{The Black Swan: The Impact of the Highly Improbable}},
year = {2007}
}

@article{Benko19,

author = {Benk\H{o}, Zsigmond and Moldov{\'{a}}n, Kinga and Sz{\'{a}}deczky-Kardoss, Katalin and Zal{\'{a}}nyi, L{\'{a}}szl{\'{o}} and Borb{\'{e}}ly, S{\'{a}}ndor and Vil{\'{a}}gi, Ildik{\'{o}} and Somogyv{\'{a}}ri, Zolt{\'{a}}n},
doi = {10.1038/s41598-019-41554-x},
issn = {20452322},
journal = {Scientific Reports},
number = {1},
pages = {1--12},
title = {{Causal relationship between local field potential and intrinsic optical signal in epileptiform activity in vitro}},
volume = {9},
year = {2019}
}

@article{Selmeczy19,
author = {Selmeczy, G{\'{e}}za B. and Abonyi, Andr{\'{a}}s and Krienitz, Lothar and Kasprzak, Peter and Casper, Peter and Telcs, Andr{\'{a}}s and Somogyv{\'{a}}ri, Zolt{\'{a}}n and Padis{\'{a}}k, Judit},
doi = {10.1007/s10750-018-3793-7},
issn = {15735117},
journal = {Hydrobiologia},
title = {{Old sins have long shadows: climate change weakens efficiency of trophic coupling of phyto- and zooplankton in a deep oligo-mesotrophic lowland lake (Stechlin, Germany)—a causality analysis}},
year = {2019}
}

@article{Benko18,
journal = {arXiv:1808.10806},
author = {Benk\H{o}, Zsigmond and Zlatniczki, {\'{A}}d{\'{a}}m and Stippinger, Marcell and Fab{\'{o}}, D{\'{a}}niel and S{\'{o}}lyom, Andr{\'{a}}s and Er\H{o}ss, Lor{\'{a}}nd and Telcs, Andr{\'{a}}s and Somogyv{\'{a}}ri, Zolt{\'{a}}n},
eprint = {1808.10806},
month = {aug},
pages = {1--43},
title = {{Complete Inference of Causal Relations between Dynamical Systems}},
year = {2018}
}

@article{Packard80,
author = {Packard, N. H. and Crutchfield, J. P. and Farmer, J. D. and Shaw, R. S.},
doi = {10.1103/PhysRevLett.45.712},
issn = {0031-9007},
journal = {Physical Review Letters},
month = {sep},
number = {9},
pages = {712--716},
title = {{Geometry from a Time Series}},
url = {https://link.aps.org/doi/10.1103/PhysRevLett.45.712},
volume = {45},
year = {1980}
}

@article{Oehmcke15,
author = {Oehmcke, S and Zielinski, O and Kramer, O},
editor = {{Holldobler S. Krotzsch M.}, Rudolph S Penaloza R},
isbn = {9783319244884},
issn = {03029743},
journal = {Lecture Notes in Computer Science (including subseries Lecture Notes in Artificial Intelligence and Lecture Notes in Bioinformatics)},
pages = {279--286},
publisher = {Springer Verlag},
title = {Event detection in marine time series data},
volume = {9324},
year = {2015}
}

@ARTICLE{2020SciPy-NMeth,
       author = {{Virtanen}, Pauli and {Gommers}, Ralf and {Oliphant},
         Travis E. and {Haberland}, Matt and {Reddy}, Tyler and
         {Cournapeau}, David and {Burovski}, Evgeni and {Peterson}, Pearu
         and {Weckesser}, Warren and {Bright}, Jonathan and {van der Walt},
         St{\'e}fan J.  and {Brett}, Matthew and {Wilson}, Joshua and
         {Jarrod Millman}, K.  and {Mayorov}, Nikolay and {Nelson}, Andrew
         R.~J. and {Jones}, Eric and {Kern}, Robert and {Larson}, Eric and
         {Carey}, CJ and {Polat}, {\.I}lhan and {Feng}, Yu and {Moore},
         Eric W. and {Vand erPlas}, Jake and {Laxalde}, Denis and
         {Perktold}, Josef and {Cimrman}, Robert and {Henriksen}, Ian and
         {Quintero}, E.~A. and {Harris}, Charles R and {Archibald}, Anne M.
         and {Ribeiro}, Ant{\^o}nio H. and {Pedregosa}, Fabian and
         {van Mulbregt}, Paul and {Contributors}, SciPy 1. 0},
        title = "{SciPy 1.0: Fundamental Algorithms for Scientific
                  Computing in Python}",
      journal = {Nature Methods},
      year = "2020",
      volume={17},
      pages={261--272},
      adsurl = {https://rdcu.be/b08Wh},
      doi = {https://doi.org/10.1038/s41592-019-0686-2},
}

@article{scikit-learn,
 title={Scikit-learn: Machine Learning in {P}ython},
 author={Pedregosa, F. and Varoquaux, G. and Gramfort, A. and Michel, V.
         and Thirion, B. and Grisel, O. and Blondel, M. and Prettenhofer, P.
         and Weiss, R. and Dubourg, V. and Vanderplas, J. and Passos, A. and
         Cournapeau, D. and Brucher, M. and Perrot, M. and Duchesnay, E.},
 journal={Journal of Machine Learning Research},
 volume={12},
 pages={2825--2830},
 year={2011}
}

@inproceedings{Szepesvari07,
address = {New York, New York, USA},
author = {{Massoud Farahmand}, Amir and Szepesv{\'{a}}ri, Csaba and Audibert, Jean-Yves},
booktitle = {Proceedings of the 24th international conference on Machine learning - ICML '07},
pages = {265--272},
publisher = {ACM Press},
title = {{Manifold-adaptive dimension estimation}},
year = {2007}
}

@Misc{DeptJustice12,
  author = 	 {{Department of Justice of The United States}},
  title = 	 {Barclays Bank {PLC} Admits Misconduct Related to Submissions for the London Interbank Offered Rate and the Euro Interbank Offered Rate and Agrees to Pay \$160 Million Penalty},
  howpublished = {https://www.justice.gov/opa/pr/barclays-bank-plc-admits-misconduct-related-submissions-london-interbank-offered-rate-and},
  month = 	 {Jun 27},
  year = 	 2012}

@Article{SniderYoule09,
  author = 	 {Connan Snider and Thomas Youle},
  title = 	 {Diagnosing the Libor: strategic manipulation member portfolio positions},
  journal = 	 {Working paper- faculty.washington.edu},
  year = 	 2009}

@Article{SniderYoule10,
  author = {Connan Snider and Thomas Youle},
  title = {Does the Libor Reflect Banks' Borrowing Costs?},
  journal = {Social Science Research Network: SSRN.1569603},
  year =  {2010},
  month = {April}
}

@Article{SniderYoule12,
  author = {Connan Snider and Thomas Youle},
  title = {The Fix Is in: Detecting Portfolio Driven Manipulation of the Libor},
  journal = {Social Science Research Network: SSRN.2189015},
  year = {2012},
  month ={April}
}

@misc{Braei20,
archivePrefix = {arXiv},
arxivId = {2004.00433},
author = {Braei, Mohammad and Wagner, Sebastian},
eprint = {2004.00433},
month = {apr},
title = {{Anomaly Detection in Univariate Time-series: A Survey on the State-of-the-Art}},
url = {http://arxiv.org/abs/2004.00433},
year = {2020}
}

@article{Qi20,
author = {Di Qi and Andrew J. Majda},
doi = {10.1073/pnas.1917285117},
issn = {10916490},
journal = {Proceedings of the National Academy of Sciences of the United States of America},
number = {1},
pages = {52--59},
title = {{Using machine learning to predict extreme events in complex systems}},
volume = {117},
year = {2020}
}

@misc{Chalapathy19,
arxivId = {1901.03407},
author = {Chalapathy, Raghavendra and Chawla, Sanjay},
eprint = {1901.03407},
month = {jan},
pages = {1--50},
title = {{Deep Learning for Anomaly Detection: A Survey}},
url = {http://arxiv.org/abs/1901.03407},
year = {2019}
}

@article{Kwon19,
author = {Donghwoon Kwon and Hyunjoo Kim and Jinoh Kim and Sang C. Suh and Ikkyun Kim and Kuinam J. Kim},
doi = {10.1007/s10586-017-1117-8},
issn = {15737543},
journal = {Cluster Computing},
title = {A survey of deep learning-based network anomaly detection},
year = {2019}
}

@misc{Abbott19,
arxivId = {1912.11716},
author = {Abbott, R. and Abbott, T. D. and Abraham, S. and Acernese, F. and Ackley, K. and Adams, C. and Adhikari, R. X. and Adya, V. B. and Affeldt, C. and Agathos, M. and Agatsuma, K. and Aggarwal, N. and Aguiar, O. D. and Aich, A. and Aiello, L. and Ain, A. and Ajith, P. and Allen, G. and Allocca, A. and Altin, P. A. and Amato, A. and Anand, S. and Ananyeva, A. and Anderson, S. B. and Anderson, W. G. and Angelova, S. V. and Ansoldi, S. and Antier, S. and Appert, S. and Arai, K. and Araya, M. C. and Areeda, J. S. and Ar{\`{e}}ne, M. and Arnaud, N. and Aronson, S. M. and Arun, K. G. and Ascenzi, S. and Ashton, G. and Aston, S. M. and Astone, P. and Aubin, F. and Aufmuth, P. and AultONeal, K. and Austin, C. and Avendano, V. and Babak, S. and Bacon, P. and Badaracco, F. and Bader, M. K. M. and Bae, S. and Baer, A. M. and Baird, J. and Baldaccini, F. and Ballardin, G. and Ballmer, S. W. and Bals, A. and Balsamo, A. and Baltus, G. and Banagiri, S. and Bankar, D. and Bankar, R. S. and Barayoga, J. C. and Barbieri, C. and Barish, B. C. and Barker, D. and Barkett, K. and Barneo, P. and Barone, F. and Barr, B. and Barsotti, L. and Barsuglia, M. and Barta, D. and Bartlett, J. and Bartos, I. and Bassiri, R. and Basti, A. and Bawaj, M. and Bayley, J. C. and Bazzan, M. and B{\'{e}}csy, B. and Bejger, M. and Belahcene, I. and Bell, A. S. and Beniwal, D. and Benjamin, M. G. and Bentley, J. D. and Bergamin, F. and Berger, B. K. and Bergmann, G. and Bernuzzi, S. and Berry, C. P. L. and Bersanetti, D. and Bertolini, A. and Betzwieser, J. and Bhandare, R. and Bhandari, A. V. and Bidler, J. and Biggs, E. and Bilenko, I. A. and Billingsley, G. and Birney, R. and Birnholtz, O. and Biscans, S. and Bischi, M. and Biscoveanu, S. and Bisht, A. and Bissenbayeva, G. and Bitossi, M. and Bizouard, M. A. and Blackburn, J. K. and Blackman, J. and Blair, C. D. and Blair, D. G. and Blair, R. M. and Bobba, F. and Bode, N. and Boer, M. and Boetzel, Y. and Bogaert, G. and Bondu, F. and Bonilla, E. and Bonnand, R. and Booker, P. and Boom, B. A. and Bork, R. and Boschi, V. and Bose, S. and Bossilkov, V. and Bosveld, J. and Bouffanais, Y. and Bozzi, A. and Bradaschia, C. and Brady, P. R. and Bramley, A. and Branchesi, M. and Brau, J. E. and Breschi, M. and Briant, T. and Briggs, J. H. and Brighenti, F. and Brillet, A. and Brinkmann, M. and Brockill, P. and Brooks, A. F. and Brooks, J. and Brown, D. D. and Brunett, S. and Bruno, G. and Bruntz, R. and Buikema, A. and Bulik, T. and Bulten, H. J. and Buonanno, A. and Buskulic, D. and Byer, R. L. and Cabero, M. and Cadonati, L. and Cagnoli, G. and Cahillane, C. and Bustillo, J. Calder{\'{o}}n and Callaghan, J. D. and Callister, T. A. and Calloni, E. and Camp, J. B. and Canepa, M. and Cannon, K. C. and Cao, H. and Cao, J. and Carapella, G. and Carbognani, F. and Caride, S. and Carney, M. F. and Carullo, G. and Diaz, J. Casanueva and Casentini, C. and Casta{\~{n}}eda, J. and Caudill, S. and Cavagli{\`{a}}, M. and Cavalier, F. and Cavalieri, R. and Cella, G. and Cerd{\'{a}}-Dur{\'{a}}n, P. and Cesarini, E. and Chaibi, O. and Chakravarti, K. and Chan, C. and Chan, M. and Chao, S. and Charlton, P. and Chase, E. A. and Chassande-Mottin, E. and Chatterjee, D. and Chaturvedi, M. and Chen, H. Y. and Chen, X. and Chen, Y. and Cheng, H. -P. and Cheong, C. K. and Chia, H. Y. and Chiadini, F. and Chierici, R. and Chincarini, A. and Chiummo, A. and Cho, G. and Cho, H. S. and Cho, M. and Christensen, N. and Chu, Q. and Chua, S. and Chung, K. W. and Chung, S. and Ciani, G. and Ciecielag, P. and Cie{\{}{\'{s}}{\}}lar, M. and Ciobanu, A. A. and Ciolfi, R. and Cipriano, F. and Cirone, A. and Clara, F. and Clark, J. A. and Clearwater, P. and Clesse, S. and Cleva, F. and Coccia, E. and Cohadon, P. -F. and Cohen, D. and Colleoni, M. and Collette, C. G. and Collins, C. and Colpi, M. and Constancio, M. and Conti, L. and Cooper, S. J. and Corban, P. and Corbitt, T. R. and Cordero-Carri{\'{o}}n, I. and Corezzi, S. and Corley, K. R. and Cornish, N. and Corre, D. and Corsi, A. and Cortese, S. and Costa, C. A. and Cotesta, R. and Coughlin, M. W. and Coughlin, S. B. and Coulon, J. -P. and Countryman, S. T. and Couvares, P. and Covas, P. B. and Coward, D. M. and Cowart, M. J. and Coyne, D. C. and Coyne, R. and Creighton, J. D. E. and Creighton, T. D. and Cripe, J. and Croquette, M. and Crowder, S. G. and Cudell, J. -R. and Cullen, T. J. and Cumming, A. and Cummings, R. and Cunningham, L. and Cuoco, E. and Curylo, M. and Canton, T. Dal and D{\'{a}}lya, G. and Dana, A. and Daneshgaran-Bajastani, L. M. and D'Angelo, B. and Danilishin, S. L. and D'Antonio, S. and Danzmann, K. and Darsow-Fromm, C. and Dasgupta, A. and Datrier, L. E. H. and Dattilo, V. and Dave, I. and Davier, M. and Davies, G. S. and Davis, D. and Daw, E. J. and DeBra, D. and Deenadayalan, M. and Degallaix, J. and {De Laurentis}, M. and Del{\'{e}}glise, S. and Delfavero, M. and {De Lillo}, N. and {Del Pozzo}, W. and DeMarchi, L. M. and D'Emilio, V. and Demos, N. and Dent, T. and {De Pietri}, R. and {De Rosa}, R. and {De Rossi}, C. and DeSalvo, R. and de Varona, O. and Dhurandhar, S. and D{\'{i}}az, M. C. and Diaz-Ortiz, M. and Dietrich, T. and {Di Fiore}, L. and {Di Fronzo}, C. and {Di Giorgio}, C. and {Di Giovanni}, F. and {Di Giovanni}, M. and {Di Girolamo}, T. and {Di Lieto}, A. and Ding, B. and {Di Pace}, S. and {Di Palma}, I. and {Di Renzo}, F. and Divakarla, A. K. and Dmitriev, A. and Doctor, Z. and Donovan, F. and Dooley, K. L. and Doravari, S. and Dorrington, I. and Downes, T. P. and Drago, M. and Driggers, J. C. and Du, Z. and Ducoin, J. -G. and Dupej, P. and Durante, O. and D'Urso, D. and Dwyer, S. E. and Easter, P. J. and Eddolls, G. and Edelman, B. and Edo, T. B. and Edy, O. and Effler, A. and Ehrens, P. and Eichholz, J. and Eikenberry, S. S. and Eisenmann, M. and Eisenstein, R. A. and Ejlli, A. and Errico, L. and Essick, R. C. and Estelles, H. and Estevez, D. and Etienne, Z. B. and Etzel, T. and Evans, M. and Evans, T. M. and Ewing, B. E. and Fafone, V. and Fairhurst, S. and Fan, X. and Farinon, S. and Farr, B. and Farr, W. M. and Fauchon-Jones, E. J. and Favata, M. and Fays, M. and Fazio, M. and Feicht, J. and Fejer, M. M. and Feng, F. and Fenyvesi, E. and Ferguson, D. L. and Fernandez-Galiana, A. and Ferrante, I. and Ferreira, E. C. and Ferreira, T. A. and Fidecaro, F. and Fiori, I. and Fiorucci, D. and Fishbach, M. and Fisher, R. P. and Fittipaldi, R. and Fitz-Axen, M. and Fiumara, V. and Flaminio, R. and Floden, E. and Flynn, E. and Fong, H. and Font, J. A. and Forsyth, P. W. F. and Fournier, J. -D. and Frasca, S. and Frasconi, F. and Frei, Z. and Freise, A. and Frey, R. and Frey, V. and Fritschel, P. and Frolov, V. V. and Fronz{\`{e}}, G. and Fulda, P. and Fyffe, M. and Gabbard, H. A. and Gadre, B. U. and Gaebel, S. M. and Gair, J. R. and Galaudage, S. and Ganapathy, D. and Gaonkar, S. G. and Garc{\'{i}}a-Quir{\'{o}}s, C. and Garufi, F. and Gateley, B. and Gaudio, S. and Gayathri, V. and Gemme, G. and Genin, E. and Gennai, A. and George, D. and George, J. and Gergely, L. and Ghonge, S. and Ghosh, Abhirup and Ghosh, Archisman and Ghosh, S. and Giacomazzo, B. and Giaime, J. A. and Giardina, K. D. and Gibson, D. R. and Gier, C. and Gill, K. and Glanzer, J. and Gniesmer, J. and Godwin, P. and Goetz, E. and Goetz, R. and Gohlke, N. and Goncharov, B. and Gonz{\'{a}}lez, G. and Gopakumar, A. and Gossan, S. E. and Gosselin, M. and Gouaty, R. and Grace, B. and Grado, A. and Granata, M. and Grant, A. and Gras, S. and Grassia, P. and Gray, C. and Gray, R. and Greco, G. and Green, A. C. and Green, R. and Gretarsson, E. M. and Griggs, H. L. and Grignani, G. and Grimaldi, A. and Grimm, S. J. and Grote, H. and Grunewald, S. and Gruning, P. and Guidi, G. M. and Guimaraes, A. R. and Guix{\'{e}}, G. and Gulati, H. K. and Guo, Y. and Gupta, A. and Gupta, Anchal and Gupta, P. and Gustafson, E. K. and Gustafson, R. and Haegel, L. and Halim, O. and Hall, E. D. and Hamilton, E. Z. and Hammond, G. and Haney, M. and Hanke, M. M. and Hanks, J. and Hanna, C. and Hannam, M. D. and Hannuksela, O. A. and Hansen, T. J. and Hanson, J. and Harder, T. and Hardwick, T. and Haris, K. and Harms, J. and Harry, G. M. and Harry, I. W. and Hasskew, R. K. and Haster, C. -J. and Haughian, K. and Hayes, F. J. and Healy, J. and Heidmann, A. and Heintze, M. C. and Heinze, J. and Heitmann, H. and Hellman, F. and Hello, P. and Hemming, G. and Hendry, M. and Heng, I. S. and Hennes, E. and Hennig, J. and Heurs, M. and Hild, S. and Hinderer, T. and Hoback, S. Y. and Hochheim, S. and Hofgard, E. and Hofman, D. and Holgado, A. M. and Holland, N. A. and Holt, K. and Holz, D. E. and Hopkins, P. and Horst, C. and Hough, J. and Howell, E. J. and Hoy, C. G. and Huang, Y. and H{\"{u}}bner, M. T. and Huerta, E. A. and Huet, D. and Hughey, B. and Hui, V. and Husa, S. and Huttner, S. H. and Huxford, R. and Huynh-Dinh, T. and Idzkowski, B. and Iess, A. and Inchauspe, H. and Ingram, C. and Intini, G. and Isac, J. -M. and Isi, M. and Iyer, B. R. and Jacqmin, T. and Jadhav, S. J. and Jadhav, S. P. and James, A. L. and Jani, K. and Janthalur, N. N. and Jaranowski, P. and Jariwala, D. and Jaume, R. and Jenkins, A. C. and Jiang, J. and Johns, G. R. and Jones, A. W. and Jones, D. I. and Jones, J. D. and Jones, P. and Jones, R. and Jonker, R. J. G. and Ju, L. and Junker, J. and Kalaghatgi, C. V. and Kalogera, V. and Kamai, B. and Kandhasamy, S. and Kang, G. and Kanner, J. B. and Kapadia, S. J. and Karki, S. and Kashyap, R. and Kasprzack, M. and Kastaun, W. and Katsanevas, S. and Katsavounidis, E. and Katzman, W. and Kaufer, S. and Kawabe, K. and K{\'{e}}f{\'{e}}lian, F. and Keitel, D. and Keivani, A. and Kennedy, R. and Key, J. S. and Khadka, S. and Khalili, F. Y. and Khan, I. and Khan, S. and Khan, Z. A. and Khazanov, E. A. and Khetan, N. and Khursheed, M. and Kijbunchoo, N. and Kim, Chunglee and Kim, G. J. and Kim, J. C. and Kim, K. and Kim, W. and Kim, W. S. and Kim, Y. -M. and Kimball, C. and King, P. J. and Kinley-Hanlon, M. and Kirchhoff, R. and Kissel, J. S. and Kleybolte, L. and Klimenko, S. and Knowles, T. D. and Koch, P. and Koehlenbeck, S. M. and Koekoek, G. and Koley, S. and Kondrashov, V. and Kontos, A. and Koper, N. and Korobko, M. and Korth, W. Z. and Kovalam, M. and Kozak, D. B. and Kringel, V. and Krishnendu, N. V. and Kr{\'{o}}lak, A. and Krupinski, N. and Kuehn, G. and Kumar, A. and Kumar, P. and Kumar, Rahul and Kumar, Rakesh and Kumar, S. and Kuo, L. and Kutynia, A. and Lackey, B. D. and Laghi, D. and Lalande, E. and Lam, T. L. and Lamberts, A. and Landry, M. and Lane, B. B. and Lang, R. N. and Lange, J. and Lantz, B. and Lanza, R. K. and {La Rosa}, I. and Lartaux-Vollard, A. and Lasky, P. D. and Laxen, M. and Lazzarini, A. and Lazzaro, C. and Leaci, P. and Leavey, S. and Lecoeuche, Y. K. and Lee, C. H. and Lee, H. M. and Lee, H. W. and Lee, J. and Lee, K. and Lehmann, J. and Leroy, N. and Letendre, N. and Levin, Y. and Li, A. K. Y. and Li, J. and Li, K. and Li, T. G. F. and Li, X. and Linde, F. and Linker, S. D. and Linley, J. N. and Littenberg, T. B. and Liu, J. and Liu, X. and Llorens-Monteagudo, M. and Lo, R. K. L. and Lockwood, A. and London, L. T. and Longo, A. and Lorenzini, M. and Loriette, V. and Lormand, M. and Losurdo, G. and Lough, J. D. and Lousto, C. O. and Lovelace, G. and L{\"{u}}ck, H. and Lumaca, D. and Lundgren, A. P. and Ma, Y. and Macas, R. and Macfoy, S. and MacInnis, M. and Macleod, D. M. and MacMillan, I. A. O. and Macquet, A. and Hernandez, I. Maga{\~{n}}a and Maga{\~{n}}a-Sandoval, F. and Magee, R. M. and Majorana, E. and Maksimovic, I. and Malik, A. and Man, N. and Mandic, V. and Mangano, V. and Mansell, G. L. and Manske, M. and Mantovani, M. and Mapelli, M. and Marchesoni, F. and Marion, F. and M{\'{a}}rka, S. and M{\'{a}}rka, Z. and Markakis, C. and Markosyan, A. S. and Markowitz, A. and Maros, E. and Marquina, A. and Marsat, S. and Martelli, F. and Martin, I. W. and Martin, R. M. and Martinez, V. and Martynov, D. V. and Masalehdan, H. and Mason, K. and Massera, E. and Masserot, A. and Massinger, T. J. and Masso-Reid, M. and Mastrogiovanni, S. and Matas, A. and Matichard, F. and Mavalvala, N. and Maynard, E. and McCann, J. J. and McCarthy, R. and McClelland, D. E. and McCormick, S. and McCuller, L. and McGuire, S. C. and McIsaac, C. and McIver, J. and McManus, D. J. and McRae, T. and McWilliams, S. T. and Meacher, D. and Meadors, G. D. and Mehmet, M. and Mehta, A. K. and Villa, E. Mejuto and Melatos, A. and Mendell, G. and Mercer, R. A. and Mereni, L. and Merfeld, K. and Merilh, E. L. and Merritt, J. D. and Merzougui, M. and Meshkov, S. and Messenger, C. and Messick, C. and Metzdorff, R. and Meyers, P. M. and Meylahn, F. and Mhaske, A. and Miani, A. and Miao, H. and Michaloliakos, I. and Michel, C. and Middleton, H. and Milano, L. and Miller, A. L. and Millhouse, M. and Mills, J. C. and Milotti, E. and Milovich-Goff, M. C. and Minazzoli, O. and Minenkov, Y. and Mishkin, A. and Mishra, C. and Mistry, T. and Mitra, S. and Mitrofanov, V. P. and Mitselmakher, G. and Mittleman, R. and Mo, G. and Mogushi, K. and Mohapatra, S. R. P. and Mohite, S. R. and Molina-Ruiz, M. and Mondin, M. and Montani, M. and Moore, C. J. and Moraru, D. and Morawski, F. and Moreno, G. and Morisaki, S. and Mours, B. and Mow-Lowry, C. M. and Mozzon, S. and Muciaccia, F. and Mukherjee, Arunava and Mukherjee, D. and Mukherjee, S. and Mukherjee, Subroto and Mukund, N. and Mullavey, A. and Munch, J. and Mu{\~{n}}iz, E. A. and Murray, P. G. and Nagar, A. and Nardecchia, I. and Naticchioni, L. and Nayak, R. K. and Neil, B. F. and Neilson, J. and Nelemans, G. and Nelson, T. J. N. and Nery, M. and Neunzert, A. and Ng, K. Y. and Ng, S. and Nguyen, C. and Nguyen, P. and Nichols, D. and Nichols, S. A. and Nissanke, S. and Nocera, F. and Noh, M. and North, C. and Nothard, D. and Nuttall, L. K. and Oberling, J. and O'Brien, B. D. and Oganesyan, G. and Ogin, G. H. and Oh, J. J. and Oh, S. H. and Ohme, F. and Ohta, H. and Okada, M. A. and Oliver, M. and Olivetto, C. and Oppermann, P. and Oram, Richard J. and O'Reilly, B. and Ormiston, R. G. and Ortega, L. F. and O'Shaughnessy, R. and Ossokine, S. and Osthelder, C. and Ottaway, D. J. and Overmier, H. and Owen, B. J. and Pace, A. E. and Pagano, G. and Page, M. A. and Pagliaroli, G. and Pai, A. and Pai, S. A. and Palamos, J. R. and Palashov, O. and Palomba, C. and Pan, H. and Panda, P. K. and Pang, P. T. H. and Pankow, C. and Pannarale, F. and Pant, B. C. and Paoletti, F. and Paoli, A. and Parida, A. and Parker, W. and Pascucci, D. and Pasqualetti, A. and Passaquieti, R. and Passuello, D. and Patricelli, B. and Payne, E. and Pearlstone, B. L. and Pechsiri, T. C. and Pedersen, A. J. and Pedraza, M. and Pele, A. and Penn, S. and Perego, A. and Perez, C. J. and P{\'{e}}rigois, C. and Perreca, A. and Perri{\`{e}}s, S. and Petermann, J. and Pfeiffer, H. P. and Phelps, M. and Phukon, K. S. and Piccinni, O. J. and Pichot, M. and Piendibene, M. and Piergiovanni, F. and Pierro, V. and Pillant, G. and Pinard, L. and Pinto, I. M. and Piotrzkowski, K. and Pirello, M. and Pitkin, M. and Plastino, W. and Poggiani, R. and Pong, D. Y. T. and Ponrathnam, S. and Popolizio, P. and Porter, E. K. and Powell, J. and Prajapati, A. K. and Prasai, K. and Prasanna, R. and Pratten, G. and Prestegard, T. and Principe, M. and Prodi, G. A. and Prokhorov, L. and Punturo, M. and Puppo, P. and P{\"{u}}rrer, M. and Qi, H. and Quetschke, V. and Quinonez, P. J. and Raab, F. J. and Raaijmakers, G. and Radkins, H. and Radulesco, N. and Raffai, P. and Rafferty, H. and Raja, S. and Rajan, C. and Rajbhandari, B. and Rakhmanov, M. and Ramirez, K. E. and Ramos-Buades, A. and Rana, Javed and Rao, K. and Rapagnani, P. and Raymond, V. and Razzano, M. and Read, J. and Regimbau, T. and Rei, L. and Reid, S. and Reitze, D. H. and Rettegno, P. and Ricci, F. and Richardson, C. J. and Richardson, J. W. and Ricker, P. M. and Riemenschneider, G. and Riles, K. and Rizzo, M. and Robertson, N. A. and Robinet, F. and Rocchi, A. and Rodriguez-Soto, R. D. and Rolland, L. and Rollins, J. G. and Roma, V. J. and Romanelli, M. and Romano, R. and Romel, C. L. and Romero-Shaw, I. M. and Romie, J. H. and Rose, C. A. and Rose, D. and Rose, K. and Rosi{\'{n}}ska, D. and Rosofsky, S. G. and Ross, M. P. and Rowan, S. and Rowlinson, S. J. and Roy, P. K. and Roy, Santosh and Roy, Soumen and Ruggi, P. and Rutins, G. and Ryan, K. and Sachdev, S. and Sadecki, T. and Sakellariadou, M. and Salafia, O. S. and Salconi, L. and Saleem, M. and Samajdar, A. and Sanchez, E. J. and Sanchez, L. E. and Sanchis-Gual, N. and Sanders, J. R. and Santiago, K. A. and Santos, E. and Sarin, N. and Sassolas, B. and Sathyaprakash, B. S. and Sauter, O. and Savage, R. L. and Savant, V. and Sawant, D. and Sayah, S. and Schaetzl, D. and Schale, P. and Scheel, M. and Scheuer, J. and Schmidt, P. and Schnabel, R. and Schofield, R. M. S. and Sch{\"{o}}nbeck, A. and Schreiber, E. and Schulte, B. W. and Schutz, B. F. and Schwarm, O. and Schwartz, E. and Scott, J. and Scott, S. M. and Seidel, E. and Sellers, D. and Sengupta, A. S. and Sennett, N. and Sentenac, D. and Sequino, V. and Sergeev, A. and Setyawati, Y. and Shaddock, D. A. and Shaffer, T. and Shahriar, M. S. and Sharma, A. and Sharma, P. and Shawhan, P. and Shen, H. and Shikauchi, M. and Shink, R. and Shoemaker, D. H. and Shoemaker, D. M. and Shukla, K. and ShyamSundar, S. and Siellez, K. and Sieniawska, M. and Sigg, D. and Singer, L. P. and Singh, D. and Singh, N. and Singha, A. and Singhal, A. and Sintes, A. M. and Sipala, V. and Skliris, V. and Slagmolen, B. J. J. and Slaven-Blair, T. J. and Smetana, J. and Smith, J. R. and Smith, R. J. E. and Somala, S. and Son, E. J. and Soni, S. and Sorazu, B. and Sordini, V. and Sorrentino, F. and Souradeep, T. and Sowell, E. and Spencer, A. P. and Spera, M. and Srivastava, A. K. and Srivastava, V. and Staats, K. and Stachie, C. and Standke, M. and Steer, D. A. and Steinke, M. and Steinlechner, J. and Steinlechner, S. and Steinmeyer, D. and Stocks, D. and Stops, D. J. and Stover, M. and Strain, K. A. and Stratta, G. and Strunk, A. and Sturani, R. and Stuver, A. L. and Sudhagar, S. and Sudhir, V. and Summerscales, T. Z. and Sun, L. and Sunil, S. and Sur, A. and Suresh, J. and Sutton, P. J. and Swinkels, B. L. and Szczepa{\'{n}}czyk, M. J. and Tacca, M. and Tait, S. C. and Talbot, C. and Tanasijczuk, A. J. and Tanner, D. B. and Tao, D. and T{\'{a}}pai, M. and Tapia, A. and Martin, E. N. Tapia San and Tasson, J. D. and Taylor, R. and Tenorio, R. and Terkowski, L. and Thirugnanasambandam, M. P. and Thomas, M. and Thomas, P. and Thompson, J. E. and Thondapu, S. R. and Thorne, K. A. and Thrane, E. and Tinsman, C. L. and Saravanan, T. R. and Tiwari, Shubhanshu and Tiwari, S. and Tiwari, V. and Toland, K. and Tonelli, M. and Tornasi, Z. and Torres-Forn{\'{e}}, A. and Torrie, C. I. and Melo, I. Tosta e and T{\"{o}}yr{\"{a}}, D. and Trail, E. A. and Travasso, F. and Traylor, G. and Tringali, M. C. and Tripathee, A. and Trovato, A. and Trudeau, R. J. and Tsang, K. W. and Tse, M. and Tso, R. and Tsukada, L. and Tsuna, D. and Tsutsui, T. and Turconi, M. and Ubhi, A. S. and Ueno, K. and Ugolini, D. and Unnikrishnan, C. S. and Urban, A. L. and Usman, S. A. and Utina, A. C. and Vahlbruch, H. and Vajente, G. and Valdes, G. and Valentini, M. and Vallisneri, M. and van Bakel, N. and van Beuzekom, M. and van den Brand, J. F. J. and Broeck, C. Van Den and Vander-Hyde, D. C. and van der Schaaf, L. and {Van Heijningen}, J. V. and van Veggel, A. A. and Vardaro, M. and Varma, V. and Vass, S. and Vas{\'{u}}th, M. and Vecchio, A. and Vedovato, G. and Veitch, J. and Veitch, P. J. and Venkateswara, K. and Venugopalan, G. and Verkindt, D. and Veske, D. and Vetrano, F. and Vicer{\'{e}}, A. and Viets, A. D. and Vinciguerra, S. and Vine, D. J. and Vinet, J. -Y. and Vitale, S. and Vivanco, Francisco Hernandez and Vo, T. and Vocca, H. and Vorvick, C. and Vyatchanin, S. P. and Wade, A. R. and Wade, L. E. and Wade, M. and Walet, R. and Walker, M. and Wallace, G. S. and Wallace, L. and Walsh, S. and Wang, J. Z. and Wang, S. and Wang, W. H. and Wang, Y. F. and Ward, R. L. and Warden, Z. A. and Warner, J. and Was, M. and Watchi, J. and Weaver, B. and Wei, L. -W. and Weinert, M. and Weinstein, A. J. and Weiss, R. and Wellmann, F. and Wen, L. and We{\ss}els, P. and Westhouse, J. W. and Wette, K. and Whelan, J. T. and Whiting, B. F. and Whittle, C. and Wilken, D. M. and Williams, D. and Williams, R. D. and Willis, J. L. and Willke, B. and Winkler, W. and Wipf, C. C. and Wittel, H. and Woan, G. and Woehler, J. and Wofford, J. K. and Wong, C. and Wright, J. L. and Wu, D. S. and Wysocki, D. M. and Xiao, L. and Yamamoto, H. and Yang, L. and Yang, Y. and Yang, Z. and Yap, M. J. and Yazback, M. and Yeeles, D. W. and Yu, Hang and Yu, Haocun and Yuen, S. H. R. and Zadro{\.{z}}ny, A. K. and Zadro{\.{z}}ny, A. and Zanolin, M. and Zelenova, T. and Zendri, J. -P. and Zevin, M. and Zhang, J. and Zhang, L. and Zhang, T. and Zhao, C. and Zhao, G. and Zhou, M. and Zhou, Z. and Zhu, X. J. and Zimmerman, A. B. and Zucker, M. E. and Zweizig, J.},
eprint = {1912.11716},
title = {{Open data from the first and second observing runs of Advanced LIGO and Advanced Virgo}},
url = {http://arxiv.org/abs/1912.11716},
year = {2019}
}

@article{Brown15,
author = {Brown, Russell A},
issn = {2331-7418},
journal = {Journal of Computer Graphics Techniques (JCGT)},
month = {mar},
number = {1},
pages = {50--68},
title = {{Building a Balanced $k$-d Tree in $O(kn \log n)$ Time}},
url = {http://jcgt.org/published/0004/01/03/},
volume = {4},
year = {2015}
}

@article{Bentley75,
author = {Bentley, Jon Louis},
doi = {10.1145/361002.361007},
journal = {Communications of the ACM},
month = {sep},
number = {9},
pages = {509--517},
title = {{Multidimensional binary search trees used for associative searching}},
url = {https://dl.acm.org/doi/10.1145/361002.361007},
volume = {18},
year = {1975}
}

@article{Senin15,
author = {Senin, Pavel and Lin, Jessica and Wang, Xing and Oates, Tim and Gandhi, Sunil and Boedihardjo, Arnold P. and Chen, Crystal and Frankenstein, Susan},
doi = {10.5441/002/edbt.2015.42},
isbn = {9783893180677},
journal = {EDBT 2015 - 18th International Conference on Extending Database Technology, Proceedings},
pages = {481--492},
title = {{Time series anomaly discovery with grammar-based compression}},
year = {2015}
}

@software{duncan_macleod_2020_3598469,
  author       = {Duncan Macleod and
                  Alex L. Urban and
                  Scott Coughlin and
                  Thomas Massinger and
                  Matt Pitkin and
                  paulaltin and
                  Joseph Areeda and
                  Eric Quintero and
                  The Gitter Badger and
                  Leo Singer and
                  Katrin Leinweber},
  title        = {gwpy/gwpy: 1.0.1},
  month        = jan,
  year         = 2020,
  publisher    = {Zenodo},
  version      = {v1.0.1},
  doi          = {10.5281/zenodo.3598469},
  url          = {https://doi.org/10.5281/zenodo.3598469}
}

@article{jmotif,
  author       = {Senin, Pavel},
  title        = {jMotif},
  month        = {02},
  year         = {2020},
  version      = {1.1.1},
  journal          = {https://github.com/jMotif/jmotif-R}{jmotif}}

@article{Zevin_2017,
   title={Gravity Spy: integrating advanced LIGO detector characterization, machine learning, and citizen science},
   volume={34},
   ISSN={1361-6382},
   url={http://dx.doi.org/10.1088/1361-6382/aa5cea},
   DOI={10.1088/1361-6382/aa5cea},
   number={6},
   journal={Classical and Quantum Gravity},
   publisher={IOP Publishing},
   author={Zevin, M and Coughlin, S and Bahaadini, S and Besler, E and Rohani, N and Allen, S and Cabero, M and Crowston, K and Katsaggelos, A K and Larson, S L and et al.},
   year={2017},
   month={Feb},
   pages={064003}
}

@article{Abbott16c,
   title={GW150914: First results from the search for binary black hole coalescence with Advanced LIGO},
   volume={93},
   ISSN={2470-0029},
   url={http://dx.doi.org/10.1103/PhysRevD.93.122003},
   DOI={10.1103/physrevd.93.122003},
   number={12},
   journal={Physical Review D},
   publisher={American Physical Society (APS)},
   author={Abbott, B. P. and Abbott, R. and Abbott, T. D. and Abernathy, M. R. and Acernese, F. and Ackley, K. and Adams, C. and Adams, T. and Addesso, P. and Adhikari, R. X. and et al.},
   year={2016},
   month={Jun}
}
\end{refsection}

\end{document}